\documentclass{article}

\PassOptionsToPackage{numbers, compress, sort}{natbib}

\usepackage[eandd, final]{neurips_2026}

\usepackage[utf8]{inputenc} 
\usepackage[T1]{fontenc}    
\usepackage{hyperref}       
\usepackage{url}            
\usepackage{booktabs}       
\usepackage{amsfonts}       
\usepackage{nicefrac}       
\usepackage{microtype}      
\usepackage[table]{xcolor}  
\usepackage{multirow}

\usepackage{pifont}

\usepackage{enumitem}

\usepackage{amsmath}
\usepackage{amssymb}
\usepackage{bm}
\usepackage{mathtools}
\usepackage{stmaryrd}

\usepackage[capitalize,noabbrev]{cleveref}
\usepackage{caption}
\crefformat{equation}{(#2#1#3)}
\crefrangeformat{equation}{(#3#1#4--#5#2#6)}
\crefmultiformat{equation}{(#2#1#3)}{ and (#2#1#3)}{, (#2#1#3)}{ and (#2#1#3)}

\hypersetup{colorlinks=true, citecolor=blue!50!black, linkcolor=red!50!black, urlcolor=green!50!black} 

\usepackage{tikz}
\usetikzlibrary{arrows.meta}

\usepackage{graphicx}
\usepackage{adjustbox}
\usepackage{subcaption}
\usepackage{wrapfig}
\graphicspath{{figures/}{figures/icons/}}
\newcommand{\icon}[1]{
  \raisebox{-0.28\height}{
    \includegraphics[height=1.6em,keepaspectratio]{#1}
  }
}

\usepackage{titletoc} 

\usepackage{tabularx}
\usepackage{booktabs}
\usepackage{makecell}
\usepackage{pifont}
\newcommand{\cmark}{\textcolor{black!75}{\ding{51}}}
\newcommand{\xmark}{\textcolor{black!20}{\ding{55}}}
\usepackage{arydshln}
\newcommand{\codecell}[1]{{\footnotesize\fontfamily{\plexttfamily}\selectfont #1}}
\usepackage{arydshln}
\usepackage{mathabx}
\usepackage{dsfont}
\newcommand{\rowarrow}{\ensuremath{\drsh}\,}

\usepackage{tcolorbox}\tcbuselibrary{listings,minted,skins}
\definecolor{codebg}{HTML}{F8FAFC}
\definecolor{codeshadow}{HTML}{94A3B8}
\definecolor{badgebg}{HTML}{1E88E5} 
\definecolor{badgefg}{HTML}{FFFFFF} 
\makeatletter
\let\origttdefault\ttdefault 
\makeatother
\usepackage{plex-mono}
\makeatletter
\edef\plexttfamily{\ttdefault} 
\let\ttdefault\origttdefault 
\makeatother

\newcommand{\problybadge}[1]{%
  \node[
    anchor=north east,
    fill=badgebg,
    text=badgefg,
    font=\tiny\sffamily\bfseries,
    rounded corners=2pt,
    inner sep=1.7pt,
    xshift=4pt,
    yshift=7pt
  ]
  at (frame.north east) {\,probly $+$ #1\,};%
}
\newcommand{\problyimagebadge}[1]{%
  \node[
    overlay,
    anchor=north east,
    fill=badgebg,
    text=badgefg,
    font=\tiny\sffamily\bfseries,
    rounded corners=2pt,
    inner sep=1.7pt,
    xshift=4.2pt,
    yshift=7pt
  ]
  at (img.north east) {\,probly $+$ #1\,};%
}

\usepackage{xspace} 
\renewcommand{\texttt}[1]{%
  {\fontfamily{\plexttfamily}\fontsize{0.82em}{1em}\selectfont #1}%
}
\pdfmapline{+probly-wordmark ProblyWordmark <figures/appendix/probly-wordmark.enc <figures/appendix/probly-wordmark.ttf}
\newsavebox{\problyReference}

\DeclareRobustCommand{\probly}{%
  \texorpdfstring{\texttt{probly}\xspace}{probly}%
}
\newsavebox{\problyTitleReference}
\DeclareRobustCommand{\problyTitle}{%
  \texorpdfstring{%
    \begingroup
    \sbox{\problyTitleReference}{b}%
    \raisebox{-0.467857\ht\problyTitleReference}{%
      \includegraphics[height=1.65\ht\problyTitleReference]{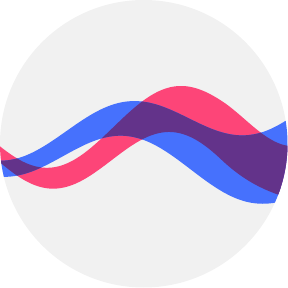}%
    }%
    \kern0.32em
    \raisebox{-0.285714\ht\problyTitleReference}{%
      \includegraphics[height=1.285714\ht\problyTitleReference,
        trim=7.5bp 7.5bp 7.5bp 7.5bp,clip]{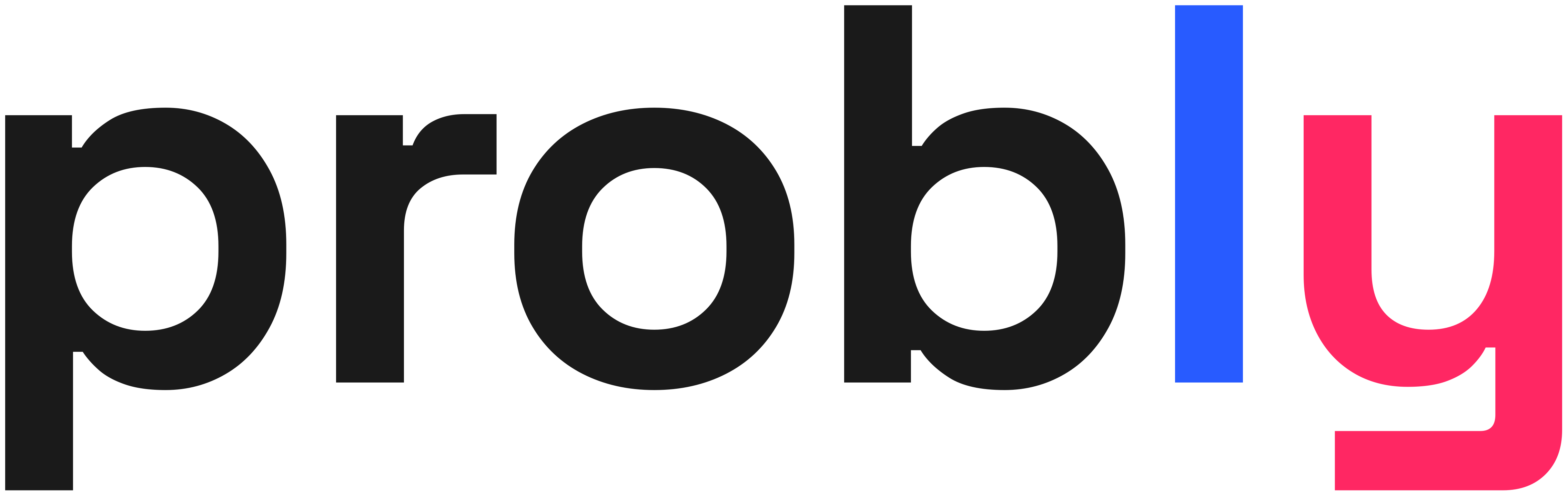}%
    }%
    \endgroup\xspace
  }{probly}%
}
\newcommand{\river}{\texttt{river}\xspace}
\newcommand{\sklearn}{\texttt{sklearn}\xspace} 
\newcommand{\torch}{\texttt{torch}\xspace} 
\newcommand{\flax}{\texttt{flax}\xspace}
\newcommand{\jax}{\texttt{jax}\xspace}
\newcommand{\githubrepo}{\url{https://github.com/pwhofman/probly}}

\usepackage{acro}
\acsetup{format/first-long=} 
\acsetup{barriers/use=true}
\acsetup{barriers/reset=true}
\acsetup{single}
\newcommand{\dac}[3]{\DeclareAcronym{#1}{short = #2, long = #3}}

\dac{ml}{ML}{machine learning}
\dac{nlp}{NLP}{natural language processing}
\dac{nn}{NN}{neural network}
\dac{mlp}{MLP}{multi-layer perceptron}
\dac{rnn}{RNN}{recurrent neural network}
\dac{cnn}{CNN}{convolutional neural network}
\dac{svm}{SVM}{support vector machine}
\dac{gnn}{GNN}{graph neural network}
\dac{ggpn}{GGPN}{Generalized Graph Posterior Network}
\dac{gpn}{GPN}{Graph Posterior Network}
\dac{cuqgnn}{CUQ-GNN}{Committee-based Uncertainty Quantification Graph Neural Network}
\dac{llop}{LLOP}{log-linear opinion pooling}
\dac{lop}{LOP}{linear opinion pooling}
\dac{lopgpn}{LOP-GPN}{linear opinion pooled graph posterior network}
\dac{bnn}{BNN}{Bayesian neural network}
\dac{appnp}{APPNP}{approximate personalized propagation of neural predictions}
\dac{ce}{CE}{cross-entropy}
\dac{uce}{UCE}{uncertain cross-entropy}
\dac{ppr}{PPR}{personalized page-rank}
\dac{arc}{ARC}{accuracy-rejection curve}
\dac{ood}{OoD}{out-of-distribution}
\dac{id}{iD}{in-distribution}
\dac{rv}{RV}{random variable}
\dac{cdf}{CDF}{cumulative distribution function}
\dac{pdf}{PDF}{probability density function}
\dac{auroc}{AUROC}{area under receiver operating characteristic}
\dac{uq}{UQ}{uncertainty quantification}
\dac{au}{AU}{aleatoric uncertainty}
\dac{eu}{EU}{epistemic uncertainty}
\dac{tu}{TU}{total uncertainty}
\dac{iid}{IID}{independent and identically distributed}
\dac{llm}{LLM}{Large Language Model}
\dac{nli}{NLI}{natural language inference}
\dac{arf}{ARF}{adaptive random forest}

\newcommand{\vw}{\bm{w}}
\newcommand{\vx}{\bm{x}}

\newcommand{\bR}{\mathbb{R}}
\newcommand{\cC}{\mathcal{C}}

\newcommand{\cQ}{\mathcal{Q}}
\newcommand{\cX}{\mathcal{X}}
\newcommand{\cY}{\mathcal{Y}}

\newcommand{\prob}{p}

\renewcommand{\vec}[1]{\boldsymbol{#1}}
\newcommand{\given}{\, | \,}
\newcommand{\fromto}{\longrightarrow}
\newcommand{\onecirc}{\ding{172}}
\newcommand{\twocirc}{\ding{173}}
\newcommand{\threecirc}{\ding{174}}
\newcommand{\fourcirc}{\ding{175}}
\DeclareMathOperator*{\argmax}{arg\,max}

\title{\problyTitle: Uncertainty-Aware Machine Learning}

\author{%
  \textbf{Paul Hofman\textsuperscript{\,1,2,}\thanks{Equal Contribution}} \quad
  \textbf{Clemens Damke\textsuperscript{\,1,2,}\footnotemark[1]} \quad
  \textbf{Timo L\"ohr\textsuperscript{\,1,2,}\footnotemark[1]} \\
  \textbf{Santo M.\ A.\ R.\ Thies\textsuperscript{\,1,2}} \quad
  \textbf{Alireza Javanmardi\textsuperscript{\,1,2}} \quad
  \textbf{Valentin Margraf\textsuperscript{\,1,2}} \\
  \textbf{Jakub Paplham\textsuperscript{\,4}} \quad
  \textbf{Yusuf Sale\textsuperscript{\,1,2}} \quad
  \textbf{Eyke H\"ullermeier\textsuperscript{\,1,2,3}} \\
  \textbf{Maximilian Muschalik\textsuperscript{\,1,2}} \\[4pt]
  \textsuperscript{\textbf{1}}LMU Munich \quad
  \textsuperscript{\textbf{2}}Munich Center for Machine Learning (MCML) \\
  \textsuperscript{\textbf{3}}German Research Center for Artificial Intelligence (DFKI) \\
  \textsuperscript{\textbf{4}}Czech Technical University in Prague \\
}

\SetSymbolFont{stmry}{bold}{U}{stmry}{m}{n}
\makeatletter
\DeclareRobustCommand\showhyphens[1]{\setbox0\vbox{\color@begingroup\everypar{}\parfillskip\z@skip\hsize\maxdimen\normalfont\pretolerance\m@ne\tolerance\m@ne\hbadness\z@\showboxdepth\z@\ #1\color@endgroup}}
\makeatother

\begin{document}

\maketitle

\begin{abstract}
As machine learning systems are increasingly deployed in real-world applications, the question of how to represent and quantify uncertainty has moved from a methodological side issue to a central concern. In practice, however, making a model uncertainty-aware is still surprisingly difficult: relevant tools tend to be scattered across libraries, each tied to a particular framework and a particular approach to modeling uncertainty, and the choice of representation, quantification, and evaluation is typically left to the user without much guidance. In this paper, we present \textbf{\probly}, a Python package that addresses these issues in a single, modular framework. \textbf{\probly} offers (i) lightweight transformations that turn existing models into uncertainty-aware ones, currently supporting PyTorch, scikit-learn, and Flax, (ii) several representations, including second-order distributions, credal sets, and conformal prediction, (iii) the corresponding quantification measures, and (iv) a number of evaluation protocols, which can be combined more or less arbitrarily. Using \textbf{\probly}, we conduct a benchmark study on three standard tasks --- selective prediction, out-of-distribution detection, and active learning. In addition, and unlike existing benchmarks, we evaluate suitable methods on first-order data, i.e., datasets for which the target itself is a distribution over outcomes. We further illustrate the flexibility of the package on a number of less standard case studies, including large language models, graph neural networks, and data streams. The package is publicly available at \githubrepo.
\end{abstract}
\acbarrier

\section{Introduction}

\looseness=-1 The rise of machine learning, and its growing impact across the sciences and (safety-critical) domains, has revived an old problem in a new guise: how can we \emph{trust} the predictions of a machine learning system? The past decades have seen unprecedented advances in predictive power, notably in deep neural networks, yet their reliability remains an open issue. Accordingly, a growing literature addresses how (predictive) uncertainty should be \emph{represented}, \emph{quantified}, and importantly \emph{evaluated}. Despite these methodological advancements, the use of uncertainty quantification in practice remains limited.

A core obstacle is the lack of a unified evaluation structure for uncertainty methods stemming from the following problems. First, implementations are scattered across research repositories with inconsistent interfaces, undocumented hyperparameter choices, and questionable reproducibility. Existing libraries that do consolidate methods, such as \texttt{torch-uncertainty}, \texttt{lightning-uq-box}, and \texttt{untangle} \citep{lafageTorchUncertainty2025,lehmannLightningUQBox2024,mucsanyiBenchmarkingUncertainty2024}, target a single framework, leaving practitioners with no solution that works across their existing tools and pipelines. Second, there is no standardized evaluation protocol. Methods targeting the same goal are benchmarked on different tasks, datasets, and metrics, making a direct comparison impossible. Third, without a \textit{shared} evaluation structure, it is impossible to know whether a method actually outperforms another for a given use case, or whether any method meets the requirements needed in a given practical application.

To close this gap, we introduce \probly: a Python package and benchmark suite providing library-agnostic, modular implementations of uncertainty methods across all major formalisms (e.g., first-order distributions, second-order distributions, credal sets, and conformal prediction sets). Any model, regardless of the underlying ML framework, can be transformed into an uncertainty-aware predictor and evaluated under a unified protocol across tasks such as selective prediction, out-of-distribution detection, and active learning. \probly is designed to be extensible: researchers can contribute new methods and evaluation protocols, with the goal of establishing a shared standard for uncertainty evaluation, acknowledging that further evaluation schemes will be needed as the field matures. The flexibility of \probly further enables uncertainty quantification in non-standard settings such as data streams, graph neural networks, and large language models, lowering the barrier for practitioners and researchers to apply principled uncertainty methods beyond standard classification tasks.

\probly consolidates uncertainty quantification in machine learning with the following three contributions. \onecirc\ \textbf{A unified uncertainty library}: \probly transforms models from any ML framework into uncertainty-aware predictors through a composable, extensible interface, hence, enabling practitioners to add uncertainty quantification to their existing pipelines. \twocirc\ \textbf{Benchmark evaluation across uncertainty methods}: \probly provides reproducible implementations of methods from published papers across all major uncertainty paradigms and evaluates them under common protocols, opening the door to new evaluation paradigms. \threecirc\ \textbf{Uncertainty quantification beyond standard domains}: we demonstrate the ease-of-use of \probly on a range of settings where principled uncertainty quantification has been largely unexplored, including large language models, graph-structured data, and data streams.

\section{Uncertainty in Machine Learning}

\begin{wrapfigure}{l}{0.4\textwidth}
  \centering
  \vspace{-1.2em}
  \includegraphics[width=0.4\textwidth]{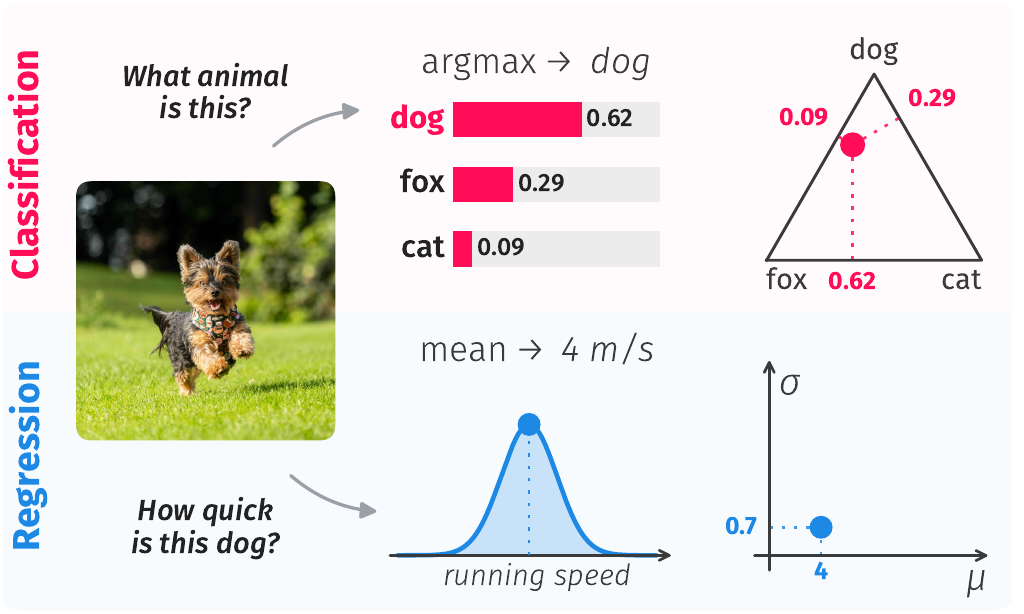}
  \vspace{-1em}
  \caption{Machine learning relies on \textit{probabilistic} models.}
  \label{fig:explainer}
  \vspace{-1em}
\end{wrapfigure}

A standard way to capture uncertainty when predicting the outcome $y \in \mathcal{Y}$ associated with a query instance $\vec{x} \in \mathcal{X}$ is in terms of \textbf{probability distributions} on the outcome space $\cY$ (cf.\ Figure \ref{fig:explainer}). To this end, one trains predictive models (hypotheses) $h: \cX \fromto \mathbb{P}(\cY)$ that produce predictions in the form of distributions $\prob(\cdot \given \vec{x}, h) = h(\vec{x})$ approximating the (assumed) true conditional probability $\prob( \cdot \given \vec{x})$. In classification, this comes down to predicting a (conditional) probability $\prob(y \given \vec{x} , h)$ for each class $y \in \mathcal{Y}$, whereas in regression, $\prob( \cdot \given \vec{x}, h)$ is a density function on $\mathbb{R}$. Probabilistic predictors of that kind can be trained by minimizing suitable loss functions, so-called \emph{proper scoring rules} \citep{gnei_sp05}.

In practice, probabilistic predictions tend to be biased and are often less accurate than they should be in theory. The process of improving predicted probabilities is commonly referred to as \textbf{calibration}. For example, in \emph{post-hoc} calibration, a calibration function is trained that maps inputs in the form of empirical probability estimates to outputs in the form of corrected probability estimates that are hopefully closer to the ground truth \citep{silv_cc23}. 

Going beyond the induction of a single probabilistic predictor, the \textbf{Bayesian approach} proceeds from a prior on a hypothesis space $\mathcal{H}$ of probabilistic predictors, and learning essentially consists of updating the prior with the posterior probability. Uncertainty about a prediction is represented by the \emph{predictive posterior distribution}, which is obtained via model averaging. Gaussian processes \citep{seeg_gp04} realize Bayesian inference in a non-parametric way and are widely used in machine learning.
A robust version of Bayesian inference is based on the concept of \textbf{credal sets}, i.e., convex sets of probability distributions \citep{wall_sr}. The basic idea is to replace a single prior distribution by a (credal) set of candidate priors. Bayesian inference can then be applied to each candidate prior, leading to a set of posteriors or quantities derived from a posterior.
Methods of that kind have also been used in the context of machine learning \citep{cora_lr08,dera_cs17}.

\begin{figure}[h]
    \centering
    \includegraphics[width=\linewidth]{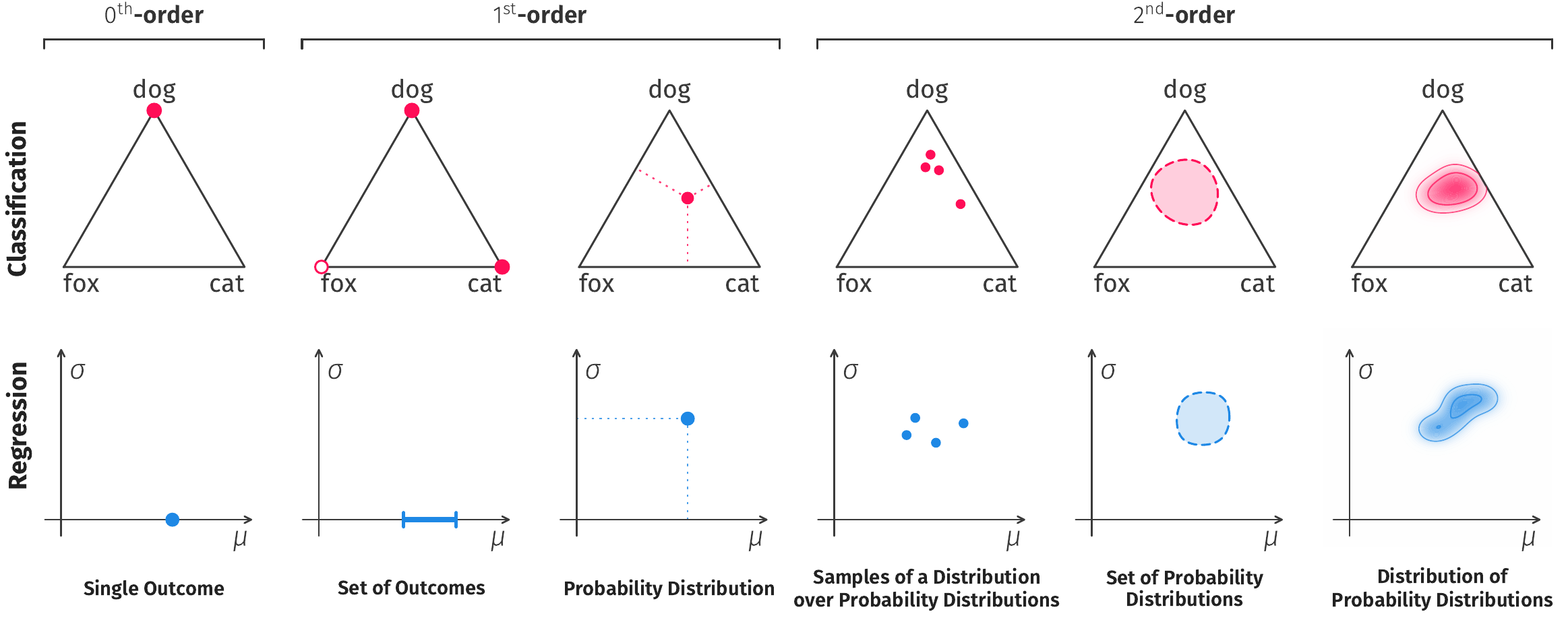}
    \caption{Uncertainty in machine learning can be represented with different formalisms.}
    \label{fig:representations}
\end{figure}

An important feature of the Bayesian (or credal) approach is that it delivers a distribution (or a set) of probabilistic predictors rather than just a single one (cf. \cref{fig:representations}). Thus, the learner's uncertainty about the right model can be represented, and hence its uncertainty about its (uncertain) prediction $h(\vec{x})$ of $\prob(\cdot \given \vec{x})$. This \textbf{epistemic} uncertainty (EU), which is caused by the learner's lack of knowledge, can be distinguished from the \textbf{aleatoric} uncertainty (AU) that is caused by the inherent randomness in the data-generating process\,---\,even if the true distribution $\prob(\cdot \given \vec{x})$ were known, the outcome $y$ could not be predicted with certainty \citep{hullermeier2021aleatoric}. Formally, a \textbf{second-order predictive model} $H: \cX \fromto \mathbb{Q}(\cY)$ maps instances $\vec{x}$ to second-order distributions $(\mathbb{Q}(\cY) = \mathbb{P}(\mathbb{P}(\cY)))$, or, alternatively, to credal sets $(\mathbb{Q}(\cY) = 2^{\mathbb{P}(\cY)})$. 

Training second-order predictors, which are able to represent both aleatoric and epistemic uncertainty, has received increasing attention in machine learning. In addition to Bayesian approaches such as Bayesian neural networks \citep{blundellWeightUncertainty2015} and ensemble-based approximations \cite{laks_sa17}, alternative methods based on other uncertainty calculi have been developed \citep{sens_ed18,wangCredalWrapper2025,loehrCredalPrediction2025,hofmanEfficientCredalPrediction2026}. Going beyond the mere \textbf{representation} of uncertainty in terms of second-order predictions $Q = h(\vec{x})$, these methods typically also offer means for uncertainty \textbf{quantification} (UQ), i.e., numerical measures of the uncertainty associated with a prediction $Q$ that quantify the (total) amount of uncertainty in terms of a single scalar, and maybe disentangle that value into an aleatoric and an epistemic component \cite{hofmanOneSize2026}. Such quantifications can be seen as extensions of measures like entropy or variance for first-order distributions.  

An alternative to representing uncertainty in terms of (first-order) distributions on $\cY$ is in terms of \textbf{set-valued predictions}, i.e., subsets $Y \subseteq \cY$ of candidate outcomes that comprise the true outcome with high probability. \textbf{Conformal prediction (CP)} is a framework for producing reliable predictions of that kind \citep{vovk2022algorithmic}. It can be seen as a meta-learning technique, which can be put on top of any base learner, i.e., any standard classifier producing ``point predictions'' $\hat{y} \in \mathcal{Y}$. It merely requires a measure of (non-)conformity of a (hypothetical) data point, i.e., a measure of how well a combination of query instance $\vec{x}$ and outcome $\hat{y}$ fits with the training data seen so far. While conformal prediction is rooted in classical frequentist statistics and hypothesis testing, Bayesian approaches to set-valued prediction have been proposed as well \citep{mortier21}.

\section{Practical Uncertainty in Machine Learning with \probly}\label{sec:practical}
\providecommand{\processrowsep}{0.6em}

\begin{figure*}[!ht]
\centering
\input{figures/main/process/code_snippets_per_stage}
\noindent\includegraphics[width=\linewidth]{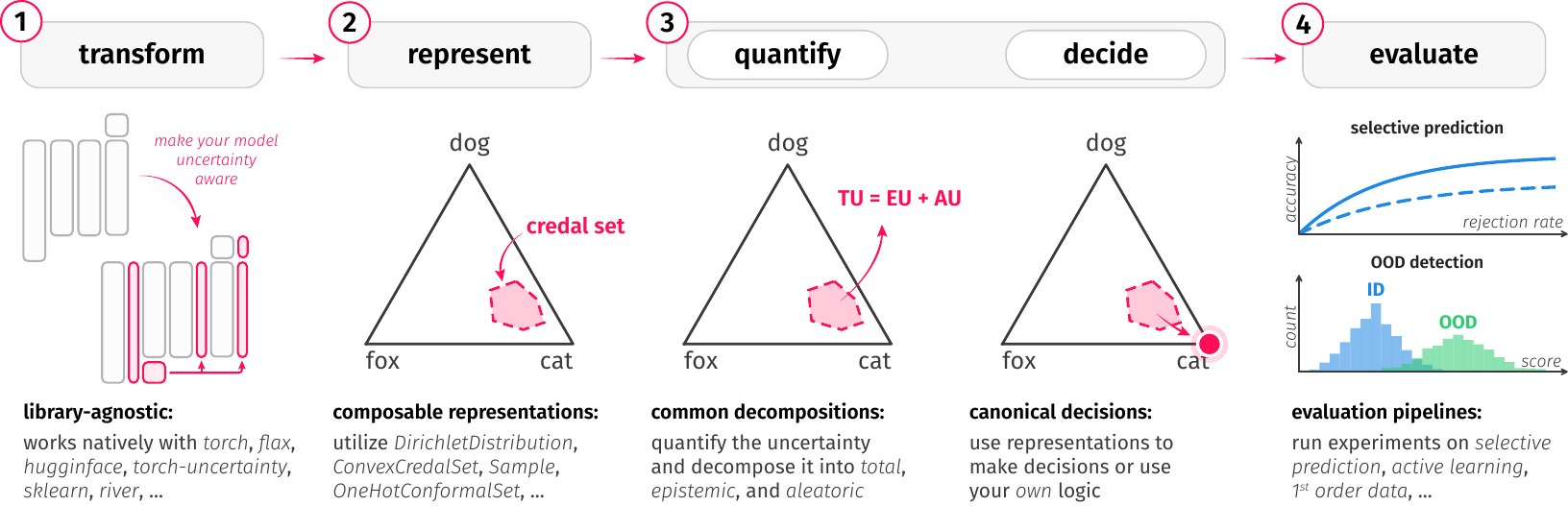}
\caption{\probly enables a composable four-stage workflow.
}
\label{fig:probly-process}
\vspace{-0.5cm}
\end{figure*}

A wide range of uncertainty methods has been developed over the last decades, but their implementations are scattered across separate libraries and code bases. Existing packages often focus on specific ecosystems, modeling paradigms, uncertainty representations, or downstream tasks (see \cref{sec:appx_extended_related_work:libraries}). As a result, researchers and practitioners frequently have to reimplement substantial parts of an uncertainty-aware pipeline, or manually glue together components that were not designed to interact.

We introduce \probly to fill this gap.
Its overarching goal is to accelerate research on uncertainty-aware \ac{ml} while making the resulting methods readily accessible to practitioners. To this end, \probly is built around three core principles:
\onecirc~\textbf{Interoperability.}
    \probly is designed to operate across \ac{ml} ecosystems. Models, data structures, and training pipelines may come from \torch, \jax, \sklearn, or other frameworks.
\twocirc~\textbf{Extensibility.}
    \probly is built to be extended post hoc. New model classes, representations, measures, and decision rules can be integrated via lightweight bindings. This allows \probly to be adapted freely to new research ideas or frameworks by the user.
\threecirc~\textbf{End-to-end coverage.}
    \probly is holistic in scope: it does not focus only on model construction, training, calibration, or evaluation. Instead, it covers the full pipeline, from building an uncertainty-aware model to making decisions based on it.

These principles are reflected in the organization of \probly around a composable four-stage workflow illustrated in \cref{fig:probly-process}. Each stage exposes modular building blocks that can be combined to recover existing methods or to construct new ones. At the same time, \probly ships with a broad range of uncertainty methods and sensible defaults, allowing practitioners to apply uncertainty-aware techniques without having to assemble every component manually.

\subsection{Transforming Models with \probly}

The \probly pipeline starts with \emph{transformations}. A transformation takes an existing model and modifies it to be uncertainty-aware, while leaving the surrounding workflow intact. This is achieved by inserting uncertainty-aware components, such as stochastic layers, duplicating parts of the model to obtain an ensemble, or by wrapping the model to allow for calibration or to add other capabilities. Crucially, transformations are not tied to a particular model class or backend. Internally, \probly provides extensible traversal utilities that can walk through Python containers and model structures, identify compatible components, and replace or augment them in place. This makes the same transformation logic applicable across ecosystems such as \texttt{torch}, \sklearn, or \jax.

Adhering to the principle of interoperability, \probly has no hard dependencies on any \ac{ml} framework; \torch- or \sklearn-specific implementations are loaded as needed based on the model provided by the user. \probly annotates transformed models using virtual predictor type protocols, which advertise the added uncertainty-awareness capabilities to the downstream pipeline stages.

\subsection{Representing Uncertainty with \probly}\label{sec:practical:representation}

After transforming a model, \probly can query it to obtain representations of its predictive uncertainty by means of a \emph{representer}. Representers construct uncertainty representations by sampling stochastic models or post-processing model predictions. \probly automatically selects an appropriate representer for a given model based on the predictor type annotations added during model transformation.

\probly supports first- and second-order uncertainty representations based on sets and distributions (see~\cref{sec:appendix:implemented_methods}). All provided representation types adhere to the Python array API standard~\citep{meurer_python_2023}; this means all representation objects provided by \probly are $n$-dimensional arrays of representation objects. To make these representations library-agnostic, \probly separates abstract representation types from their concrete backend-specific realizations. For instance, a \texttt{CategoricalDistribution} can be realized by an \texttt{ArrayCategoricalDistribution}, a \texttt{TorchCategoricalDistribution}, or an analogous representation for another array backend. Those representations natively integrate with their respective backend; \probly's \texttt{ndarray}-backed representations can be passed directly to NumPy functions, while the \torch-backed representations can be used with regular \torch functions. This lets users manipulate batches of distributions, samples, credal sets, or prediction sets much like ordinary arrays, while preserving the semantic structure needed by later stages of the \probly pipeline.

\subsection{Quantifying Uncertainty and Making Decisions with \probly}

Once a prediction has been represented, \probly supports two complementary operations: \emph{quantification} and \emph{decision-making}. \textbf{Quantification} asks how uncertain the prediction is. Given a representation, \probly can compute scalar uncertainty measures such as entropy-based, variance-based, distance-based, or set-size-based scores. When possible, these scores are further decomposed into uncertainty notions, most importantly, total, aleatoric, and epistemic uncertainty. Thus, \probly does not only quantify how much uncertainty is present, but also where this uncertainty comes from.

\textbf{Decision-making} then asks what to do with the representation and its encoded uncertainty. A \emph{decider} maps an uncertainty representation to a decision space. This may mean selecting a class label, abstaining from prediction, returning a set-valued prediction, or reducing a second-order representation to a first-order predictive distribution. For instance, a second-order distribution or credal set can be collapsed to a categorical distribution by taking its mean, while more conservative decision rules may use worst-case or maximin criteria. Together, quantification and decision-making turn uncertainty representations into numerical scores and concrete actions needed by downstream applications.

\subsection{Evaluating Uncertainty with \probly}
The final stage of the \probly pipeline is \emph{evaluation}. Evaluation closes the loop by assessing whether the uncertainty representations, scores, and decisions produced by the preceding stages are useful for a given downstream task. Rather than treating uncertainty quality as a single abstract property, \probly evaluates uncertainty in context: the appropriate metric depends on whether uncertainty is used for calibration, selective prediction, \ac{ood} detection, active learning, set-valued prediction, or another uncertainty-aware task.

At the level of \textbf{individual predictions}, \probly provides metrics for evaluating first-order predictive distributions, set-valued predictions, and scalar uncertainty scores. These include standard notions such as accuracy, negative log-likelihood, calibration error, prediction-set coverage, set size, and ranking-based measures derived from uncertainty scores.

At the \textbf{task level}, \probly supports evaluation protocols for common uncertainty-aware workflows. In \emph{selective prediction}, uncertainty scores induce an ordering over examples, and performance is measured as coverage is varied. In \emph{\ac{ood} detection}, in-distribution and out-of-distribution examples are compared using ranking metrics such as \ac{auroc} or precision--recall based summaries. In \emph{active learning}, uncertainty measures are evaluated by the learning curves they induce when used as acquisition functions. This task-oriented view makes it possible to compare uncertainty methods not only by the representations they produce, but by their downstream consequences.

\section{Benchmark}\label{sec:benchmark}
We use \probly to perform a benchmark study pursuing three main goals, each going beyond what the current existing comparison in the uncertainty literature offer. First, we provide a structured overview of how different uncertainty methods perform on three of the most common downstream tasks, namely selective prediction, out-of-distribution detection, and active learning. Second, we broaden this comparison beyond what is typically considered, by including both task-specific state-of-the-art approaches, and methods that are usually excluded from comparison of this kind, such as calibration. Third, in addition to the standard evaluation, we compare suitable methods on first-order data --- a setting that has \textit{not yet} been systematically explored in existing benchmarks. 

\subsection{Selective Prediction}\label{sec:bench-sp}

\begin{figure}[tb]
    \centering
    \begin{minipage}[t]{0.49\linewidth}
        \centering
        \includegraphics[height=9.6em]{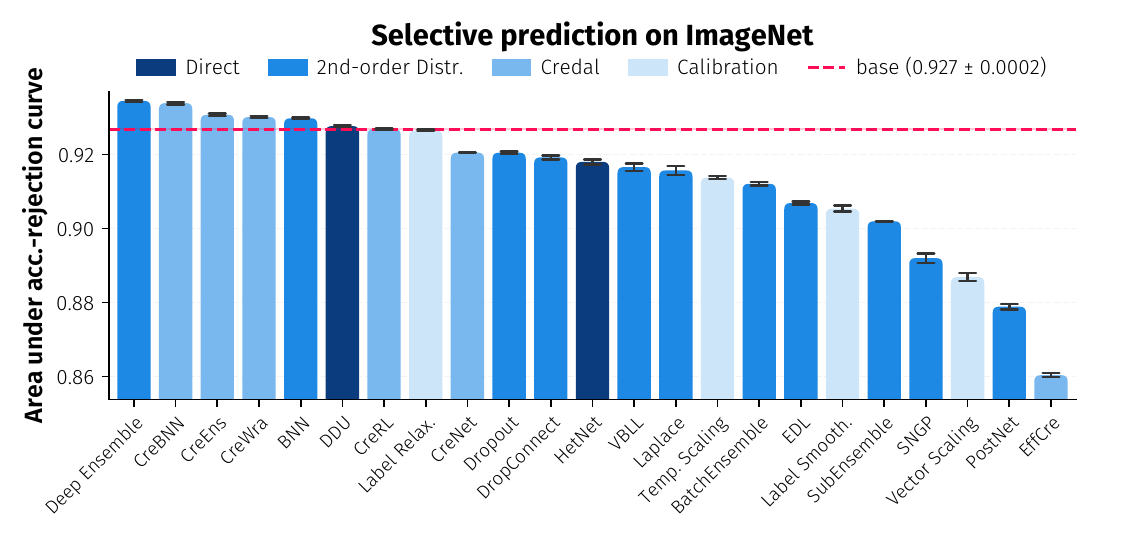}
        \captionof{figure}{Selective prediction on ImageNet. Mean area under the accuracy–rejection curve across three runs. Higher is better.}
        \label{fig:selective-prediction}
    \end{minipage}
    \hfill
    \begin{minipage}[t]{0.49\linewidth}
        \centering
        \includegraphics[height=9.6em]{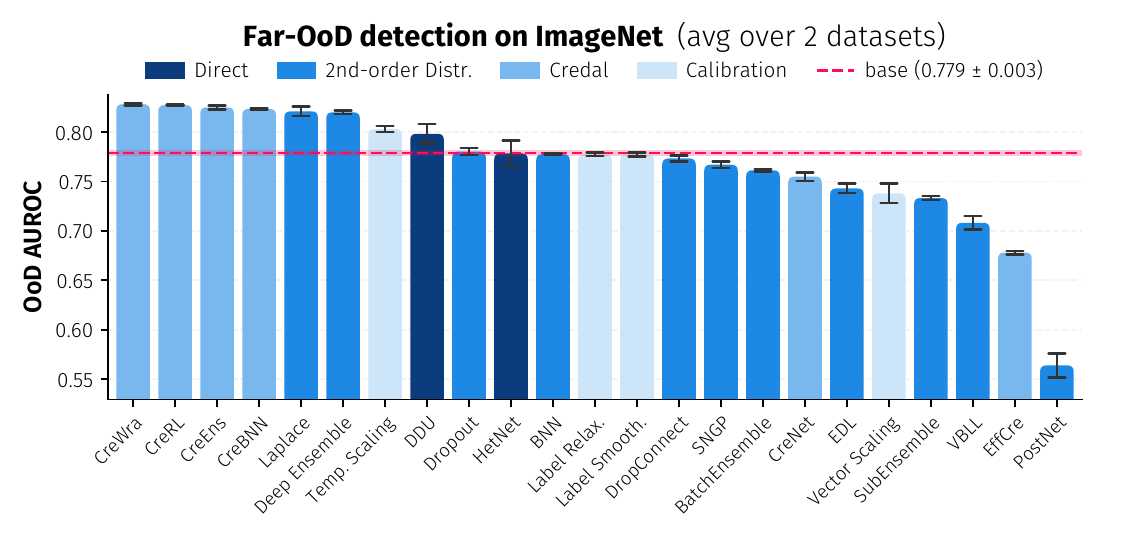}
        \captionof{figure}{Far out-of-distribution detection on ImageNet, averaged over Textures and iNaturalist. Mean \ac{auroc} across three runs. Higher is better.}
        \label{fig:ood-detection}
    \end{minipage}
    \vspace{-0.5cm}
\end{figure}

Selective prediction has become one of the standard tasks for evaluating uncertainty: the predictor is allowed to abstain on a fraction of the (test) instances and is evaluated on the remainder. The objective therefore reduces to identifying instances on which the predictor is likely to be wrong --- a task for which uncertainty is a natural candidate. A predictor equipped with proper uncertainty representation and quantification should attain higher performance on the non-rejected subset, with the performance gain increasing as more rejections are permitted. The standard protocol therefore sweeps the rejection rate over the interval $[0,1)$ and reports the area under the resulting accuracy-rejection curve.

We compare methods on ImageNet \citep{dengImageNet2009} using a ResNet50 \citep{heDeepResidualLearning2016} base model; full implementation and training details are deferred to \cref{app:implementation-details,sec:appendix_data_availability_reproducibility}. In line with recent literature \citep{hofmanOneSize2026}, we use total uncertainty as the rejection criterion. For methods that do not specify a measure of total uncertainty, we use an alternative measure from the original paper. We report the mean area under the accuracy-rejection curve over three runs.

\textbf{Results.} \cref{fig:selective-prediction} shows on the left the results for the selective prediction task. Three findings stand out. First, the single model baseline is highly competitive. Second, the methods that do beat it span both distribution and credal second-order formalisms, suggesting the choice of representation matters less for this task than the construction of the underlying ensemble. Third, the evidential-based methods (PostNet \citep{charpentierPosteriorNetwork2020} and EDL \citep{sens_ed18}) underperform the baseline consistently. 

\subsection{Out-of-Distribution Detection}\label{sec:bench-ood}

\Acl{ood} detection has become a second standard task for evaluating uncertainty, and is most commonly used as a proxy for the quality of epistemic uncertainty. The intuition is straightforward: a predictor's epistemic uncertainty should reflect its (lack of) knowledge, and an instance drawn from a distribution it has never seen is the prototypical case of ignorance. The setting introduces the predictor to such \acf{ood} data alongside \acf{id} ones, and proper epistemic uncertainty should be systematically higher on the former than the latter. The standard protocol formalizes this as a binary classification task and reports the \acf{auroc}.

We follow the evaluation of \citep{zhangOpenOOD2024}, using a ResNet50 with ImageNet as the \ac{id} setting. Near-\acs{ood} performance is evaluated on SSB-hard \citep{vazeSSBHard2022} and NINCO \citep{bitterwolfNINCO2023}, and far-\ac{ood} performance on iNaturalist \citep{vanHornINaturalist2018} and Textures \citep{cimpoiTextures2014}. For each method we compute epistemic uncertainty on \ac{id} and \ac{ood} instances and report the mean \ac{auroc} over three runs, averaged separately for the near- and far-\ac{ood} settings. Additional datasets and metrics are reported in \cref{sec:additional-results}.

\textbf{Results.} \cref{fig:ood-detection} shows on the right the results for the far-\ac{ood} setting. In contrast to selective prediction, the single model baseline is now clearly outperformed by CreRL and Deep Ensemble, among other methods. Once again, the top methods span second-order distribution and credal-based formalisms. Evidential models (EDL, PostNet) are again among the worst performers --- adding empirical evidence to recent critiques about their ability to faithfully represent epistemic uncertainty \citep{juergensEpistemicUncertaintyEvidential2024,bengsPitfallsEpistemicUncertainty2022,meinertUnreasonableEffectiveness2023}.

\subsection{Active Learning}\label{sec:bench-al}

\begin{figure}[tb]
    \centering
    \begin{minipage}[c]{0.49\linewidth}
        \centering
        \includegraphics[width=\linewidth]{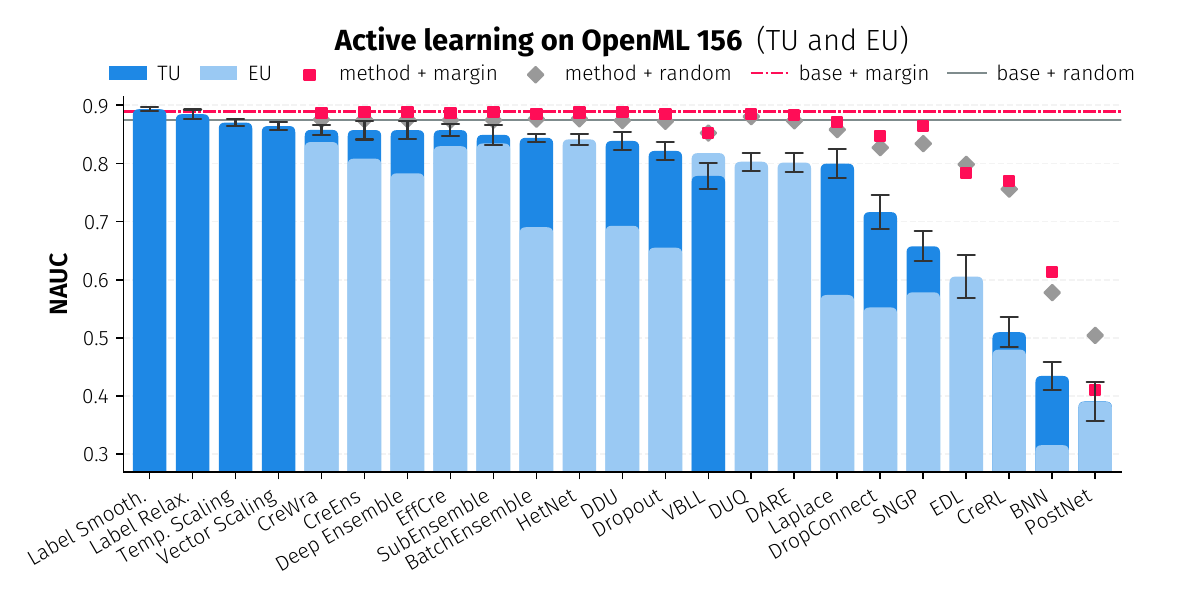}
    \end{minipage}
    \hfill
    \begin{minipage}[c]{0.49\linewidth}
        \centering
        \includegraphics[width=\linewidth]{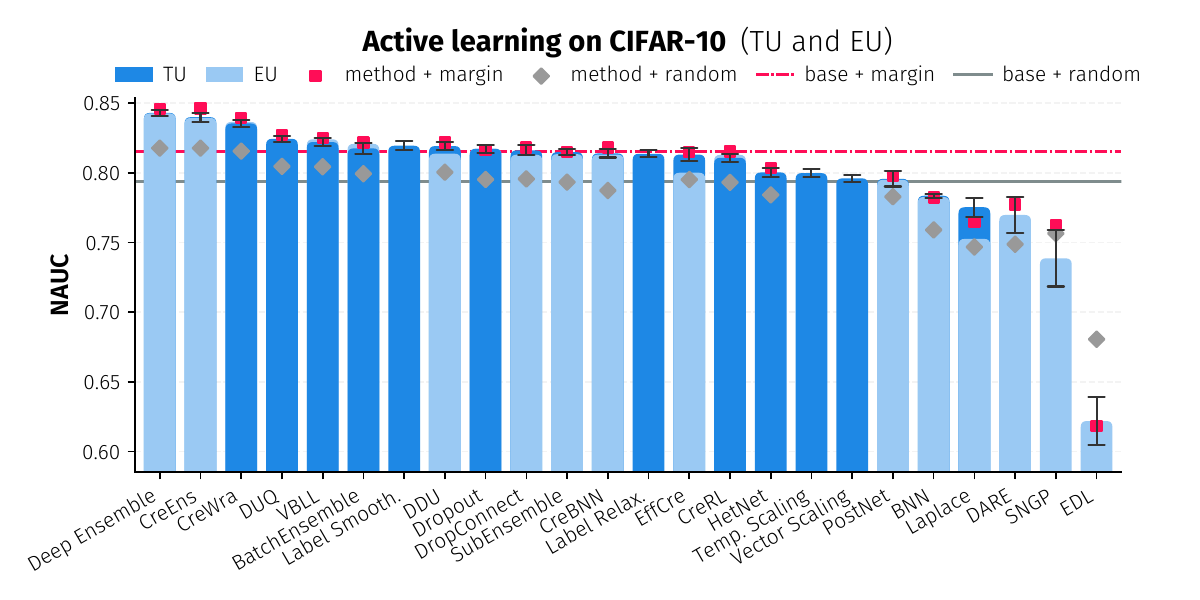}
    \end{minipage}
    \caption{Mean normalized area under the budget curve (NAUC, $\uparrow$) on OpenML-156 \textbf{(left)} and CIFAR-10 \textbf{(right)}.}
\label{fig:al_results}
\vspace{-0.5cm}
\end{figure}

Active learning provides a third standard protocol for uncertainty evaluation, and arguably the one in which epistemic uncertainty plays the most direct role. The aim is to obtain good predictive performance with as little labeled data as possible: starting from a small labeled pool, the learner is allowed at each iteration to query labels for a subset of an unlabeled pool, which is then added to the training set. The natural strategy is to query instances on which the predictor is most uncertain, and epistemic uncertainty is the most principled choice --- by definition, it captures ignorance that more data can in principle reduce \citep{nguyenEpistemicUncertaintySampling2019}. 

\begin{wrapfigure}{r}{0.37\textwidth}
    \vspace{-1em}
    \centering
    \includegraphics[width=\linewidth]{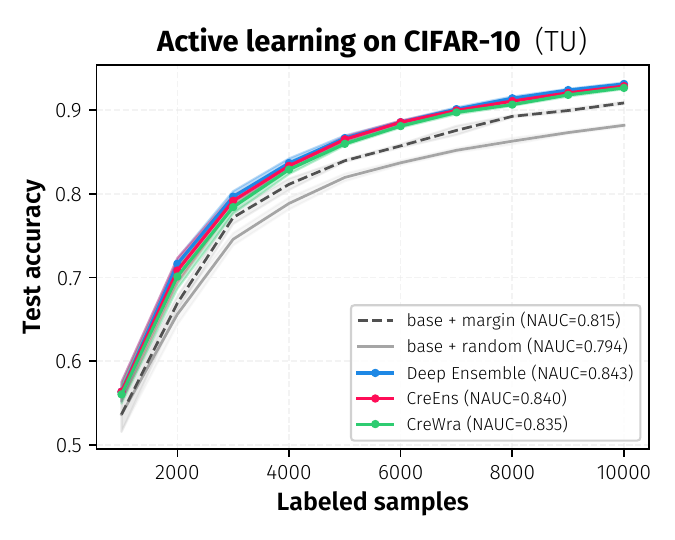}
    \caption{Budget performance plot on CIFAR-10 (see also \cref{fig:al_results:cifar10}) for the five best-performing methods.}
    \label{fig:al_results-cifar10}
\end{wrapfigure}
We run experiments on two tabular datasets from OpenML (IDs 6 and 156), following the protocol of \citep{activelearningSandraECML}. The base model is a two-layer \ac{mlp} with a hidden dimension of 1024. Each of the 10 runs starts from 1000 randomly sampled labeled instances and proceeds for ten acquisition rounds, querying 1000 instances per round. Moreover, we perform experiments on the CIFAR-10 \citep{krizhevsky2009learning} dataset using a ResNet18 \citep{heDeepResidualLearning2016}. Following \citet{yooLearningLoss2019,luethNavigatingPitfalls2023}, we perform 5 runs, starting with 1000 instances and perform nine acquisition rounds, querying 1000 instances per round. To isolate the contribution of uncertainty, we compare each method in three settings: with its native representation and quantification as described in the original paper, with the same base model but random acquisition, and with the same base model and representation under margin- and least-confidence-sampling, which have been found to be among the strongest active learning strategies \citep{bahri2022margin,crossDomainActiveLearningClassification24}. We report the mean normalized area under the budget curve over the runs. Additional configurations and results are reported in \cref{sec:additional-results}.

\textbf{Results.} \cref{fig:al_results,fig:al_results-cifar10,fig:al_results:openml_156,fig:al_results:openml_6,fig:al_results:cifar10} present active learning results. For the tabular datasets, the most striking observation is that no method beats the base model with margin sampling, and --- more surprisingly --- most methods fail to beat the base model with random sampling as well. The strongest performers are the calibration methods, which can only capture aleatoric uncertainty, suggesting its usefulness in active learning. The within-method comparison points to the same conclusion: across uncertainty-aware methods, total uncertainty consistently outperforms epistemic uncertainty, also showing the benefit of adding the aleatoric component. We hypothesize that aleatoric uncertainty is high around decision boundaries, and that querying there provides a useful signal for active learning.

For CIFAR-10, the base model with margin sampling is not the strongest configuration. 
Specifically, the ensemble-based methods --- deep ensemble, credal ensembling, and credal wrapper --- clearly outperform it. 
Yet, their advantage is mainly attributable to ensembling resulting in a stronger predictor, as with random acquisition alone, they already match the base model with margin sampling. 
Notably, however, the gain is not exclusively obtained by ensembling: the DUQ and VBLL methods come close to the performance of the ensembling methods, while relying on a single model. 
As on the tabular datasets, total uncertainty performs at least as well as epistemic uncertainty, and margin sampling, which, computed on the (first-order) predictive distribution, is itself a measure of total uncertainty, remains the best performing strategy. 
The gap in performance between total uncertainty and epistemic uncertainty, which is substantial on OpenML 156, is not as defined for CIFAR-10, which may be explained by the lack of aleatoric uncertainty present in CIFAR-10 instances. 
Taken together, these results indicate that the role of the (epistemic) uncertainty in active learning is less clear-cut than the typical motivation suggests. This may be a matter of practice rather than of principle, however, as an acquisition strategy can only be as good as the estimate it relies on.
Furthermore, the results also caution against drawing strong conclusions from a single dataset, or even from a single modality, as the ranking of methods may change considerably.

\subsection{Evaluation with First-Order Data}\label{sec:bench-fo}

\begin{figure}[t]
    \centering
    \begin{minipage}[t]{0.38\textwidth}
        \vspace{0pt}
        \centering
        \includegraphics[width=\linewidth]{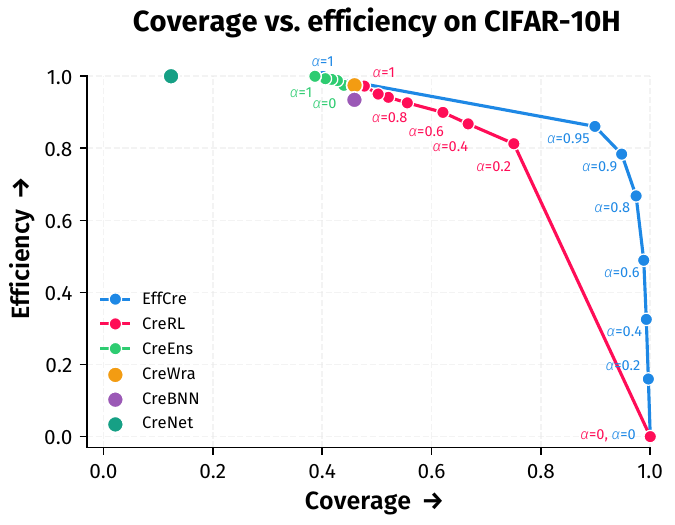}
        \caption{Credal methods trade coverage for efficiency. Higher coverage requires larger credal sets, hence less efficient predictions.}
        \label{fig:cov_eff}
    \end{minipage}
    \hfill
    \begin{minipage}[t]{0.6\textwidth}
        \vspace{0pt}
        \centering
        \captionsetup{position=top}
        \captionof{table}{Conformal prediction ($\alpha=0.1$, 10 seeds) with the ensemble (CreEns) mean as the classifier.}\label{tab:cp}
        \begin{adjustbox}{max width=\linewidth}
        \setlength{\tabcolsep}{3pt}
\renewcommand{\arraystretch}{1.05}
\begin{tabular}{@{}l@{\hspace{2pt}}c c c c c@{}}
\toprule
\multicolumn{2}{@{}l}{\textbf{Metric}} & \textbf{LAC} \cite{sadinle2019least}& \textbf{APS} \cite{romanoAPS2020}& \textbf{RAPS} \cite{angelopoulosRAPS2021} & \textbf{SAPS} \cite{huang2024conformal} \\
\hdashline\noalign{\vskip 0.45em}
\rowcolor[gray]{0.92}[0pt][0pt]
\multicolumn{6}{@{}l@{}}{\textbf{micebone}} \\
\addlinespace[0.15em]
Cond.\ Sat. & $\uparrow$   & $0.742_{\pm 0.046}$          & $\mathbf{0.789_{\pm 0.018}}$ & $0.745_{\pm 0.039}$          & $0.750_{\pm 0.047}$ \\
Set Size    & $\downarrow$ & $\mathbf{1.461_{\pm 0.106}}$ & $1.818_{\pm 0.036}$          & $\mathbf{1.488_{\pm 0.060}}$ & $1.508_{\pm 0.096}$ \\
Marg.\ Cov. & $\uparrow$   & $0.900_{\pm 0.022}$          & $0.899_{\pm 0.015}$          & $0.904_{\pm 0.020}$          & $0.902_{\pm 0.021}$ \\
\addlinespace[0.4em]
\rowcolor[gray]{0.92}[0pt][0pt]
\multicolumn{6}{@{}l@{}}{\textbf{qualitymri}} \\
\addlinespace[0.15em]
Cond.\ Sat. & $\uparrow$   & $0.720_{\pm 0.098}$          & $\mathbf{0.810_{\pm 0.116}}$ & $0.810_{\pm 0.116}$          & $0.720_{\pm 0.098}$ \\
Set Size    & $\downarrow$ & $\mathbf{1.720_{\pm 0.098}}$ & $1.810_{\pm 0.116}$          & $1.810_{\pm 0.116}$          & $1.720_{\pm 0.098}$ \\
Marg.\ Cov. & $\uparrow$   & $0.857_{\pm 0.084}$          & $0.917_{\pm 0.081}$          & $0.917_{\pm 0.081}$          & $0.857_{\pm 0.084}$ \\
\addlinespace[0.4em]
\rowcolor[gray]{0.92}[0pt][0pt]
\multicolumn{6}{@{}l@{}}{\textbf{synthetic}} \\
\addlinespace[0.15em]
Cond.\ Sat. & $\uparrow$   & $0.608_{\pm 0.022}$          & $\mathbf{0.636_{\pm 0.029}}$ & $0.607_{\pm 0.020}$          & $\mathbf{0.638_{\pm 0.019}}$ \\
Set Size    & $\downarrow$ & $\mathbf{1.877_{\pm 0.052}}$ & $1.911_{\pm 0.037}$          & $\mathbf{1.863_{\pm 0.042}}$ & $\mathbf{1.856_{\pm 0.040}}$ \\
Marg.\ Cov. & $\uparrow$   & $0.897_{\pm 0.012}$          & $0.893_{\pm 0.013}$          & $0.897_{\pm 0.012}$          & $0.896_{\pm 0.009}$ \\
\bottomrule
\end{tabular}
        \end{adjustbox}
    \end{minipage}
\end{figure}

In this section, we consider a richer setting, namely \emph{first-order data}, in which each input $\vx \in \cX$ is associated with its conditional label distribution $\prob( \cdot \given \vx)$.
Such data naturally arises in applications where annotations are provided by multiple human experts \cite{peterson2019cifar10h, beyer2020we,nie2020can,armato2011lung}.
In classification settings, the empirical distribution of annotator votes can be aggregated into a categorical distribution, which serves as a reasonable proxy for the underlying conditional label distribution. \cref{tab:first-order-data} in Appendix C summarizes examples of first-order datasets available in \probly. From the perspective of uncertainty quantification, first-order data is particularly valuable, as it directly captures the inherent \emph{aleatoric uncertainty}. This enables principled evaluation of uncertainty estimates, avoiding the need for indirect proxy metrics. 
Note that we adopt the perspective of the learner here, for which the annotation process is simply part of the data-generating process. The uncertainty of the annotators themselves, whether of aleatoric or epistemic nature, is thus not considered further. 

The following two use cases for evaluation highlight the fact that first-order data can provide additional insights going beyond what standard data can do. 

\textbf{Coverage in credal prediction.} 
Given first-order data, the important property of coverage for credal set predictions can be directly assessed in terms of $ \mathds{1}\big( \prob( \cdot \given \vx)\in Q\big)$. Such evaluation has been considered in recent works \cite{javanmardi2024conformalized, caprio2025conformalized, loehrCredalPrediction2025,hofmanEfficientCredalPrediction2026}. \cref{fig:cov_eff} compares the coverage (\cref{eq:cov:hull} for CreEns and CreBNN, \cref{eq:cov:interval} for the interval-based methods) and efficiency \cref{eq:eff} of credal set predictors on CIFAR-10H while \cref{tab:credal_coverage} presents additional results on three other first-order datasets.

\textbf{Conditional coverage in conformal prediction.}
In conformal prediction, \emph{conditional} coverage \cite{foygel2021limits} strengthens the property of marginal coverage and requires $\mathbb{P}\bigl( y \in \cC(\vx) \mid \vx \bigr) \geq 1 - \alpha$ for each individual input $\vx$, where $\cC(\vx) \subseteq \cY$ is the prediction set for $\vec{x}$. In the standard (zero-order) setting, conditional coverage can only be assessed indirectly through proxy metrics (see Section~3.1 in \cite{angelopoulos2023gentle}). In contrast, first-order data enables a more direct evaluation by checking if $\sum_{y \in \mathcal{C}(\mathbf{x})} p(y\mid \mathbf{x}) \geq 1-\alpha$ holds \cite{javanmardi2025optimal}. \cref{tab:cp} compares the performance of 4 well-known conformal prediction approaches for classification. 

\section{Practical Application of \probly}\label{sec:showcase}

In \cref{sec:benchmark}, we evaluated \probly on standard label prediction datasets. To showcase the versatility of \probly we now go beyond this standard setting and explore more specific applications of \acs{uq} in modern \ac{ml}. \onecirc~\textbf{Non-standard targets.} We apply \probly to generative \ac{ml} and quantify uncertainty on \ac{llm} responses. \twocirc~\textbf{Non-standard inputs.} We apply \probly to relational graph data, where the common \acl{iid} assumption does not hold. \threecirc~\textbf{Non-standard learning settings.} We combine \probly with \river for online learning on incremental datastreams. \fourcirc~\textbf{Non-standard methods.} We utilize the modularity of \probly and demonstrate how existing approaches and libraries can be composed, enabling \textit{accessible} future work.

\subsection{Semantic Entropy on LLMs}

Most \acl{uq} approaches assume finite or Euclidean target spaces and thus only work for classification or regression tasks. \Ac{uq} for \acfp{llm} requires special care. \probly provides two state-of-the-art \acs{llm}-\acs{uq} approaches for this: \emph{semantic entropy}~\citep{farquharSemanticEntropy2024} and \emph{spectral entropy}~\citep{walhaSpectralUncertainty2026}. Both obtain \acf{tu} estimates by generating multiple responses for a given prompt and measuring their semantic diversity.

\begin{figure}[t]
\captionsetup[subfigure]{font=footnotesize}
\newlength{\demollmSubfigHeight}
\setlength{\demollmSubfigHeight}{0.11\textheight}
\newlength{\demollmMiddleShift}
\setlength{\demollmMiddleShift}{0.25em}
\centering

\begin{subfigure}[t]{0.41\linewidth}
\centering
\vbox to \demollmSubfigHeight{%
  \hbox to \linewidth{%
    \hfil
    \begin{minipage}{\linewidth}
    \centering
    \begin{tcblisting}{
      listing only,
      listing engine=minted,
      minted language=python,
      minted style=lovelace,
      minted options={
        fontsize=\fontsize{5.8pt}{7pt}\selectfont,
        autogobble,
        fontfamily=\plexttfamily
      },
      colback=codebg,
      colframe=codebg,
      arc=3mm,
      boxrule=0pt,
      enhanced,
      left=2pt,
      right=2pt,
      top=4pt,
      bottom=1pt,
      overlay={
        \problybadge{LLM}
      },
    }
from probly.uncertainty.llm import (
    semantic_entropy, spectral_entropy)
from transformers import AutoModelForCausalLM

# other models like openai, anthropic, ...
llm = AutoModelForCausalLM.from_pretrained(
    "google/gemma-4-E2B-it")
TU = semantic_entropy(llm, questions)
TU = spectral_entropy(llm, questions)
    \end{tcblisting}
    \end{minipage}
    \hfil
  }%
  \vss
}
\caption{Code example for applying LLM uncertainty quantification methods in \probly.}
\label{fig:demollm_code}
\end{subfigure}
\hfill
\begin{subfigure}[t]{0.22\linewidth}
\centering
\vbox to \demollmSubfigHeight{%
  \vspace*{\demollmMiddleShift}
  \hbox to \linewidth{%
    \hfil
    \includegraphics[width=\linewidth]{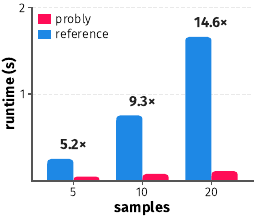}%
    \hfil
  }%
  \vss
}
\caption{Sampling performance of semantic entropy NLI.}
\label{fig:demollm_entropy}
\end{subfigure}
\hfill
\begin{subfigure}[t]{0.33\linewidth}
\centering
\vbox to \demollmSubfigHeight{%
  \hbox to \linewidth{%
    \hfil
    \begin{tikzpicture}
      \node[inner sep=0] (img) {
        \includegraphics[width=\linewidth]{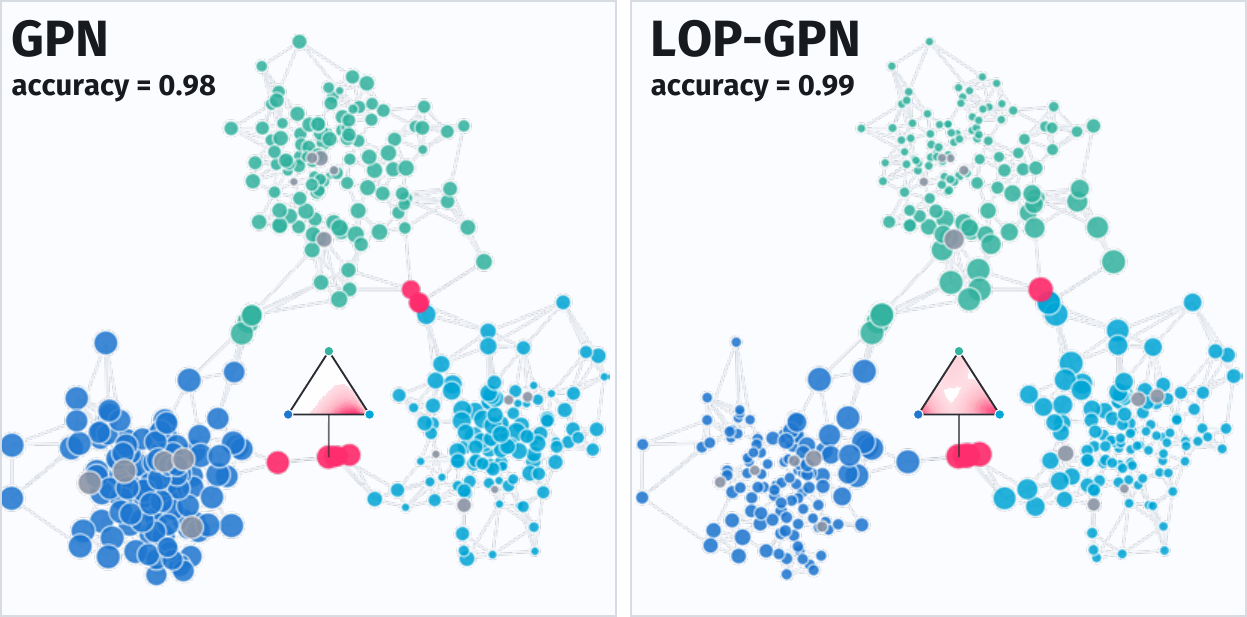}
      };
      \problyimagebadge{GNN}
    \end{tikzpicture}
    \hfil
  }%
  \vss
}
\caption{Node-level uncertainties, vertex size = EU, red = misclass., see~\cref{fig:demogpn_full}.}
\label{fig:demollm_graph}
\end{subfigure}
\caption{Examples of uncertainty quantification workflows supported by \probly.}
\label{fig:demollm}
\end{figure}

\emph{Semantic entropy} computes the entropy over response clusters obtained via a \acf{nli} model. \emph{Spectral entropy} measures uncertainty as the von Neumann entropy of a response embedding kernel. \probly provides efficient batched implementations of those methods for \texttt{huggingface}, see \cref{fig:demollm_entropy}. Additionally, \probly implements \emph{input clarification}~\citep{houInputClarification2024}, a technique for decomposing semantic and spectral entropy-based \ac{tu} estimates into \acs{au} and \acs{eu}. \Cref{fig:demollm_code} demonstrates \probly's \ac{uq} pipeline for a \texttt{huggingface} model.

\subsection{Uncertainty for Graph Machine Learning}
Akin to \acp{llm}, node-level problems on graphs do not satisfy the assumptions of standard \ac{uq} approaches either. Here, instances are vertices of a graph with a relational inductive bias induced by its edges. \Acp{gnn} account for such biases by incorporating graph structure in their predictions. Estimating the uncertainty of a \ac{gnn} thus also requires structure-aware \ac{uq}.

\Acfp{gpn}~\cite{stadler2021} combine \acsp{appnp}~\citep{gasteiger2018} and posterior networks~\citep{charpentierPosteriorNetwork2020}; they propagate predictive uncertainties along edges and directly predict second-order Dirichlet distributions and are thus inherently uncertainty-aware. \Acs{lopgpn}~\citep{damke2024} and \acs{cuqgnn}~\citep{damke2024a} are variants of \ac{gpn} which provide alternative structural uncertainty propagation mechanisms. \probly interoperates with \texttt{torch-geometric}~\citep{fey2019} to provide this family of \ac{gpn} methods. \cref{fig:demollm_graph} illustrates node-level predictive uncertainties of \ac{gpn} and \ac{lopgpn} on a small synthetic dataset with vertex sizes indicating predicted \ac{eu}.
\Ac{lopgpn} assigns high \ac{eu} to vertices that are close to the cluster boundaries, indicating data sparsity in this region of the instance space.
All misclassifications (red vertices) of \ac{lopgpn} occur in these data sparse regions of the graph.
Meanwhile, \ac{gpn} distributes \ac{eu} more evenly among nodes, with most of the bottom left cluster having high \ac{eu}, 
\begin{wrapfigure}{r}{0.33\textwidth}
\centering
\vspace{-0.75em}
\begin{tcblisting}{
  listing only,
  listing engine=minted, minted language=python, minted style=lovelace,
  minted options={fontsize=\fontsize{4.7}{6}, autogobble, fontfamily=\plexttfamily},
  colback=codebg, colframe=codebg,
  arc=3mm, boxrule=0pt, enhanced,
  left=2pt, right=2pt, top=2pt, bottom=1pt,
  overlay={
    \problybadge{river}
  }
}
from river.forest import ARFClassifier
from probly.quantification import quantify
from probly.representer import representer

arf = ARFClassifier(n_models=10, seed=0)
for step, (x, y) in enumerate(stream):
    quantify(representer(arf).represent(x))
    arf.learn_one(x, y) # standard river
\end{tcblisting}
\vspace{-0.4em}
\includegraphics[width=\linewidth]{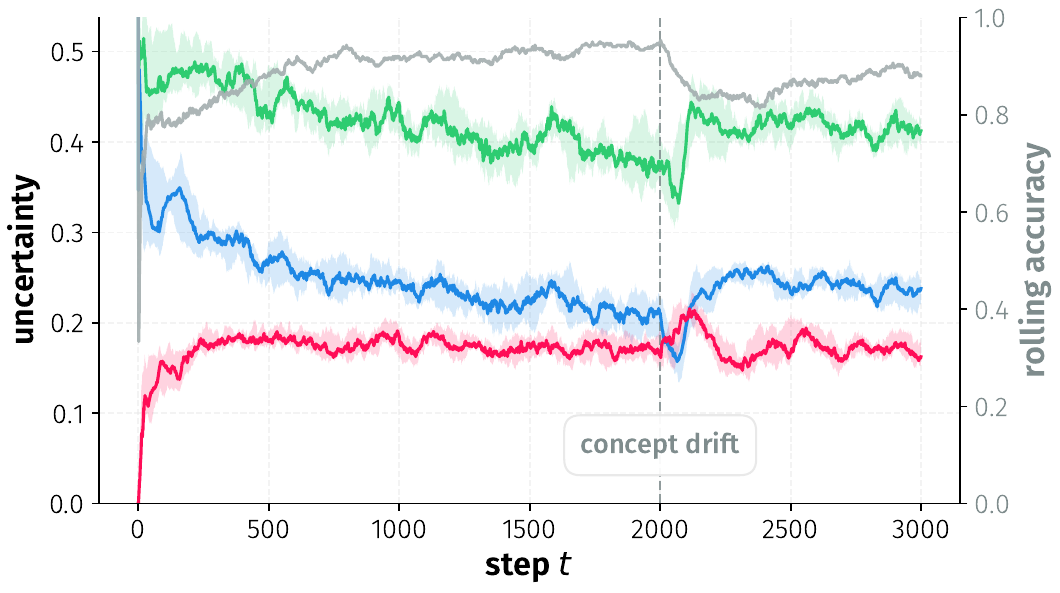}
\caption{\probly on a \river \cite{Montiel.2021} data stream (see~\cref{fig:appendix:demo:data_stream}).}
\label{fig:demo:data_stream}
\vspace{-4em}
\end{wrapfigure}
making it less informative for tasks, such as selective prediction.

\subsection{Using \probly for Online Learning on Data Streams}

Online learning from incremental data streams~\cite{Gama.2014,Lu.2019} departs sharply from the standard i.i.d.\ supervised setting: models observe one instance at a time, are evaluated prequentially, and must adapt continuously as the data distribution shifts --- a phenomenon known as \emph{concept drift}. While uncertainty has been used as a label-free proxy for drift detection~\cite{Lu0L025}, no general-purpose \ac{uq} pipeline has so far been integrated with the stream-learning ecosystem. \probly closes this gap by interoperating with \river, the de facto standard library for online learning, and by directly supporting specialized stream learners such as the well-established \ac{arf}~\cite{Gomes.2017}. At every step of the prequential loop (\cref{fig:demo:data_stream}), \probly returns a \ac{tu}/\ac{au}/\ac{eu} decomposition of the \ac{arf}'s prediction before the model is updated on the new sample.

This unlocks a diagnostic view of streams that goes beyond rolling accuracy~\cite{Lu0L025}. As shown in \cref{fig:demo:data_stream,fig:appendix:demo:data_stream}, different drift regimes leave distinct uncertainty signatures: abrupt virtual drift produces the decomposition's intended separation, with \ac{eu} briefly spiking and then subsiding as the learner accumulates evidence about the new region, while \ac{au}, and consequently \ac{tu}, remains elevated to reflect the irreducible label noise that further training cannot eliminate.

\subsection{Stacking Uncertainty Methods}

\begin{wrapfigure}{r}{0.35\textwidth}
\vspace{-2.3em}
\centering
\begin{tcblisting}{
  listing only,
  listing engine=minted, minted language=python, minted style=lovelace,
  minted options={fontsize=\fontsize{4.5}{6}, autogobble, fontfamily=\plexttfamily},
  colback=codebg, colframe=codebg,
  arc=3mm, boxrule=0pt, enhanced,
  left=2pt, right=2pt, top=2pt, bottom=2pt,
  overlay={
    \problybadge{torch-uncertainty}
  }
}
from probly.transformation import dropout
model = nn.Sequential(
    nn.Linear(4, 16), nn.ReLU(), 
    nn.Linear(16, 3))
do_model = dropout(model, p=0.2)
train(do_model, X_train, y_train)

import torch_uncertainty.post_processing as tup
cp_model = tup.ConformalClsTHR(0.1, do_model)
cp_model.fit(DataLoader(TensorDataset(X, y)))

from probly.representer import representer
prediction_sets = representer(cp_model)(X_test)
prediction_sets.set_size # Tensor of set sizes
\end{tcblisting}
\vspace{-1em}
\caption{Combining \probly with \texttt{torch-uncertainty}~\citep{lafageTorchUncertainty2025}.}
\label{fig:demo:compose}
\vspace{-1em}
\end{wrapfigure}

The previous application examples illustrate that the modular design of \probly makes it straightforward to integrate with a wide variety of existing libraries, even those it was not explicitly designed for. \Cref{fig:demo:compose} highlights this with a roundtrip: starting from a \probly Dropout transformation, passing through a \texttt{torch-uncertainty} conformal wrapper, and ending in a \probly\  \texttt{OneHotConformalSet} representation. Further \probly integrations already exist for \texttt{river} \citep{Montiel.2021}, \texttt{lightning} \citep{pytorch-lightning}, \texttt{PyTorch Geometric} \citep{fey2019}, and even specialized libraries for model releases, such as \texttt{WeatherNext3} \citep{alet2026weathernext,alet2025skillful}.

\section{Implications and Future Work}\label{sec:conclusion}
We contributed \probly, a library for \textit{uncertainty-aware} machine learning, and use it to conduct an extensive benchmark across established evaluation tasks that sheds light on how uncertainty methods perform across \textit{varied} representations. Through case studies, we further showed the composability and extensibility of the package across libraries and settings. This puts in place the tools for further research, and also points to our most natural limitation: the case studies are deliberately illustrative rather than exhaustive, and each warrants a deeper treatment. Looking further ahead, we envision treating the step from uncertainty representations to final predictions more generally through explicit decision rules, which would, for example, allow for joint predictions that combine multiple, potentially orthogonal, representations, a direction we are keen to support within \probly. We hope \probly contributes to developing these directions more holistically, and bringing uncertainty within reach of application areas where it has so far played a peripheral role for practitioners and researchers alike. 

\textbf{Broader Impacts.} This work contributes to the development of reliable machine learning models by improving \ac{uq}. We do not foresee any direct negative broader impacts arising from this.

\begin{ack}
First and foremost, we thank every contributor who has helped shape \probly into what it is today, whether through code, issues, feedback, or discussions that informed its design. We are also grateful to the broader open-source community, whose libraries and tooling this work builds on, and to our colleagues for many valuable conversations along the way. We further want to thank \textit{Elias Barth}, \textit{Tjark Maurice Fromm}, and \textit{Celina Surrer} for their work on the project. 

\textit{Timo Löhr}, \textit{Maximilian Muschalik} and \textit{Eyke Hüllermeier} gratefully acknowledge funding by the Deutsche Forschungsgemeinschaft (DFG, German Research Foundation): TRR 318/3 2026 – 438445824. \textit{Yusuf Sale} is supported by the DAAD program Konrad Zuse Schools of Excellence in Artificial Intelligence, sponsored by the Federal Ministry of Education and Research. \textit{Alireza Javanmardi} was supported by the Klaus Tschira Stiftung (project 00.019.2024). \textit{Jakub Paplhám} acknowledges support from the Czech Science Foundation under project No. 26-22444S. The authors gratefully acknowledge LMU Klinikum for providing computing resources on their Clinical Open Research Engine (CORE).
\end{ack}

\clearpage
\bibliographystyle{abbrvnat}
\bibliography{references}

\appendix
\onecolumn

\begin{center}
  \includegraphics[width=3cm]{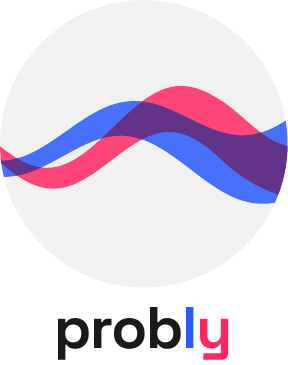}
\end{center}
\vspace{-0.8em}

\section*{Organization}

\startcontents[sections]
\printcontents[sections]{l}{1}{\setcounter{tocdepth}{2}}

\clearpage
\section{Extended Related Work}\label{sec:appx_extended_related_work}
\subsection{Uncertainty Libraries}\label{sec:appx_extended_related_work:libraries}
The growing recognition that uncertainty is essential for trustworthy machine learning has driven the development of a rapidly expanding software ecosystem. Several recent toolkits aim at collecting uncertainty methods for deep learning models under a single interface, including \texttt{Lightning UQ Box} \citep{lehmannLightningUQBox2024}, \texttt{Torch-Uncertainty} \citep{lafageTorchUncertainty2025}, \texttt{Uncertainty Baselines} \citep{nadoUncertaintyBaselines2021}, \texttt{untangle} \citep{mucsanyiBenchmarkingUncertainty2024}, \texttt{Fortuna} \citep{detomassoFortunaLibrary2024}, and the \texttt{UNIQUE} framework \citep{laniniUNIQUEFramework2024}. While each of these libraries gathers a useful collection of methods, they typically commit to a single representation of uncertainty.

This single representation focus is even more pronounced in libraries specialized to particular families of methods. On the second-order distribution and Bayesian side, \texttt{laplace-torch} \citep{daxbergerLaplaceRedux2021} and \texttt{laplax} \citep{weberLaplaxLaplaceApproximations2025} provide post-hoc Laplace approximation in \torch and \jax, respectively, \texttt{posteriors} \citep{duffieldPosteriors2025} targets scalable Bayesian inference, while \texttt{Bayesian layers} \citep{tranBayesianLayers2019}, \texttt{bayesian-torch} \citep{krishnan2022bayesiantorch}, \texttt{BLiTZ} \citep{esposito2020blitzbdl}, and \texttt{torch\_blue} \citep{weyrauchTorchBlue2026} expose drop-in stochastic layers for variational \ac{bnn}s. Conformal prediction has a parallel ecosystem of dedicated tools, including \texttt{TorchCP} \citep{hangTorchCP2025}, \texttt{MAPIE} \citep{cordierMAPIE2023}, and \texttt{crepes} \citep{bostroemCrepes2022}. A further line of work concentrates entirely on \emph{calibration}, packaging methods for post-hoc adjustment of predicted probabilities and metrics such as ECE, as in \texttt{netcal} \citep{kueppersNetcal2020} and \texttt{SmartCal} \citep{abdelrahmanSmartCal2025}. To the best of our knowledge, however, no existing library provides \emph{credal set} representations of uncertainty, and, more importantly, none provides a unified interface that makes the aforementioned representations and tools interchangeable. 

A second issue is more practical: most of these libraries wrap the training stack, forcing the user to express the model inside the library (e.g., a custom \texttt{LightningModule}, or replacing standard layers by custom ones). \probly does not: it may make in-place modification where needed, but does not impose its own training abstractions or rewrite the architecture. This makes \probly composable with existing tools --- a user can train a \texttt{Torch-Uncertainty} model or fit a Laplace approximation with \texttt{laplace-torch} and build on those outputs with \probly.

\subsection{Uncertainty Benchmarks}
In parallel with the libraries reviewed above, a complementary literature has emerged that benchmarks uncertainty methods rather than merely implementing them; indeed, several of the libraries cited above (notably \texttt{Torch-Uncertainty}, \texttt{Lightning UQ Box}, and \texttt{Uncertainty Baselines}) come with benchmark studies. Closest in spirit to our work is the line of evaluation studies focused on uncertainty disentanglement, most prominently \citet{mucsanyiBenchmarkingUncertainty2024} and the related work of \citet{deJongHowDisentangled2024}, which test uncertainty approaches on tasks such as selective prediction and out-of-distribution detection. A complementary line of work benchmarks uncertainty methods under \emph{distribution shift}: \citet{snoekPredictiveUncertaintyShift2019} do so for image classification and \citet{seligmannBayesianDistributionShift2023} extend it to scalable Bayesian deep learning, while OpenOOD \citep{yangOpenOOD2022, zhangOpenOOD2024} focus on out-of-distribution detection. 

Across this body of work, each benchmark commits to a single uncertainty representation. The recent comparative study of \citet{wangSetBasedvsDistributionBased2026} is, to our knowledge, the first to evaluate credal-set-based and second-order-distribution-based representation of uncertainty on the same tasks, but is restricted to selective prediction and out-of-distribution detection. Our benchmark extends this in a number of directions. First, it spans more representations, including direct uncertainty methods and conformal prediction. Second, it includes more tasks, providing active learning in addition to other common tasks. Third, it adds a first-order data comparison, which jointly compares credal coverage and efficiency alongside conformal conditional coverage and conditional calibration. We see this as a first step toward a broader empirical evaluation of uncertainty in machine learning. 

\subsection{Uncertainty Representation}

The most prominent formalism to \emph{represent} (epistemic) uncertainty in machine learning are second-order distributions, i.e., distributions over distributions. From a Bayesian perspective, this second-order object is the posterior over model parameters. Since exact posteriors are intractable for deep neural networks, much of the literature has focused on scalable approximations, such as variational inference \citep{blundellWeightUncertainty2015}, Monte-Carlo dropout \citep{galDropout2016}, Stochastic Weight Averaging-Gaussian \citep{maddoxSWAG2019}, and Laplace approximations restricted to parameter subsets such as the last layer \citep{daxbergerLaplaceRedux2021, kristiadi2020being, harrisonVBLL2024}. Deep ensembles \citep{laks_sa17}, although non-Bayesian in spirit, are typically interpreted as Monte Carlo samples from a posterior and remain the de facto standard baseline for predictive uncertainty, with refinements that compress them via parameter sharing \citep{durasovMasksembles2021, laurentPackedEnsembles2023} or distill them into a single network \citep{malinin2020ensemble}. 

A parallel line of work, mostly motivated by the cost of many forward passes (or full training runs), tries to learn the second-order distribution directly. Rather than averaging over parameter samples, the network outputs a second-order distribution, a Dirichlet over class probabilities, or a Normal-Inverse-Gamma over a Gaussian likelihood, in a single deterministic pass. This family of methods is commonly referred to as evidential deep learning \citep{ulmer2023prior}. Prior networks \citep{malininPriorNetworks2018, malininRegressionPriorNetworks2020} and posterior networks \citep{charpentierPosteriorNetwork2020} parameterize a Dirichlet whose concentration represents the strength of evidence. In this regard, \citep{sens_ed18} reframed this in the language of subjective logic, while \citep{aminiDeepEvidentialRegression2020} extended the idea to regression. Motivated by the same computational burdens, a parallel line of work represents uncertainty using a single deterministic forward pass. Deterministic uncertainty quantification \citep{vanAmersfoortDUE2021} uses RBF-style centroids with a Jacobian penalty, Spectral-normalized Neural Gaussian Processes \citep{liuSNGP2020} couple a bi-Lipschitz feature extractor with a Gaussian-process output layer, and Deep Deterministic Uncertainty \citep{mukhotiDDU2023} shows that a regularized softmax network combined with post-hoc density estimation on its features is comparable to deep ensembles for \ac{ood} detection.

An alternative to second-order distributions, with roots in the imprecise probability literature \citep{wall_sr, augustin2014introduction}, represents (epistemic) uncertainty by a credal set, i.e., a closed convex set of probability distributions. The aim is to learn a credal predictor, a model that, for each input, returns a set of plausible probability distributions over the label space rather than a single one. Many methods to obtain such a credal set in practice have been proposed. These include credal sets obtained as the convex hull of ensemble predictions \citep{wangCredalWrapper2025, nguyenCredalEnsembling2025}, Bayesian credal networks that propagate prior or likelihood imprecision through a Bayesian neural network \citep{caprioCredalBNNs2024}, interval-head networks trained end-to-end to output class-wise plausible intervals \citep{wangCredalIntervalNNs2025}, and likelihood-based approaches that take the credal set to be the predictive image of a relative likelihood $\alpha$-cut around the maximum-likelihood solution \citep{loehrCredalPrediction2025, hofmanEfficientCredalPrediction2026}. 

A separate line of works moves from point predictions to set- (or interval-) valued predictions equipped with finite-sample, distribution-free coverage guarantees. Conformal prediction \citep{vovk2022algorithmic,shafer2008tutorial, angelopoulos2023gentle} turns any heuristic notion of confidence (e.g., softmax scores, residuals, density estimates, etc.) into a prediction region whose marginal coverage probability is provably at least $1 - \alpha$, conditional only on exchangeability of the data. Conformal methods are model-agnostic and can be wrapped around existing predictors without modifying training. Further, we note that in conformal prediction, aleatoric and epistemic uncertainty cannot be represented in a way that clearly separates the two. In this spirit, several recent papers connect conformal prediction with credal sets \citep{javanmardi2024conformalized, javanmardi2025optimal}.

\subsection{Uncertainty Quantification}

Complementary to the representation of uncertainty is its \emph{quantification}, where, given a suitable representation, we seek summaries, or uncertainty measures, that capture the extent of predictive uncertainty. Related to that, is the question of disaggregation, namely, how to decompose (total) uncertainty into its aleatoric and epistemic components.

For second-order distributions, the standard practice, popularized especially in deep learning by \citep{houlsby2011bayesian, kendall2017uncertainties, depewegDecompositionOfUncertainty2018}, decomposes the entropy of the (Bayesian) model average into conditional-entropy (aleatoric), and mutual-information (epistemic). Variance-based variants are also standard in regression \citep{kendall2017uncertainties, aminiDeepEvidentialRegression2020}. While widely used, the information theoretic measures also received criticism. In particular, \cite{wimmer2023quantifying} show that it violates several desiderata one would expect of a principled aleatoric-epistemic split. Several alternatives have been proposed in response, including variance-based measures (for classification) \citep{sale2024label}, distance-based measures \citep{saleDistanceBased2024}, and a unifying scoring-rule framework that exploits the decomposition of any strictly proper scoring rule into a divergence and an entropy component, recovering the entropy chain rule as the log-loss special case \citep{kotelevskii2024risk, fishkov2025uncertainty, hofmanOneSize2026}.

For credal sets, the quantification problem is much older (due to its imprecise-probabilistic foundations). Aleatoric and epistemic uncertainty are also called conflict and non-specificity in the broader uncertainty literature \citep{yager1983entropy, klir1999uncertainty}. The canonical measures are the generalized Hartley measure for non-specificity \citep{abellan2000non}, and the upper Shannon entropy for conflict \citep{abellan2003maximum}. Further, \citet{abellan2006disaggregated} combine these additively into a measure of total uncertainty. Recent work has extended the proper-scoring-rule framework for uncertainty quantification to credal sets \citep{hofman2024quantifying}. Complementarily, \citet{chau2026integral} introduced integral imprecise probability metrics, a Choquet-based generalization that yields tractable measures for credal sets. Similarly to the case of second-order distributions, also a distance-based approach was proposed for credal sets \citep{gonzalez2026quantification}.

\clearpage
\section{Technical Overview of \probly}
A central design principle of \probly is framework-agnostic interoperability. The package is designed to work out of the box with heterogeneous ecosystems such as \torch, \flax, and \sklearn, despite their substantially different programming models and user interfaces. This flexibility is enabled by a set of dedicated infrastructure libraries, most notably \texttt{pytraverse}, \texttt{flexdispatch} and \texttt{ndlike}, which provide the abstractions required to integrate arbitrarily complex framework-specific objects without imposing a common frontend on the user.

We consider the setting in which a user provides a specialized neural network implemented in an arbitrary framework. Such models typically consist of interconnected components, such as layers or modules, that define a computation from inputs to outputs and may be represented by deeply nested Python objects. In \probly, the workflow begins by transforming this object into an uncertainty-aware model. Irrespective of the specific transformation being applied, this requires identifying and visiting the relevant components of the underlying Python structure. \texttt{pytraverse} provides the corresponding abstraction: a flexible mechanism for traversing and manipulating arbitrary nested Python objects in a framework-independent manner.

\textbf{\texttt{pytraverse}.}
The goal of \texttt{pytraverse} is to provide a highly flexible interface for traversing arbitrary Python objects while supporting specialized element-wise transformations. It follows a functional design based on higher-order functions. At a high level, \texttt{pytraverse} applies user-defined functions recursively to nested data structures while maintaining state across the recursive call stack. Its central primitive, \texttt{traverse}, takes an object and a traversal function and applies the latter to the former while propagating auxiliary state variables. The expressivity of this mechanism is determined by the traversal function, which specifies how to iterate over an object's elements and, optionally, how to modify them. Since traversal functions can themselves be composed, complex object transformations can be constructed from simpler traversal rules. Moreover, global or call-stack-level state can be used to carry metadata throughout the traversal, enabling context-dependent transformations of deeply nested Python objects.

Consider a simple framework in which layers are connected sequentially via a linked list. In \texttt{pytraverse}, the traversal of such objects can be defined by an element-wise traversal function that is applied to the current node and then recursively applied to the subsequent layer. Once this traversal rule has been specified, downstream transformations only need to define their behavior on individual layers; they no longer need to encode how those layers are accessed within the network. This separation makes it straightforward to implement custom transformations, such as inserting dropout layers into the linked network or replacing each deterministic layer with a Bayesian counterpart.

While this mechanism is powerful on its own, it does not by itself address a central design goal of \probly: \textit{maintaining lightweight dependencies while retaining high flexibility}. A naive solution would be to define a dedicated interface for each supported framework. However, this is undesirable from a software-engineering perspective, as it imposes unnecessary work on users and leads to poor maintainability. Instead, \probly aims to expose a single user-facing entry point. Supporting multiple frameworks through such an entry point is non-trivial: if framework-specific functionality is imported directly at the entry point, then every user must have all supported frameworks installed and importable at runtime. This is undesirable, since a \torch user, for example, should not be required to inflate their environment with unrelated dependencies, which may be large, unnecessary, or even incompatible with existing packages.

This issue arises from Python's eager evaluation semantics: import statements are executed when they are encountered, rather than when the corresponding functionality is actually required. This contrasts with the principle of \textit{lazy evaluation}, where an expression is represented without being evaluated until its value is needed. In the context of \probly, the analogous requirement is to defer framework-specific imports until the user invokes functionality that depends on them. This makes it possible to maintain a lightweight dependency stack, since users do not need to install every framework supported by \probly, and it avoids potential conflicts between unrelated framework dependencies.

Beyond lightweight dependencies, \probly also requires maintainable, extensible code. Consider again the user who applies a Bayesian transformation to an existing model. Such a transformation may require replacing many different layer types with corresponding Bayesian variants, which naturally induces branching over framework-specific and layer-specific cases. Implementing this logic as a single monolithic function would quickly become difficult to maintain as the number of supported cases grows. Moreover, this branching is not limited to layer transformations: several stages of the \probly pipeline require dispatching over multiple components, including representation, decision, and transformation steps.

A functional remedy is \textit{single dispatch}, where the implementation applied to an input is selected based on the input's type. In this setting, specialized functions are registered for specific types, and the dispatcher acts as a lookup table that maps types to their corresponding implementations. Conceptually, this moves branching logic from the body of a monolithic function to the level of function registration, resulting in a substantially simpler and more modular code base.

While single dispatch improves maintainability, \probly additionally requires lazy evaluation of framework-specific functionality, which is not supported by Python's native dispatcher. This is precisely the role of \texttt{flexdispatch}: it provides lazy, input-type-based dispatching. As a result, \probly obtains the modularity benefits of single dispatch while preserving lightweight dependencies through deferred imports and evaluation.

\paragraph{\texttt{flexdispatch}.}
We developed \texttt{flexdispatch} as an extension of Python's native single-dispatch mechanism by supporting lazy evaluation of registered types. It achieves this through efficient lookup tables that map string-based type representations to specialized implementations. Instead of dispatching directly on concrete Python types, \texttt{flexdispatch} represents types by string expressions and performs type matching at the string level. This enables dispatch rules to be registered without importing the corresponding framework-specific classes at registration time.

In \probly, each user-facing entry point is implemented as a \texttt{flexdispatch} dispatcher, which can be extended to arbitrary frameworks through additional registrations. When the dispatcher is called, it searches the string registry for the best-matching type and applies the corresponding implementation to the input. Since registrations are immutable once defined, lookup results can be cached safely, reducing function-matching overhead and making dispatch costs negligible in typical use.

While the interplay between these two abstractions appears throughout the library, it is most apparent in model transformations. The user-facing entry points are implemented as \texttt{flexdispatch} dispatchers, which route the input object to the appropriate framework-specific implementation. Within that implementation, uncertainty transformations are typically applied using \texttt{pytraverse}: the model is traversed, relevant elements are identified and manipulated, and the resulting object is returned as an uncertainty-aware model.

\paragraph{\texttt{RegistryMeta}.}
When applying transformations to models, \probly annotates those transformed models with the transformation type that was applied to it, such that downstream representers can build an appropriate uncertainty representation out of them.
One way to realize this requirement is to introduce wrapper types around transformed models; the main downside of this approach is that transformations that only change the internals of the model also have to change its outer type and interface.
For example, when applying the \texttt{dropout} transformation to a Torch Lightning model with a custom train method, those custom methods should ideally be preserved.
Another problem with the wrapper-based approach is that a variant of every wrapper type would have to be implemented for every supported framework in order to support their full feature set.

\probly avoids the issues of custom wrapper types by introducing the notion of ``virtual'' \texttt{RegistryMeta} types.
\texttt{RegistryMeta} is a custom Python metaclass, which allows arbitrary objects to be registered as instances of classes with the \texttt{RegistryMeta} metaclass without changing the registered object's original type.
This mechanism can be seen as a natural extension of Python's builtin \texttt{ABCMeta} metaclass, which allows arbitrary classes to be dynamically added as subclasses of the \texttt{ABCMeta} class.
\texttt{RegistryMeta}-based instance registrations are deeply integrated into Python's type system, supporting \texttt{isinstance} checks, single-dispatch for dynamically registered objects and even dynamic registration-preserving serialization support via \texttt{pickle}.
The dynamic dispatching and typing facilities used by \probly are exposed by the \texttt{flextype}\footnote{\url{https://github.com/Cortys/flextype}} sibling project.

\paragraph{\texttt{ndlike}.}
\probly provides batched representation types that behave like $n$-dimensional arrays of uncertainty representations.
As explained in \cref{sec:practical:representation}, these representation types conform to Python's array API standard~\citep{meurer_python_2023} and interoperate with the array functions provided by the supported array backends (NumPy, \torch and \jax).
The custom $n$-dimensional array types are defined using the \texttt{ndlike}\footnote{\url{https://github.com/Cortys/ndlike}} sibling library developed alongside \probly.
At its core, \texttt{ndlike} provides the \texttt{ArrayLike} protocol, defining the shared interface for all $n$-dimensional representation types.
Most representations offered by \probly can be represented by a coordinated set of regular numerical arrays:
an array of univariate Gaussian distributions can be represented by an array of means and an array of variances;
likewise, an array of probability intervals can be represented by an array of upper- and an array of lower-probabilities.
\texttt{ndlike} provides the backend-specific \texttt{AxisProtected} abstraction which wraps such sets of base-arrays and transparently coordinates indexing and shaping operations across the underlying arrays.
Additionally, \texttt{AxisProtected} provides hooks to define custom logic for vectorized NumPy, \torch, or \jax operations, such as addition or multiplication.
\probly uses these mechanisms to support a broad set of arithmetic operations on its representation types;
this includes, both, backend functions like \mintinline{python}{np.mean(categorical_distributions)} and infix operations like \mintinline{python}{2 * gaussian_distributions + 1}.
Last, all of \probly's representation types are regular Python dataclasses; if the coordinated access provided by \texttt{AxisProtected} is too restrictive for a specific use case, the underlying arrays are always accessible as attributes.

\begin{figure}
\centering
\begin{minipage}[c]{0.8\linewidth}
\begin{tcblisting}{
  listing only,
  listing engine=minted, minted language=python, minted style=lovelace,
  minted options={fontsize=\fontsize{6}{6}, autogobble, fontfamily=\plexttfamily},
  colback=codebg, colframe=codebg,
  arc=3mm, boxrule=0pt, enhanced,
  left=2pt, right=2pt, top=4pt, bottom=1pt,
  overlay={
    \problybadge{probly.plot}
  }
}
import matplotlib.pyplot as plt
import numpy as np
from probly.plot import plot_credal_set
from probly.representation.credal_set.array import ArrayProbabilityIntervalsCredalSet

# Binary credal set (K=2) -> rendered on an interval
binary = ArrayProbabilityIntervalsCredalSet(
    lower_bounds=np.array([[0.2, 0.4], [0.5, 0.1]]),
    upper_bounds=np.array([[0.6, 0.8], [0.9, 0.5]]),
)
plot_credal_set(binary, title="Probability Intervals (binary)")

# Simplex credal set (K=3) -> rendered on the ternary simplex
simplex = ArrayProbabilityIntervalsCredalSet(
    lower_bounds=np.array([[0.1, 0.2, 0.3], [0.3, 0.1, 0.1]]),
    upper_bounds=np.array([[0.4, 0.5, 0.6], [0.6, 0.3, 0.7]]),
)
plot_credal_set(simplex, title="Probability Intervals")

# Spider credal set (K>=4) -> rendered as a radar/spider chart
spider = ArrayProbabilityIntervalsCredalSet(
    lower_bounds=np.array([[0.05, 0.02, 0.02, 0.01, 0.01, 0.01, 0.01, 0.01]]),
    upper_bounds=np.array([[0.80, 0.30, 0.20, 0.15, 0.10, 0.10, 0.08, 0.05]]),
)
plot_credal_set(spider, title="Probability Intervals",
                labels=[f"Class {i}" for i in range(8)])
plt.show()
\end{tcblisting}
\end{minipage}
\\
\hfill
\begin{minipage}[c]{0.32\linewidth}
\includegraphics[width=\linewidth]{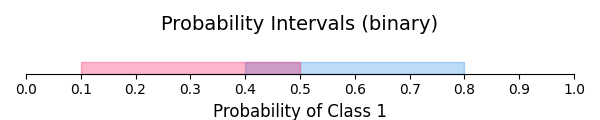}
\end{minipage}
\begin{minipage}[c]{0.32\linewidth}\includegraphics[width=\linewidth]{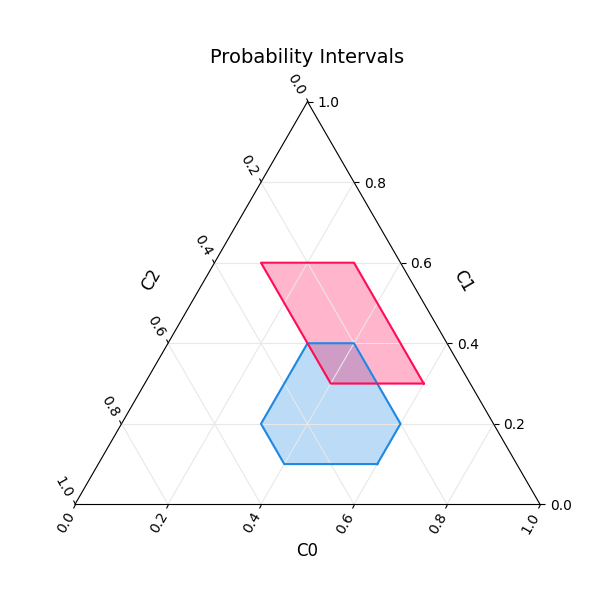}
\end{minipage}
\begin{minipage}[c]{0.32\linewidth}\includegraphics[width=\linewidth]{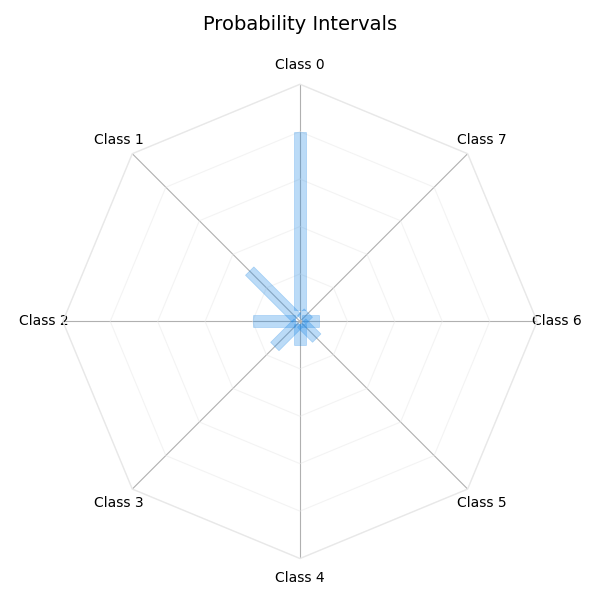}
\end{minipage}
\hfill
\caption{Example visualizations with \texttt{probly}.}
\label{fig:example:visualization}
\end{figure}

\clearpage
\section{Methods Available in \probly}\label{sec:appendix:implemented_methods}
\probly currently supports a substantial portion of the uncertainty methods found in the (recent) literature, spanning second-order distributions, credal sets, conformal prediction, calibration, and direct uncertainty quantification, among others. A complete list is given in \cref{tab:methods}, alongside the corresponding API calls and a comparison with three closely related packages: \texttt{torch-uncertainty}, \texttt{lightning-uq-box}, and \texttt{untangle}. The table omits a number of auxiliary functions that \probly provides, such as intersection probabilities and coverage metrics. 

For transparency, \cref{tab:missing-methods} lists methods that are \emph{not yet} implemented in \probly but are available in one or more of the packages above. The omissions reflect a deliberate prioritization: for this first release, we have focused on breadth across representations, ensuring that each major category of uncertainty method is represented by at least one canonical implementation, and have left greater depth within each category for subsequent releases. We expect most of the methods listed in \cref{tab:missing-methods} to be incorporated in due course, either through our own development or through community contributions. 

\probly is intended as a living code base, developed in close contact with the research it aims to support. The set of methods, the available representations, and the underlying abstractions can all be expected to grow as the field itself matures, and we welcome contributions in any of these directions. To this end, the codebase follows standard open-source conventions: contributions are accepted via pull request, the API is versioned semantically, and breaking changes are documented in a changelog. We refer the reader to the project repository for the current state of the package and for guidance on extending it.

\begin{wraptable}{r}{1.0\textwidth}
  \centering
  \vspace{-1em}
  \footnotesize
  \setlength{\tabcolsep}{3pt}
  \caption{\probly's first-order data.}
  \vspace{-0.7em}
  \begin{tabular}{@{}l@{\hspace{4pt}}l r r@{}}
    \toprule
    \multicolumn{2}{@{}l}{\textbf{Dataset}} & $n_{\text{samples}}$ & $n_{\text{classes}}$ \\
    \midrule
    CIFAR-10H            & \cite{peterson2019cifar10h} & 10\,000 & 10     \\
    ImageNet-ReaL        & \cite{beyer2020we}        & 50\,000 & 1\,000 \\
    \addlinespace[0.25em]
    \rowcolor[gray]{0.92}
    \textit{DCIC benchmark}  & \cite{schmarje2022one} &         &    \\
    \rowarrow Benthic        &      \cite{Langenkamper2020GearStudy, schoening2020Megafauna}                    & 4\,867  & 10  \\
    \rowarrow MiceBone        &   \cite{schmarje2019}                        & 7\,240  & 3  \\
    \rowarrow Pig        &       \cite{schmarje2022one}                    & 10\,237  & 4  \\
    \rowarrow Plankton       &      \cite{schmarje2021foc}                     & 12\,280 & 10 \\
    \rowarrow QualityMRI     &     \cite{obuchowicz2020qualityMRI, stepien2021cnnQuality}                      & 310     & 2  \\
     \rowarrow Synthetic     &     \cite{schmarje2022one}                      & 15\,000     & 6  \\
    \rowarrow Treeversity\#1 &    \cite{arnold_arboretum_data_resources}                           & 9\,489  & 6  \\
    \rowarrow Treeversity\#6 &    \cite{arnold_arboretum_data_resources}                      & 9\,826  & 6  \\
    \rowarrow Turkey &         \cite{volkmann2021turkeys, volkmann2022keypoint}                  & 8\,040  & 3\\
    \bottomrule
  \end{tabular}
  \label{tab:first-order-data}
\end{wraptable}

\begin{table}[t]
  \centering
  \small
  \renewcommand{\arraystretch}{1.15}
  \setlength{\tabcolsep}{4pt}
  \caption{All methods implemented in \probly{} and coverage in three reference libraries:
    \texttt{TU} = \texttt{torch-uncertainty},
    \texttt{LUQ} = \texttt{lightning-uq-box},
    and \texttt{UT} = \texttt{untangle}.}
  \vspace{0.5em}
  \label{tab:methods}
  \resizebox{\linewidth}{!}{%
  \begin{tabular}{@{}l@{\hspace{4pt}}l l c c c c@{}}
    \toprule
    \multicolumn{2}{@{}l}{\textbf{Method}} & \textbf{How to use} & \textbf{\probly} & \textbf{\texttt{TU}} & \textbf{\texttt{LUQ}} & \textbf{\texttt{UT}} \\
    & & & {\scriptsize\textbf{[ours]}} & {\scriptsize\cite{lafageTorchUncertainty2025}} & {\scriptsize\cite{lehmannLightningUQBox2024}} & {\scriptsize\cite{mucsanyiBenchmarkingUncertainty2024}} \\
    \hdashline\noalign{\vskip 0.25em}
    \rowcolor[gray]{0.92}[0pt][0pt]
  \multicolumn{7}{@{}l@{}}{\textbf{Second-order Distributions}} \\
  \addlinespace[0.15em]
  \cite{galDropout2016} & MC Dropout & \codecell{probly.method.dropout} & \cmark & \cmark & \cmark & \cmark \\
  \cite{wanDropConnect2013} & DropConnect & \codecell{probly.method.dropconnect} & \cmark & \xmark & \xmark & \xmark \\
  \cite{blundellWeightUncertainty2015} & Mean-field BNN & \codecell{probly.method.bayesian} & \cmark & \cmark & \cmark & \xmark \\
  \cite{daxbergerLaplaceRedux2021} & Laplace approximation & \codecell{probly.method.laplace} & \cmark & \cmark & \cmark & \cmark \\
  \cite{laks_sa17} & Deep Ensemble & \codecell{probly.method.ensemble} & \cmark & \cmark & \cmark & \cmark \\
  \cite{wenBatchEnsemble2020} & BatchEnsemble & \codecell{probly.method.batchensemble} & \cmark & \cmark & \xmark & \xmark \\
  \cite{valdenegroToroSubEnsemble2023} & SubEnsemble & \codecell{probly.method.subensemble} & \cmark & \xmark & \xmark & \cmark \\
  \cite{deMathelinDARE2023} & DARE & \codecell{probly.method.dare} & \cmark & \xmark & \xmark & \xmark \\
  \cite{liuSNGP2020} & SNGP & \codecell{probly.method.sngp} & \cmark & \xmark & \cmark & \cmark \\
  \cite{sens_ed18} & Evidential classification & \codecell{probly.method.evidential\_classification} & \cmark & \cmark & \xmark & \cmark \\
  \cite{charpentierPosteriorNetwork2020} & Posterior Network & \codecell{probly.method.posterior\_network} & \cmark & \xmark & \xmark & \cmark \\
  \cite{charpentierNaturalPosteriorNetwork2022} & Natural Posterior Network & \codecell{probly.method.natural\_posterior\_network} & \cmark & \xmark & \xmark & \xmark \\
  \cite{malininPriorNetworks2018} & Prior Network & \codecell{probly.method.prior\_network} & \cmark & \xmark & \xmark & \xmark \\
  \cite{aminiDeepEvidentialRegression2020} & Evidential regression & \codecell{probly.method.evidential\_regression} & \cmark & \cmark & \cmark & \xmark \\
  \cite{harrisonVBLL2024} & Variational Bayesian last layer (VBLL) & \codecell{probly.method.vbll} & \cmark & \xmark & \cmark & \xmark \\
  \cite{maddoxSWAG2019} & SWAG & \codecell{probly.method.swag} & \cmark & \cmark & \cmark & \cmark \\
  \cite{durasovMasksembles2021} & Masksembles & \codecell{probly.method.masksembles} & \cmark & \cmark & \cmark & \xmark \\
  \addlinespace[0.4em]
  \rowcolor[gray]{0.92}[0pt][0pt]
  \multicolumn{7}{@{}l@{}}{\textbf{Credal Sets}} \\
  \addlinespace[0.15em]
  \cite{wangCredalWrapper2025} & Credal Wrapper & \codecell{probly.method.credal\_wrapper} & \cmark & \xmark & \xmark & \xmark \\
  \cite{nguyenCredalEnsembling2025} & Credal Ensembling & \codecell{probly.method.credal\_ensembling} & \cmark & \xmark & \xmark & \xmark \\
  \cite{caprioCredalBNNs2024} & Credal BNN & \codecell{probly.method.credal\_bnn} & \cmark & \xmark & \xmark & \xmark \\
  \cite{wangCredalIntervalNNs2025} & Credal Net & \codecell{probly.method.credal\_net} & \cmark & \xmark & \xmark & \xmark \\
  \cite{loehrCredalPrediction2025} & Credal Relative Likelihood & \codecell{probly.method.credal\_relative\_likelihood} & \cmark & \xmark & \xmark & \xmark \\
  \cite{hofmanEfficientCredalPrediction2026} & Efficient Credal Prediction & \codecell{probly.method.efficient\_credal\_prediction} & \cmark & \xmark & \xmark & \xmark \\
  \cite{javanmardi2024conformalized} & Conformal credal set & \codecell{probly.method.conformal\_credal\_set} & \cmark & \xmark & \xmark & \xmark \\
  \cite{javanmardi2024conformalized} & \rowarrow Conformal credal (inner product) & \codecell{probly.method.conformal\_inner\_product} & \cmark & \xmark & \xmark & \xmark \\
  \cite{javanmardi2024conformalized} & \rowarrow Conformal credal (KL) & \codecell{probly.method.conformal\_kullback\_leibler} & \cmark & \xmark & \xmark & \xmark \\
  \cite{javanmardi2024conformalized} & \rowarrow Conformal credal (TV) & \codecell{probly.method.conformal\_total\_variation} & \cmark & \xmark & \xmark & \xmark \\
  \cite{javanmardi2024conformalized} & \rowarrow Conformal credal (Wasserstein) & \codecell{probly.method.conformal\_wasserstein\_distance} & \cmark & \xmark & \xmark & \xmark \\
  \cite{javanmardi2024conformalized} & \rowarrow Conformal credal (Dirichlet RL) & \codecell{probly.method.conformal\_dirichlet\_relative\_likelihood} & \cmark & \xmark & \xmark & \xmark \\
  \addlinespace[0.4em]
  \rowcolor[gray]{0.92}[0pt][0pt]
  \multicolumn{7}{@{}l@{}}{\textbf{Direct Uncertainty Quantification}} \\
  \addlinespace[0.15em]
  \cite{vanAmersfoortDUQ2020} & DUQ & \codecell{probly.method.duq} & \cmark & \xmark & \xmark & \cmark \\
  \cite{mukhotiDDU2023} & DDU & \codecell{probly.method.ddu} & \cmark & \xmark & \xmark & \cmark \\
  \cite{leeMahalanobis2018} & Mahalanobis distance & \codecell{probly.method.mahalanobis} & \cmark & \xmark & \xmark & \cmark \\
  \cite{lahlou2021deup} & DEUP & \codecell{probly.method.deup} & \cmark & \xmark & \xmark & \cmark \\
  \cite{collierHETNet2021} & HetNet (heteroscedastic) & \codecell{probly.method.het\_net} & \cmark & \cmark & \cmark & \cmark \\
  \addlinespace[0.4em]
  \rowcolor[gray]{0.92}[0pt][0pt]
  \multicolumn{7}{@{}l@{}}{\textbf{Conformal Prediction}} \\
  \addlinespace[0.15em]
  \cite{sadinle2019least} & LAC & \codecell{probly.method.conformal.conformal\_lac} & \cmark & \cmark & \xmark & \xmark \\
  \cite{romanoAPS2020} & APS & \codecell{probly.method.conformal.conformal\_aps} & \cmark & \cmark & \xmark & \xmark \\
  \cite{huang2024conformal} & SAPS & \codecell{probly.method.conformal.conformal\_saps} & \cmark & \xmark & \xmark & \xmark \\
  \cite{angelopoulosRAPS2021} & RAPS & \codecell{probly.method.conformal.conformal\_raps} & \cmark & \cmark & \cmark & \xmark \\
  \cite{Vovk1999MachineLearningAO} & Absolute-error CP & \codecell{probly.method.conformal.conformal\_absolute\_error} & \cmark & \xmark & \xmark & \xmark \\
  \cite{Romanoetal2019} & CQR & \codecell{probly.method.conformal.conformal\_cqr} & \cmark & \xmark & \cmark & \xmark \\
  \cite{Romanoetal2019} & CQR-r & \codecell{probly.method.conformal.conformal\_cqr\_r} & \cmark & \xmark & \xmark & \xmark \\
  \cite{rosselliniIntegratingUncertainty2024} & UACQR & \codecell{probly.method.conformal.conformal\_uacqr} & \cmark & \xmark & \xmark & \xmark \\
  \addlinespace[0.4em]
  \rowcolor[gray]{0.92}[0pt][0pt]
  \multicolumn{7}{@{}l@{}}{\textbf{Calibration}} \\
  \addlinespace[0.15em]
  \cite{guoOnCalibration2017} & Temperature scaling & \codecell{probly.method.calibration.temperature\_scaling} & \cmark & \cmark & \cmark & \cmark \\
  \cite{platt1999probabilistic} & Platt scaling & \codecell{probly.method.calibration.platt\_scaling} & \cmark & \xmark & \xmark & \xmark \\
  \cite{guoOnCalibration2017} & Vector scaling & \codecell{probly.method.calibration.vector\_scaling} & \cmark & \cmark & \xmark & \xmark \\
  \cite{chakravartiIsotonicRegression1989} & Isotonic regression & \codecell{probly.method.calibration.isotonic\_regression} & \cmark & \cmark & \xmark & \xmark \\
  \addlinespace[0.4em]
  \rowcolor[gray]{0.92}[0pt][0pt]
  \multicolumn{7}{@{}l@{}}{\textbf{Losses}} \\
  \addlinespace[0.15em]
  \cite{szegedyRethinkingInceptionArchitecture2016} & Label smoothing & \codecell{probly.train.LabelSmoothingLoss} & \cmark & \cmark & \xmark & \xmark \\
  \cite{DBLP:conf/aaai/LienenH21} & Label relaxation & \codecell{probly.train.LabelRelaxationLoss} & \cmark & \xmark & \xmark & \xmark \\
  \cite{linFocalLoss2017} & Focal loss & \codecell{probly.train.FocalLoss} & \cmark & \cmark & \xmark & \xmark \\
  \addlinespace[0.4em]
    \rowcolor[gray]{0.92}[0pt][0pt]
    \multicolumn{7}{@{}l@{}}{\textbf{Uncertainty Quantification}} \\
  \addlinespace[0.15em]
  \cite{depewegDecompositionOfUncertainty2018} & Shannon entropy & \codecell{probly.quantification.measure.distribution} & \cmark & \cmark & \cmark & \cmark \\
  \cite{hullermeier2021aleatoric} & Upper and Lower Entropy & \codecell{probly.quantification.measure.distribution} & \cmark & \xmark & \xmark & \xmark \\
  \cite{sens_ed18} & Variance based & \codecell{probly.quantification.measure.sample} & \cmark & \xmark & \xmark & \xmark \\
    \cite{kendall2017uncertainties} & Dempster Shafer & \codecell{probly.quantification.measure.distribution} & \cmark & \xmark & \xmark & \cmark \\
    \cite{charpentierNaturalPosteriorNetwork2022} & Vacuity & \codecell{probly.quantification.measure.distribution} & \cmark & \xmark & \xmark & \cmark \\
    \cite{farquharSemanticEntropy2024} & Semantic entropy & \codecell{probly.quantification.measure.entropy} & \cmark & \xmark & \xmark & \xmark \\
  \cite{walhaSpectralUncertainty2026} & Spectral entropy & \codecell{probly.quantification.spectral\_entropy} & \cmark & \xmark & \xmark & \xmark \\
  \cite{houInputClarification2024} & Clarification pipeline & \codecell{probly.representer.HFQuestionClarifier} & \cmark & \xmark & \xmark & \xmark \\
    \bottomrule
  \end{tabular}%
  }

\end{table}

\begin{table}[t]
  \centering
  \small
  \renewcommand{\arraystretch}{1.15}
  \setlength{\tabcolsep}{4pt}
  \caption{Methods implemented by reference libraries and \emph{not yet} by \probly.
    \texttt{TU} = \texttt{torch-uncertainty},
    \texttt{LUQ} = \texttt{lightning-uq-box},
    and \texttt{UT} = \texttt{untangle}.}
  \vspace{0.5em}
  \label{tab:missing-methods}
  \begin{tabular}{@{}l@{\hspace{4pt}}l c c c@{}}
    \toprule
    \multicolumn{2}{@{}l}{\textbf{Method}} & \textbf{\texttt{TU}} & \textbf{\texttt{LUQ}} & \textbf{\texttt{UT}} \\
    & & {\scriptsize\cite{lafageTorchUncertainty2025}} & {\scriptsize\cite{lehmannLightningUQBox2024}} & {\scriptsize\cite{mucsanyiBenchmarkingUncertainty2024}} \\
    \hdashline\noalign{\vskip 0.25em}
    \rowcolor[gray]{0.92}[0pt][0pt]
  \multicolumn{5}{@{}l@{}}{\textbf{Second-order Distributions}} \\
  \addlinespace[0.15em]
  \cite{izmailovSWA2018} & SWA & \cmark & \xmark & \xmark \\
  \cite{polyakJuditskyAveraging1992} & EMA (weight averaging) & \cmark & \xmark & \xmark \\
  \cite{wellingSGLD2011} & SGLD & \xmark & \cmark & \xmark \\
  \cite{huangStochasticDepth2016} & Stochastic depth & \cmark & \xmark & \xmark \\
  \cite{teyeMCBatchNorm2018} & MC-BatchNorm & \cmark & \xmark & \xmark \\
  \cite{franchiABNN2024} & ABNN (adaptable BNN) & \cmark & \xmark & \xmark \\
  \cite{laurentPackedEnsembles2023} & Packed-Ensembles & \cmark & \xmark & \xmark \\
  \cite{havasiMIMO2021} & MIMO & \cmark & \xmark & \xmark \\
  \cite{vanAmersfoortDUE2021} & DUE / Deep Kernel Learning & \xmark & \cmark & \xmark \\
  \cite{Park2023DensityUL} & Density-uncertainty layers & \xmark & \cmark & \xmark \\
  \cite{hanCARD2022} & CARD (classification / regression diffusion) & \xmark & \cmark & \xmark \\
  \cite{bishopMixtureDensityNetworks1994} & Mixture density networks & \xmark & \cmark & \xmark \\
  \cite{kohlProbabilisticUNet2018} & Probabilistic U-Net & \xmark & \cmark & \xmark \\
  \cite{durasovZigZag2024} & ZigZag & \xmark & \cmark & \xmark \\
  \cite{ayhanTestTimeAugmentation2018} & Test-time augmentation & \xmark & \cmark & \xmark \\
  \addlinespace[0.4em]
  \rowcolor[gray]{0.92}[0pt][0pt]
  \multicolumn{5}{@{}l@{}}{\textbf{Conformal Prediction}} \\
  \addlinespace[0.15em]
  \cite{angelopoulosImg2ImgConformal2022} & Image-to-image conformal & \xmark & \cmark & \xmark \\
  \addlinespace[0.4em]
  \rowcolor[gray]{0.92}[0pt][0pt]
  \multicolumn{5}{@{}l@{}}{\textbf{Calibration}} \\
  \addlinespace[0.15em]
  \cite{zadroznyHistogramBinning2001} & Histogram binning & \cmark & \xmark & \xmark \\
  \cite{guoOnCalibration2017} & Matrix scaling & \cmark & \xmark & \xmark \\
  \cite{kullDirichletCalibration2019} & Dirichlet calibration & \cmark & \xmark & \xmark \\
  \cite{naeiniBBQ2015} & BBQ (Bayesian binning) & \cmark & \xmark & \xmark \\
  \addlinespace[0.4em]
  \rowcolor[gray]{0.92}[0pt][0pt]
  \multicolumn{5}{@{}l@{}}{\textbf{Losses}} \\
  \addlinespace[0.15em]
  \cite{Fellaji2024OnTC} & Conflictual loss & \cmark & \xmark & \xmark \\
  \cite{pereyraConfidencePenalty2017} & Confidence-penalty loss & \cmark & \xmark & \xmark \\
  \addlinespace[0.4em]
  \rowcolor[gray]{0.92}[0pt][0pt]
  \multicolumn{5}{@{}l@{}}{\textbf{Downstream-Task-Specific}} \\
  \addlinespace[0.15em]
  \cite{liuEnergyScore2020} & Energy score & \cmark & \xmark & \xmark \\
  \cite{hendrycksMaxSoftmax2017} & MaxLogit / MaxSoftmax & \cmark & \xmark & \xmark \\
  \cite{corbiereConfidNet2019} & Correctness prediction head & \xmark & \xmark & \cmark \\
  \cite{yooLearningLoss2019} & Loss prediction head & \xmark & \xmark & \cmark \\
    \bottomrule
  \end{tabular}
\end{table}

\begin{table}[t]
  \centering
  \small
  \renewcommand{\arraystretch}{1.05}
  \setlength{\tabcolsep}{4pt}
  \caption{All uncertainty representations implemented in \probly.}
  \vspace{0.5em}
  \label{tab:representations}
  \begin{tabular}{@{}c l c l@{}}
    \toprule
    \textbf{Order} & \textbf{Representation} & \textbf{Type} & \textbf{Description} \\
    \hdashline\noalign{\vskip 0.2em}
    \rowcolor[gray]{0.92}[0pt][0pt]
  \multicolumn{4}{@{}l@{}}{\textbf{1st-order}} \\
  \addlinespace[0.05em]
  $1^{\mathrm{st}}$ & \codecell{CategoricalDistribution} & \icon{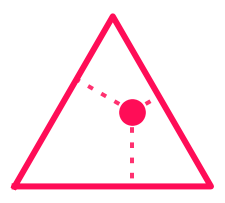} & Probability over a finite label set. \\
  $1^{\mathrm{st}}$ & \codecell{GaussianDistribution} & \icon{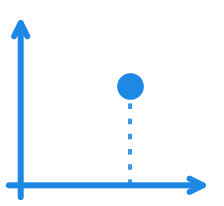} & Continuous, mean and variance. \\
  $1^{\mathrm{st}}$ & \codecell{OneHotConformalSet} & \icon{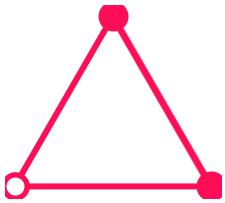} & Set of class labels. \\
  $1^{\mathrm{st}}$ & \codecell{IntervalConformalSet} & \icon{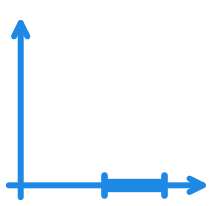} & Real-valued prediction interval. \\
  $1^{\mathrm{st}}$ & \codecell{Embedding} & \icon{reg_distribution} & Vector embedding with a protected trailing axis. \\
  $1^{\mathrm{st}}$ & \codecell{TextGeneration} & \icon{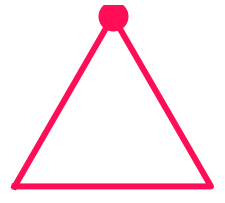} & Decoded text with sequence log-likelihoods. \\
  $1^{\mathrm{st}}$ & \codecell{TokenGeneration} & \icon{cls_label} & Token sequences with transition log-probs. \\
  \addlinespace[0.2em]
  \rowcolor[gray]{0.92}[0pt][0pt]
  \multicolumn{4}{@{}l@{}}{\textbf{2nd-order}} \\
  \addlinespace[0.05em]
  $2^{\mathrm{nd}}$ & \codecell{Sample} & \icon{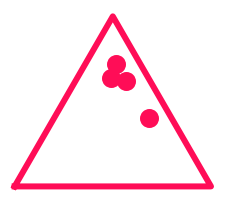} \icon{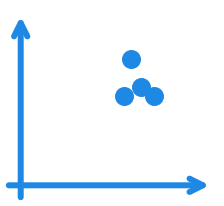} & (Weighted) samples from a 2nd-order object. \\
  $2^{\mathrm{nd}}$ & \codecell{CategoricalCredalSet} & \icon{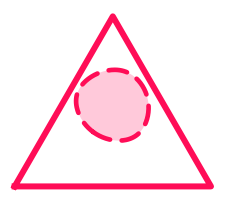} & Set of categorical distributions. \\
  $2^{\mathrm{nd}}$ & \rowarrow \codecell{ConvexCredalSet} & \icon{cls_set_distributions} & Convex hull over finite distributions. \\
  $2^{\mathrm{nd}}$ & \rowarrow \codecell{DistanceBasedCredalSet} & \icon{cls_set_distributions} & Distance ball around a nominal distribution. \\
  $2^{\mathrm{nd}}$ & \rowarrow \codecell{ProbabilityIntervalsCredalSet} & \icon{cls_set_distributions} & Lower / upper bounds on outcome probabilities. \\
  $2^{\mathrm{nd}}$ & \rowarrow \codecell{DiscreteCredalSet} & \icon{cls_set_distributions} & Finite set of extreme distributions. \\
  $2^{\mathrm{nd}}$ & \phantom{\rowarrow} \rowarrow \codecell{SingletonCredalSet} & \icon{cls_set_distributions} & Single precise distribution (degenerate). \\
  $2^{\mathrm{nd}}$ & \codecell{DirichletDistribution} & \icon{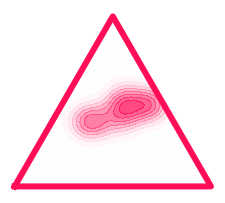} \icon{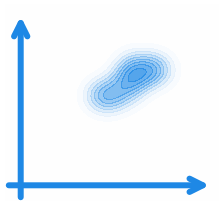} & Distribution over categorical distributions. \\
    \bottomrule
  \end{tabular}
\end{table}

\clearpage

\section{Method Implementation Details}\label{app:implementation-details}
In this section, we discuss all of the implementation-specific details of the methods used for the benchmark. This includes the choice of hyperparameters and assumptions that were made. In general, for each method we used the hyperparameters from the respective paper that introduced the method. Some papers do not include measures of total, aleatoric, and epistemic uncertainty, hence, for these papers we make assumptions about the measure of uncertainty that should be used. These assumptions will be listed clearly for each method.

\subsection{Second-Order Distribution}
\subsubsection{Ensemble}\label{app:implementation-details:ensemble}
The ensemble is constructed by training $M$ copies of the base model with different weight initializations. The randomness in the initialization and the random batches are used to produce $M$ different predictors. Given a trained ensemble, and an instance $\vx$, predictions are made by averaging the probabilistic predictions of the members $\bar{p}(\cdot \mid \vx, h_i) = \frac{1}{M}\sum_{m=1}^M p(\cdot\mid\vx,h_m)$, where $\{h_1, \ldots, h_M\}$. The (total) uncertainty associated with the prediction for $\vx$ can be decomposed into aleatoric and epistemic components using a decomposition based on entropy \citep{depewegDecompositionOfUncertainty2018}:
\begin{equation}\label{eq:entropy-decomposition}
    \operatorname{H}(\bar{p}(\cdot \mid \vx)) = \frac{1}{M}\sum_{m=1}^M \operatorname{H}(p(\cdot \mid \vx, h_m)) + \frac{1}{M}\sum_{m=1}^M\operatorname{KL}(p(\cdot \mid \vx) \parallel \bar{p}(\cdot \mid \vx)),
\end{equation}
where $\operatorname{H}$ denotes the Shannon entropy. In the benchmark, we use $M=10$. When starting with a pretrained network base model, we rely only on the randomness of stochastic gradient descent to create diversity. 
\subsubsection{SubEnsemble}
The subensemble follows a similar logic to the ensemble, but with a focus on efficiency. Instead of training the whole network, the backbone is shared and only the (linear) head is trained to create $M=10$ different members. The predictive distribution and uncertainty is computed in the same way as for the ensemble.

\subsubsection{Batchensemble}
The batchensemble  \citep{wenBatchEnsemble2020} is an efficient approximation of the ensemble that avoids training $M$ independent copies of the base model. Instead, a single shared weight matrix $\mathbf{W}$ is maintained, and each ensemble member $m$ is defined by an element-wise rescaling via rank-one matrices, yielding per-member effective weights $W_m = W \cdot (r_m s_m^\top)$. This construction reduces both memory and compute cost relative to a full ensemble while still producing $M$ diverse predictors. The predictive distribution and uncertainty decomposition follow the same procedure as for the ensemble.
In the benchmark, we use $M=10$.
The shared weights are initialized from the base model; entries of $r_m$ and $s_m$ are initialized from $\mathcal{N}(1,\sigma^2)$, with $\sigma=0.5$ on CIFAR-10 and $\sigma=0.1$ on ImageNet.

\subsubsection{Bayesian Neural Networks}\label{sec:bnns}
\Aclp{bnn} \citep{blundellWeightUncertainty2015} place a prior distribution $p(\vec{\theta})$ over the network weights and seek to compute the posterior $p(\vec{\theta} \mid \mathcal{D})$ after observing data $\mathcal{D}$. Predictions are obtained by marginalizing over the posterior: 
\begin{equation} p(\cdot \mid \vx) = \int p(\cdot \mid \vx, \vec{\theta})\, p(\vec{\theta} \mid \mathcal{D})\, \mathrm{d}\vec{\theta}, \end{equation} 
which is approximated in practice via Monte Carlo sampling using $M$ weight samples ${\vec{\theta}_1,...,\vec{\theta}_M}$.
The predictive distribution and uncertainty decomposition follow the same procedure as for the ensemble. In the benchmark, we use $M=100$ weight samples.
We approximate the posterior by a mean-field Gaussian over linear and Conv2d weights and learn its means and standard deviations. On CIFAR-10, the prior has mean zero and standard deviation $0.5$, with initial posterior standard deviation $0.01$. On ImageNet, both the prior and initial posterior means equal the pretrained weights; their standard deviations are $0.05$ and $0.001$, respectively.

\subsubsection{MC Dropout} MC Dropout \citep{galDropout2016} interprets dropout-regularized neural networks as approximate Bayesian inference. During training, dropout is applied such that each activation in the model will be set to zero with probability $p$. At test time, dropout is kept active, and $M$ stochastic forward passes are performed for a given instance $\vx$, each corresponding to a different random masking of the network activations. The resulting $M$ predictions are treated as approximate posterior predictive samples, and the predictive distribution and uncertainty decomposition follow the same procedure as for the ensemble. In the benchmark, we use $M=100$ stochastic forward passes and a dropout rate of $p=0.5$.

\subsubsection{DropConnect} DropConnect \citep{wanDropConnect2013} extends the MC Dropout approach by randomly masking individual weights rather than activations during the forward pass. This yields a finer-grained stochastic approximation to the posterior over weights. At test time, $M$ stochastic forward passes are performed analogously to MC Dropout, with independent weight masks sampled for each pass. The predictive distribution and uncertainty decomposition follow the same procedure as for the ensemble. In the benchmark, we use $M=100$ stochastic forward passes and a drop rate of $p=0.5$.

\subsubsection{Deep Deterministic Uncertainty (DDU)}
DDU \citep{mukhotiDDU2023} produces uncertainty estimates from a single deterministic forward pass by fitting a class-conditional Gaussian density model on the feature space of a trained network. To ensure that the feature space is well-suited for density estimation, the backbone is regularized using spectral normalization, which, together with the residual connections of the ResNet architecture, makes the feature map approximately bi-Lipschitz. Given an instance $\vx$, the epistemic uncertainty is quantified by the negative log-likelihood of the feature representation under the fitted Gaussian mixture, while the aleatoric uncertainty is captured by the entropy of the softmax output.
No total component is exposed by the default implementation; selective prediction therefore falls back to the softmax entropy.

\subsubsection{Deterministic Uncertainty Quantification (DUQ)}
DUQ \citep{vanAmersfoortDUQ2020} replaces the standard softmax output with a radial basis function (RBF) network that maps feature representations to per-class centroids. Given an instance $\vx$, the model computes a kernel distance between the feature representation and each class centroid, and the resulting distances are interpreted as class-wise uncertainty scores. To preserve meaningful distances in feature space, the feature extractor is regularized with a gradient penalty. Total uncertainty is measured by the maximum kernel distance across classes, with no explicit decomposition into aleatoric and epistemic components. We follow the original paper and use the kernel distance as the sole uncertainty measure.
We fix the centroid dimension to $512$, RBF length scale to $0.1$, centroid EMA decay to $0.999$, and gradient-penalty coefficient to $0.5$.

\subsubsection{Deep Anti-Regularized Ensembles (DARE)}
DARE \citep{deMathelinDARE2023} trains an ensemble using a particular loss function designed to maximize ensemble member diversity. In addition to the standard supervised loss, an anti-regularization term enforces weights to become as large as possible while performing well on train data. As a result, performance is kept for in-distribution data while it collapses on out-of-distribution data, enabling the detection of such samples. The predictive distribution and uncertainty decomposition follow the same procedure as for the ensemble. In the benchmark, we use an ensemble of $M=10$.

\subsubsection{HetNet}
HetNets \citep{collierHETNet2021} aim to model the input-dependent, also known as heteroscedastic, label noise. By placing a multivariate normal distributed latent variable on top of the final hidden layer of a neural network classifier. Rather than predicting a single logit vector, the network outputs then both a mean and a variance for each logit, from which $M$ logit samples are drawn and averaged in probability space to form the predictive distribution. This allows the model to capture label noise that varies as a function of the input. Epistemic uncertainty is not explicitly modeled; we therefore take the entropy of the predictive distribution as the total uncertainty and the mean per-sample entropy as the aleatoric component. In the benchmark, we use $M=100$ samples, a diagonal-plus-low-rank logit covariance with $15$ factors, temperature $0.9$, and the non-parameter-efficient parameterization.

\subsubsection{Evidential Deep Learning}
Evidential deep learning \citep{sens_ed18} replaces the standard softmax output with a Dirichlet distribution over the simplex, treating classification as the \emph{direct} prediction of a second-order distribution. Concretely, the network output is interpreted as a concentration vector $\bm{\alpha} \in \bR^K_{\geq 1}$ that parametrizes a Dirichlet distribution $\operatorname{Dir}(\bm{\alpha})$ with $\bm{\alpha} = \exp(h(\vx)) + 1$. Training is performed by minimizing the MSE loss from \citep{sens_ed18}, which minimizes the Bayes risk, while keeping a KL-divergence regularizer. The mean of the Dirichlet distribution provides the predicted first-order distribution $\bar{p}(y_k \mid \vx) = \frac{\alpha_k}{\sum_{k=1}^K\alpha_k}$. The total uncertainty can be computed as $H(\bar{p}(\cdot \mid \vx))$ and the epistemic uncertainty is the \emph{vacuity} $\frac{K}{\sum_{k=1}^K\alpha_k}$.
The final KL coefficient is $1.0$ with annealing over $100$ epochs on CIFAR-10, and $0.01$ with annealing over $5$ epochs on ImageNet.

\subsubsection{SNGP}
The Spectral-Normalizing Gaussian Process \citep{liuSNGP2020} is a single forward-pass method designed to make a deterministic deep neural network distance-aware. This is achieved by applying two modifications. First, spectral normalization is applied to all residual blocks in order to bound the Lipschitz constant of the feature extractor $f$ such that distances in the feature space approximately reflect distances in the input space. Second, the linear output layer, is replaced by a Gaussian process with an RBF kernel resulting in $K$ trainable parameters $\beta_k$. The model is then trained normally, and predictions are made by sampling 100 samples from the Gaussian process. Similarly, uncertainty is computed by drawing 100 samples and using \cref{eq:entropy-decomposition}.
We use $1{,}024$ frozen random features, spectral-norm bound $6.0$, ridge penalty $1.0$, and one power iteration per update. The random-projection initialization standard deviation is $1.0$ on CIFAR-10 and $0.05$ on ImageNet.

\subsubsection{Posterior Network}
Posterior Network, akin to EDL, also outputs a Dirichlet distribution over the simplex, but constructs it through density estimation in a learned latent space rather than from network logits. The models consists of an encoder $f : \vx \mapsto \bm{z}$ and a per-class radial normalizing flow $\mathbb{P}(\bm{z} \mid y_k)$ that estimates a density on the latent space for each class. The Dirichlet parameters for class $k$ for an instance $\vx$ is then $\alpha_k = 1 + N_k * \mathbb{P}(\bm{z} \mid y_k)$, where $N_k$ is the number of training instances for class $y_k$.  $N_c \mathbb{P}(\bm{z} | y_k)$ is a pseudo-count of how many training samples of class $y_k$ the latent representation looks like, and a test input far from any class density receives close to zero evidence and hence a near-uniform Dirichlet. Predictions are made by taking the argmax of the Dirichlet mean. The encoder and flows are trained jointly by minimizing a Bayesian loss combining the expected cross-entropy under the Dirichlet with an entropy regularizer. The default aleatoric score is $1-\max_k(\alpha_k/\sum_j\alpha_j)$, and epistemic uncertainty is the vacuity. No total score is exposed, so selective prediction falls back to the aleatoric score.
We fix the latent dimension to $6$ on CIFAR-10 and $16$ on ImageNet, use six radial flow transformations per class, and set the entropy-regularization coefficient to $10^{-5}$. Class counts $N_k$ are computed from the training data.

\subsection{Variational Bayesian Last Layer}
The variational Bayesian last layer approach \citep{harrisonVBLL2024} is a partially Bayesian method in which the feature extractor $f$ (all but the last layer) remains deterministic, and only the last layer $W \in \bR^{K \times d}$ is equipped with a Gaussian variational posterior $q(W) = \prod_{k=1}^K \mathcal{N}(\bar{\vw}_k, S_k)$, where $\bar{\vw}_k$ denotes the mean weight vector of row $k$ of $W$, and $S_k$ is the covariance. The covariance is parameterized through a Cholesky factor $S_k = L_kL_k^T$, where $L_k$ is either a lower-triangular matrix with positive diagonal (dense parameterization) or a positive diagonal matrix (diagonal parameterization).
Thus, this essentially amounts to a Bayesian logistic regression on learned features. 
The original paper proposes multiple variants, all of which are included in \probly, but we use the standard discriminative classifier, obtained by replacing the last layer of the base model. 
The resulting model is trained by minimizing the training objective in \citep{harrisonVBLL2024} with the hyperparameters described therein. For the tabular datasets and CIFAR-10, we use the dense parameterization, whereas for ImageNet the diagonal parameterization is used. 
At test time, a forward pass then yields a Gaussian over the logits from which $M = 100$ samples are drawn and passed through a softmax. The predictive distribution and uncertainty decomposition follow the same procedure as for ensemble models. 

\subsection{Credal Set}
\subsubsection{Credal Wrapper}
The credal wrapper is created in the same way as an ensemble --- by training $M$ members and relying on stochasticity of the initialization and training procedure to create diversity. Given an instance $\vx$, the lower and upper probabilities for each class are then computed by $\underline{p}(y_k \mid \vx) = \min_{i \in [M]} p(y_k \mid \vx, h_i)$ and $\overline{p}(y_k \mid \vx) = \max_{i \in [M]} p(y_k \mid \vx, h_i)$; using these an interval-based credal set can be constructed \citep{camposProbabilityIntervals1994}:
\begin{equation*}
    \cQ = \{p(\cdot \mid \vx) : \underline{p}(y_k \mid \vx) \leq p(y_k \mid \vx) \leq \overline{p}(y_k \mid \vx), \ \forall y_k \in \cY \}.
\end{equation*}
On the basis of this credal set, a prediction is made by considering the intersection probability \citep{cuzzolinIntersectionProbability2022}: 
\begin{equation*}
    \tilde{p}(y_k \mid \vx) = \underline{p}(y_k \mid \vx) + \alpha ( \overline{p}(y_k \mid \vx) - \underline{p}(y_k \mid \vx)), \ \forall y_k \in \cY,
\end{equation*}
where 
\begin{equation*}
    \alpha = \frac{1 - \sum_{k=1}^{K} \underline{p}(y_k \mid \vx)}{\sum_{k=1}^{K} (\overline{p}(y_k \mid \vx) - \underline{p}(y_k \mid \vx))}.    
\end{equation*}
The uncertainty is quantified by using a decomposition based on upper and lower Shannon entropy \citep{abellan2006disaggregated}:
\begin{equation}\label{eq:credal-entropy}
    \overline{H}(\cQ) = \underline{H}(\cQ) + (\overline{H}(\cQ) - \underline{H}(\cQ)),    
\end{equation}
where $\overline{H}(\cQ) = \sup_{p(\cdot \mid \vx) \in \cQ} \operatorname{H}p(\cdot \mid \vx)$ and $\underline{H}(\cQ) = \inf_{p(\cdot \mid \vx) \in \cQ} \operatorname{H}p(\cdot \mid \vx)$ are the upper and lower Shannon entropy, respectively.
We use $M=10$ and reuse the trained deep-ensemble checkpoint (\cref{app:implementation-details:ensemble}), with no additional model fitting.

\subsubsection{Credal Ensembling}
Credal ensembling is trained in the same way as Credal Wrapper and Ensemble --- that is, by training $M = 10$ members of the base model. The credal set, however, is constructed differently. Given an instance $\vx$, a representative distribution is first identified as the minimizer of the average squared distance to the ensemble members. In this setting, we choose the Euclidean distance, which means the representative distribution reduces to $\bar{p}(\cdot \mid \vx) = \frac{1}{M}\sum_{m=1}^M p(\cdot\mid\vx,h_m)$, where $\{h_1, \ldots, h_M\}$. The probability distribution $p(\cdot\mid\vx,h_m)$ are then ordered by their distance to $\bar{p}(\cdot\mid\vx)$ and the ${(1-\alpha)M}$ closest probability distributions are retained, yielding a set $H_\alpha(\vx)$. The credal set is the convex hull of this credal set $\cQ = \operatorname{Conv}(H_\alpha(\vx))$ with the threshold $\alpha \in [0,1)$ controlling the fraction of outlying members discarded. We use $\alpha = 0$, as this has been shown to produce the best results in previous work \citep{loehrCredalPrediction2025}. Predictions are made using the representative distribution and the uncertainty is computed according to \cref{eq:credal-entropy}. 

\subsubsection{Credal BNN}
Credal Bayesian Neural Networks constructs a credal set from a finite set of \ac{bnn}s, each obtained from a different prior. The paper does not list any hyperparameters related to the priors and an empty code repository \url{https://github.com/PRECISE/credal-bayesian-deep-learning} is provided. We use a set of $M=10$ priors, with 0 mean.
On CIFAR-10, we use $\{0.1, 0.15, 0.2, 0.3, 0.45, 0.7, 1.0, 1.5, 2.0, 3.0\}$ as standard deviations; on ImageNet we use $\{0.01, 0.015, 0.02, 0.03, 0.04, 0.05, 0.07, 0.09, 0.12, 0.15\}$. The \acp{bnn} are then trained following the setup in \cref{sec:bnns}. To make predictions, we take 20 samples from each \ac{bnn} and average them, the $M$ Bayesian model averages are then used to create $\cQ = \operatorname{Conv}(\{\bar{p}(\cdot \mid \vx, h_m)\})$. Class-wise predictions are then made by $\hat{y} = \argmax_{y\in\cY}\min_{m \in [M]}p(\cdot \mid \vx, h_m)$. 

\subsubsection{Credal Relative Likelihood}
The credal relative likelihood predictor also makes use of ensembling. Members are randomly initialized on CIFAR-10 and retain pretrained initializations on ImageNet; for each constrained member $m$ the bias of class $m$ is set to $\beta = 100$ following the paper's recommended setting. Before training, the user specifies an $\alpha$ --- this defines a range of $M=10$ threshold in $[\alpha,1]$, which are used for early stopping of the training procedure based on the relative likelihood compared of the predictor compared to the maximum likelihood estimate. Based on the results in \citep{loehrCredalPrediction2025} we use $\alpha=1$. Prediction are made using the probabilistic predictions of the MLE. Uncertainty is quantified based on \cref{eq:credal-entropy} using the interval-based credal set constructed using the $M$ members. 

\subsubsection{Efficient Credal Prediction}
Efficient credal prediction \citep{hofmanEfficientCredalPrediction2026} creates interval-based credal sets around a maximum likelihood estimate. That is, given a model $h^{ML}$, it constructs lower and upper bounds for the class probabilities based on the relative likelihood of the resulting ``predictors''. We use $\alpha = 0.5$ in the supervised benchmark and $\alpha = 0.8$ in active learning following the ablation in \citep{hofmanEfficientCredalPrediction2026}.

\subsubsection{Conformalized Credal Set Predictor}
The conformalized credal set predictor \citep{javanmardi2024conformalized} differentiates itself from the other credal predictors because it uses first-order data in its conformal calibration process. Starting from a trained base model $h$, the conformalized credal predictor is calibrated on the first-order data using a non-conformity score, which is the distance between the predictions from $h$ and the ground-truth distributions. In this case, we use total variation distance. Based on this, the $\alpha$-quantile $q_\alpha$ is computed and the conformal credal sets are constructed by providing TVD balls with radius $q_\alpha$ around predictions $h(\vx)$. Different $\alpha$s are chosen in the evaluation.
We reuse the base checkpoint and reserve $20\%$ of the first-order data for calibration. The radius is calibrated from these data, while $\alpha$ is fixed per run.

\subsection{Calibration}
\subsubsection{Label Smoothing}
Label smoothing \citep{szegedyRethinkingInceptionArchitecture2016} trains a 
hypothesis $h: \cX \fromto \mathbb{P}(\cY)$ by replacing the hard target 
$\delta_{y}$, concentrated on the observed label $y \in \cY$, with a 
smoothed target
\begin{equation*}
    \tilde{\prob}(y_k) 
    = (1 - \varepsilon)\, \mathds{1}[y_k = y] + \frac{\varepsilon}{K},
    \quad k = 1, \ldots, K,
\end{equation*}
where $K = |\cY|$ and $\varepsilon \in [0,1]$ controls the smoothing 
strength, redistributing probability mass uniformly across all classes. 
The hypothesis is trained to minimize the cross-entropy between 
$\tilde{\prob}$ and the predictive distribution 
$\prob(\cdot \given \vec{x}, h)$.
In the benchmark, we fix $\varepsilon=0.1$.

\subsubsection{Label Relaxation}
Label relaxation \citep{DBLP:conf/aaai/LienenH21} adopts a set-valued 
perspective: rather than committing to a single softened target, each 
observed label $y \in \cY$ is replaced by an ambiguity set of admissible 
label distributions,
\begin{equation*}
    \cQ^{\alpha}(y) 
    = \left\{ \prob \in \mathbb{P}(\cY) \,:\, 
    \sum_{y_k \neq y} \prob(y_k) \leq \alpha \right\},
\end{equation*}
where $\alpha \in [0,1]$ controls the degree of relaxation. The hypothesis 
$h$ is trained with a relaxation loss that incurs zero penalty for any 
prediction $\prob(\cdot \given \vec{x}, h) \in \cQ^{\alpha}(y)$, while 
predictions outside the set are penalised by their minimal KL divergence 
to $\cQ^{\alpha}(y)$.
In the benchmark, we fix $\alpha=0.1$.

\subsubsection{Temperature Scaling}
Temperature scaling \citep{guoOnCalibration2017} is a post-hoc 
calibration method that rescales the logits of an already-trained 
hypothesis $h: \cX \fromto \mathbb{P}(\cY)$ by a single scalar 
parameter $T > 0$. Let $z(\vec{x}) = (z_1(\vec{x}), \ldots, 
z_K(\vec{x})) \in \mathbb{R}^K$ denote the logits produced by $h$; 
the temperature-scaled predictive distribution is then
\begin{equation*}
    \prob_T(y_k \given \vec{x}, h) 
    = \frac{\exp(z_k(\vec{x}) / T)}
           {\sum_{j=1}^{K} \exp(z_j(\vec{x}) / T)},
    \quad k = 1, \ldots, K.
\end{equation*}
The temperature $T$ is fitted on a held-out calibration set by 
minimising the negative log-likelihood. As $T$ acts as a uniform 
rescaling of the logits, the argmax of the predictions is preserved 
and only their sharpness is modified: $T > 1$ flattens the 
distribution toward uniformity, while $T < 1$ sharpens it.

\subsubsection{Vector Scaling} Vector scaling \citep{guoOnCalibration2017} generalizes temperature scaling by replacing the single scalar $T$ with an affine transformation of the logits. Concretely, let $\vec{w} \in \mathbb{R}^K$ and $\vec{b} \in \mathbb{R}^K$ be class-wise scale and bias parameters; the vector-scaled predictive distribution is \begin{equation*} \prob_{\vec{w},\vec{b}}(y_k \given \vec{x}, h) = \frac{\exp(w_k z_k(\vec{x}) + b_k)} {\sum_{j=1}^{K} \exp(w_j z_j(\vec{x}) + b_j)}, \quad k = 1, \ldots, K. \end{equation*} As in temperature scaling, the parameters $\vec{w}$ and $\vec{b}$ are fitted post-hoc on a held-out calibration set by minimising the negative log-likelihood, leaving the weights of $h$ unchanged. By allowing each class its own scale and bias, vector scaling can correct for class-dependent miscalibration that a single temperature parameter cannot capture. Temperature scaling is recovered as the special case $\vec{w} = \frac{1}{T}\mathbf{1}$ and $\vec{b} = \mathbf{0}$.

\clearpage

\section{Details of the Experimental Setting and Reproducibility}\label{sec:appendix_reproducibility}

\subsection{Computational Resources}\label{sec:appendix_computational_resources}
Experiments were conducted using NVIDIA B200, H200, and A100 GPUs. Each model was trained on a single GPU with access to up to 16 CPU cores and 300 GB of RAM.
The overall computational budget of this work is estimated at approximately 1,500 GPU hours. Training times varied depending on the dataset and model type. On ImageNet-scale experiments, single models required roughly 3 hours of training, while ensemble-based models required approximately 25 hours. For CIFAR-10 experiments, single models trained in approximately 30 minutes, whereas ensemble-based models required around 5 hours.

\subsection{Data Availability and Reproducibility}\label{sec:appendix_data_availability_reproducibility}

To ensure reproducibility and fair comparison across methods, we use the same random seeds ($1, 2, 3$) in our experiments for all sources of randomness throughout the codebase. This includes model initialization, data shuffling, and dataset splitting, resulting in consistent experimental conditions across all evaluated methods.

Furthermore, all hyperparameter configurations used in our experiments are provided in our GitHub repository \githubrepo. This includes training settings, optimization parameters, and dataset-specific configurations required to reproduce the reported results.

All datasets used in this work are publicly available and commonly used benchmark datasets.

\subsection{Datasets}

\paragraph{ImageNet}
ImageNet, introduced by \citet{dengImageNet2009}, is a large-scale image dataset structured according to the WordNet hierarchy, comprising over $14$ million images spanning more than $20,000$ categories. Its ILSVRC subset, commonly referred to as ImageNet-1K, contains $1,281,167$ training images, $50,000$ validation images, and $100,000$ test images across $1,000$ classes. \textit{The dataset is freely accessible to researchers for non-commercial purposes}.

\paragraph{CIFAR-10}
CIFAR-10, introduced by \citet{krizhevsky2009learning} and Geoffrey Hinton in 2009, is a widely adopted benchmark in machine learning and computer vision. It consists of $60,000$ color images with a resolution of $32\times32$ pixels, evenly divided among $10$ classes: airplane, automobile, bird, cat, deer, dog, frog, horse, ship, and truck. The dataset is split into $50,000$ training images and $10,000$ test images, organized into five training batches and a single test batch, each containing $10,000$ images. CIFAR-10 is publicly available and has been extensively used for training and evaluating machine learning models. While the original dataset does not explicitly define a license, versions distributed through platforms such as TensorFlow datasets are provided under the \textit{Creative Commons Attribution 4.0 License}.

\paragraph{CIFAR-100}
CIFAR-100, introduced by \citet{krizhevsky2009learning}, consists of $60,000$ color images at a resolution of $32\times32$ pixels, organized into $100$ classes with $600$ images per class. Each image carries both a “fine” label, indicating its specific class, and a “coarse” label corresponding to one of $20$ broader superclasses. The dataset is divided into $50,000$ training images and $10,000$ test images. CIFAR-100 is derived from the Tiny Images dataset and is widely used for benchmarking image classification models. While the original dataset does not define a license, versions distributed through platforms such as TensorFlow Datasets are available under the \textit{Creative Commons Attribution 4.0 License}.

\paragraph{SVHN}
The SVHN dataset, introduced by \citet{netzer2011reading}, contains over $600,000$ $32\times32$ RGB images of digits ($0 - 9$) extracted from real-world house numbers in Google Street View. It is organized into three subsets: $73,257$ images for training, $26,032$ for testing, and an additional set of $531,131$ images for extended training. SVHN is intended for digit recognition tasks and requires minimal preprocessing. Although the original dataset does not specify a license, versions distributed through platforms like TensorFlow Datasets are available under the \textit{Creative Commons Attribution 4.0 License}.

\paragraph{Places365}
Places365, introduced by \citet{zhou2017places}, is a large-scale dataset for scene recognition, comprising $1.8$ million training images spanning $365$ scene categories. The validation set contains $50$ images per category, while the test set includes $900$ images per category. An expanded variant, Places365-Challenge-2016, incorporates an additional $6.2$ million images and $69$ new scene categories, bringing the total to $8$ million images across $434$ categories. Although the original dataset does not specify a license, versions distributed through platforms such as TensorFlow Datasets are available under the \textit{Creative Commons Attribution 4.0 License.}

\paragraph{MNIST}
MNIST, introduced by \citet{lecun-mnisthandwrittendigit-2010}, is a large database of handwritten digits derived from the NIST Special Database, commonly used for training and evaluating image processing and machine learning systems. It consists of $70,000$ grayscale images at a resolution of $28\times28$ pixels, spanning $10$ classes corresponding to the digits $0$ through $9$. The dataset is split into $60,000$ training images and $10,000$ test images, with $6,000$ and $1,000$ images per class, respectively. Images were size-normalized and centered, with anti-aliasing introducing grayscale levels. The dataset is made available under the \textit{Creative Commons Attribution-Share Alike 3.0 License}.

\paragraph{Textures}
The Describable Textures Dataset (DTD), introduced by \citet{cimpoiTextures2014}, is a collection of textural images gathered in the wild, annotated with human-centric perceptual attributes. It consists of $5,640$ images organized into $47$ texture categories, with $120$ images per category. Image resolutions range from $300\times300$ to $640\times640$ pixels, and each image contains at least 90\% surface coverage of the depicted texture attribute. The dataset is evenly divided into three splits—training, validation, and test—each comprising $40$ images per class. Images were collected from Google and Flickr and annotated via Amazon Mechanical Turk. The dataset is made available to the computer vision community \textit{for research purposes}.

\paragraph{iNaturalist}
The iNaturalist Species Classification and Detection Dataset, introduced by \citet{vanHornINaturalist2018}, is a large-scale fine-grained recognition benchmark sourced from the citizen science platform iNaturalist. It is characterized by a heavy class imbalance reflecting the natural observation frequency of species in the wild. The dataset covers $8,142$ species across $14$ super-categories, comprising $437,513$ training images and $24,426$ validation images. Images were collected with varying camera types and conditions, and verified by multiple citizen scientists. Although the dataset does not specify an explicit license, images distributed through platforms such as TensorFlow Datasets are provided under the \textit{Creative Commons Attribution 4.0 License}.

\paragraph{SSB-Hard}
SSB-Hard is the hard split of the Semantic Shift Benchmark (SSB), introduced by \citet{vazeSSBHard2022} as part of a study on open-set recognition. It is designed to evaluate near out-of-distribution detection for models trained on ImageNet-1K, using semantically similar but disjoint classes drawn from ImageNet-21K. The unknown classes are partitioned by their degree of semantic similarity to the ImageNet-1K training classes, with the hard split containing the classes most visually and semantically confusable with in-distribution categories. The test set consists of $49,000$ images. SSB-Hard does not specify an explicit license; images are inherited from their respective source datasets, primarily governed by the ImageNet terms of use.

\paragraph{NINCO}
NINCO (No ImageNet Class Objects), introduced by \citet{bitterwolfNINCO2023}, is a manually curated out-of-distribution benchmark designed to fix known contamination issues in existing \ac{ood} test sets for ImageNet-1K. Each of the $5,879$ images was individually inspected to ensure it contains no objects belonging to any of the $1,000$ ImageNet-1K classes, making the dataset strictly \ac{ood}. The images are organized into $64$ fine-grained \ac{ood} categories, enabling detailed analysis of an \ac{ood} detector's failure modes. In addition to the main dataset, the authors provide cleaned versions of $2,715$ \ac{ood} samples from eleven previously used \ac{ood} datasets, as well as $17$ synthetic \ac{ood} unit-tests with $400$ samples each for probing specific detector weaknesses. The dataset is released under the \textit{MIT License}.

\paragraph{CIFAR-10H}
CIFAR-10H provides human-generated soft labels for the $10,000$ images in the CIFAR-10 test set, reflecting the variability in human judgments for image classification. Introduced by \citet{peterson2019cifar10h}, the dataset contains $511,400$ annotations from $2,571$ workers on Amazon Mechanical Turk, with each image receiving around $51$ labels. Each annotation assigns an image to one of the ten CIFAR-10 classes, allowing the creation of a probability distribution over labels for every image. CIFAR-10H is publicly available under the \textit{Creative Commons BY-NC-SA 4.0 License}.

\paragraph{QualityMRI}
Introduced by \citet{obuchowicz2020qualityMRI}, the QualityMRI dataset is part of the Data-Centric Image Classification (DCIC) Benchmark, which studies the role of dataset quality in shaping model performance. It consists of $310$ magnetic resonance images that cover different quality levels, providing a resource for assessing MRI image quality. The dataset is distributed under the \textit{Creative Commons BY-SA 4.0 License}.

\paragraph{MiceBone}
The MiceBone dataset is part of the Data-Centric Image Classification (DCIC) Benchmark introduced by \citet{schmarje2022one}. It consists of $7,240$ microscopy images of collagen fibers in mice bone tissue. Each image is annotated by between one and five professional annotators and assigned to one of three classes: directed collagen fibers, undirected collagen fibers, or non-relevant structures caused by noise or background artifacts. To better capture annotation uncertainty and ambiguity, additional annotations were collected, resulting in approximately $15$ annotations per image on average. The dataset is characterized by substantial class imbalance, with around $70\%$ of samples belonging to the non-relevant class, making it a challenging benchmark for uncertainty-aware and label-distribution learning methods. The dataset is publicly available as part of the DCIC benchmark collection.

\paragraph{Synthetic}
Synthetic is a dataset of synthetically generated images featuring a single circle or ellipse in blue, red, or green on a black background \citep{schmarje2022one}. The task is to classify the object by its color and shape. To introduce ambiguity, the dataset includes interpolations in both color and axis dimensions between these classes.

\paragraph{OpenML}
OpenML \cite{OpenML2025} is an open platform for sharing datasets, machine learning tasks, and experimental results, widely used to support reproducible machine learning research. In this work, we use OpenML datasets \texttt{6} and \texttt{156}, corresponding to the \textit{letter} and \textit{RandomRBF} datasets, respectively. The \textit{letter} dataset contains $20,000$ samples of handwritten capital letters represented by $17$ numerical features and classified into $26$ classes. The \textit{RandomRBF} dataset consists of $1,000,000$ synthetic samples from $5$ classes described by $11$ attributes. Both datasets are publicly accessible through the OpenML repository and are commonly used as tabular benchmarks for classification tasks.

\subsection{Models}
Here, we describe the architecture of the models we use and the setup we use to train them. The base hyper-parameters for all models are found in \cref{tab:hyperparameters}. The ResNet models are trained without early stopping, but the best performing model on the validation set is used; the validation split is 0.1. For CIFAR-10 this split is taken from the dedicated training set, for ImageNet the validation split is split into a validation split and a test split. 

\begin{table}[h]
    \centering
    \caption{Hyperparameter configuration.}
    \begin{tabular}{l c c c}
        \toprule
        & \textbf{ResNet18} & \textbf{ResNet50} & \textbf{MLP} \\
        \midrule
        \textbf{Optimizer}     & SGD    & Lamb \citep{youLambOptimizer2020} & Adam \\
        \textbf{Epochs}        & 200    & 20                 &  100 \\
        \textbf{Batch Size}    & 128    & 2048               & 128 \\
        \textbf{Learning Rate} & 0.1    & 5e-3               & 0.001 \\
        \textbf{Weight Decay}  & 5e-4   & 1e-2               & --- \\
        \textbf{Momentum}      & 0.9    & 0                  & --- \\
        \textbf{Scheduler}     & Cosine & Cosine (2 warmup)  & Cosine \\
        \bottomrule
    \end{tabular}
    \label{tab:hyperparameters}
\end{table}

\paragraph{ResNet18}
We use the ResNet18 \citep{heDeepResidualLearning2016} implementation from \url{https://github.com/kuangliu/pytorch-cifar}. This version replaces the first convolution layer by a 3x3 stride-1 convolutional layer and removes the max-pooling layer afterwards to make it more suitable for CIFAR-10 images with a 32x32 resolution. The model is trained from scratch following the hyperparameters suggested in the implementation.

\paragraph{ResNet50}
We employ the ResNet50 \citep{heDeepResidualLearning2016} implementation from \texttt{torchvision} \citep{torchvision2016}. The model is initialized using the default ImageNet weights. We fine-tune it following a setup similar to \citep{mucsanyiBenchmarkingUncertainty2024,wightmanResNetStrikesBack2021}. 

\paragraph{Multilayer Perceptron}
We employ a \ac{mlp} with two hidden layers of
dimension 1024 and ReLU activations, following~\citep{activeLearningBadge2020}.
The model is trained from scratch with the Adam optimizer at a learning
rate of $10^{-4}$.

\subsection{Selective Prediction}
We use the trained uncertainty methods for the selective prediction task. The task is performed on the full test split of the respective dataset. At each rejection level, the instances are sorted by the total uncertainty measure reported in the respective paper and the predictions are made based on the predictive distribution put forward by the method. 

\subsection{Out-of-Distribution Detection}
For out-of-distribution detection, we use the dedicated test sets of each dataset. The \ac{ood} dataset always receives the same transformation (e.g. resizing) as the iD data. In order to balance the binary classification problem (\ac{id} vs \ac{ood}), we subsample the test set down to the size of the smaller one. The epistemic uncertainty is computed using the measure reported in the respective paper. These values are then used to discriminate between in-distribution and out-of-distribution data. Based on this the \ac{auroc} is computed. 

\subsection{Active Learning}\label{sec:appendix:active_learning}

\textbf{Pool-based protocol.}
We follow the following pool-based active learning protocol on two
tabular OpenML datasets: ID~6 (26 classes, $\sim$20{,}000 instances) and
ID~156 (5 classes, 1{,}000{,}000 instances). In each run, the dataset is
first split into 80\% training and 20\% test data using a
class-stratified split (\texttt{sklearn.train\_test\_split}) seeded by
the run's seed. Features are standardized with a \texttt{StandardScaler}
fit on the training split only, and class labels are integer-encoded.
The training labels are then hidden, partitioning the training split
into a labeled set $\mathcal{D}_L^0$ of $n_0$ instances drawn uniformly
at random and an unlabeled pool $\mathcal{D}_U^0$ containing the
remaining instances. The held-out test set remains fixed throughout the
run and is used only for evaluation.

\textbf{Query and retraining rounds.}
In each of the $T$ acquisition rounds, we (i)~train the base model
\emph{from scratch} on the current labeled set $\mathcal{D}_L^t$,
(ii)~evaluate test accuracy on the held-out test set, (iii)~score all
instances in the unlabeled pool $\mathcal{D}_U^t$ using the acquisition
strategy, and (iv)~remove the $q$ highest-scoring instances from
$\mathcal{D}_U^t$ and add them, together with their ground-truth labels,
to $\mathcal{D}_L^t$, yielding $\mathcal{D}_L^{t+1}$ and
$\mathcal{D}_U^{t+1}$. After the final round, we perform one additional
retraining and evaluation step, yielding $T+1$ accuracy measurements at
$T+1$ distinct labeled-set sizes. 

\textbf{Acquisition budget.}
The initial labeled-set size $n_0$, query size $q$, and number of
acquisition rounds $T$ follow the protocol
of~\citep{activelearningSandraECML} and are summarized
in~\cref{tab:al_budget}.

\begin{table}[h]
\centering
\caption{Active learning acquisition budget per dataset.}
\label{tab:al_budget}
\begin{tabular}{lcccc}
\toprule
OpenML ID & $n_0$ & $q$ & $T$ & Total labels \\
\midrule
6   & 1000 & 1000 & 10 & 11{,}000 \\
156 & 100  & 100  & 15 & 1{,}600  \\
\bottomrule
\end{tabular}
\end{table}

\textbf{Base model and training.}
Across all methods and acquisition strategies, we use the same base
model: a multi-layer perceptron with two hidden layers of dimension
1024 and ReLU activations. In each round, the model is trained for 100
epochs with Adam at a learning rate of $10^{-3}$ under a cosine
learning-rate schedule. The mini-batch size is 128 for ID~6 and 4096
for ID~156; the larger batch size on ID~156 amortizes the cost of
processing its substantially larger unlabeled pool and test set.

\textbf{Evaluation.}
For each run, we record test accuracy at each of the $T+1$ checkpoints
along the acquisition trajectory. The \emph{normalized area under the
budget curve} (NAUC) is defined as the mean accuracy across these
checkpoints, and the \emph{final test accuracy} as the accuracy at the
last checkpoint. Both metrics are averaged over ten random seeds, which
jointly determine the train/test split, the initial labeled set
$\mathcal{D}_L^0$, and the model initialization.

\subsection{Evaluation with First-Order Data}
We train three credal predictors, Credal Ensembling (CreEns), Credal Wrapper (CreWra), and Credal Relative Likelihood (CreRL), on three DCIC datasets: QualityMRI, MiceBone, and Synthetic. Each dataset originally is partitioned into five folds; fold 1 is held out as the test set, and folds 2 to 5 are used for training (with $10\%$ of the training data reserved for validation). All models share a pretrained ResNet-50 backbone and are trained for 50 epochs using SGD with a learning rate of $10^{-2}$ under a cosine learning-rate schedule.
Each method produces an ensemble of size $M=10$.

\clearpage
\section{Additional Details and Experimental Results with \probly}
\label{sec:additional-results}

\begin{figure}[ht]
\centering
\begin{minipage}[c]{0.49\textwidth}
\centering
\includegraphics[width=\linewidth]{figures/appendix/ood/bar_ood_far_imagenet.pdf}
\end{minipage}
\hfill
\begin{minipage}[c]{0.49\textwidth}
\centering
\includegraphics[width=\linewidth]{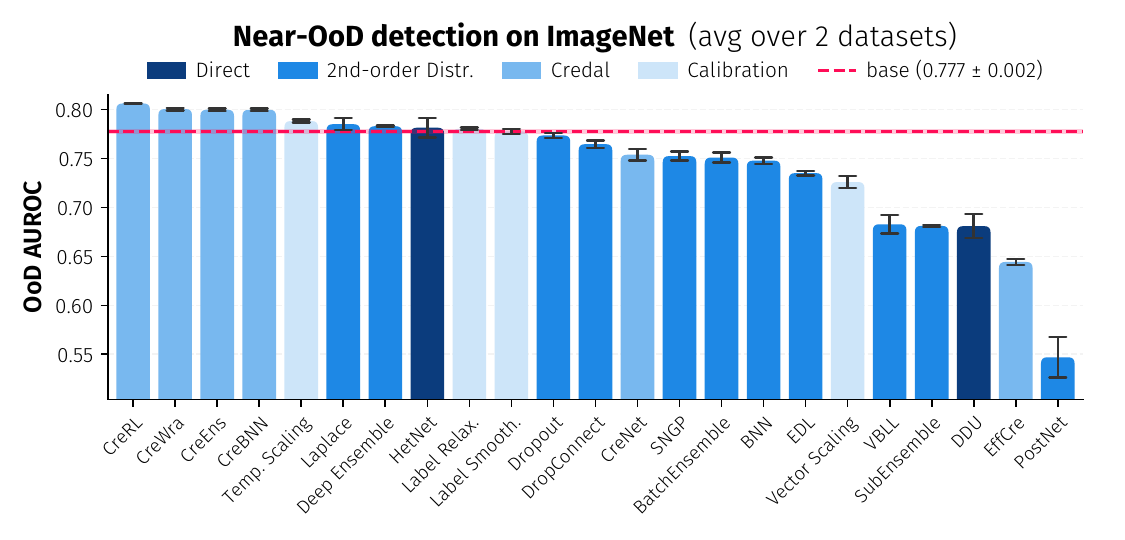}
\end{minipage}
\\
\begin{minipage}[c]{0.32\textwidth}
\centering
\includegraphics[width=\textwidth]{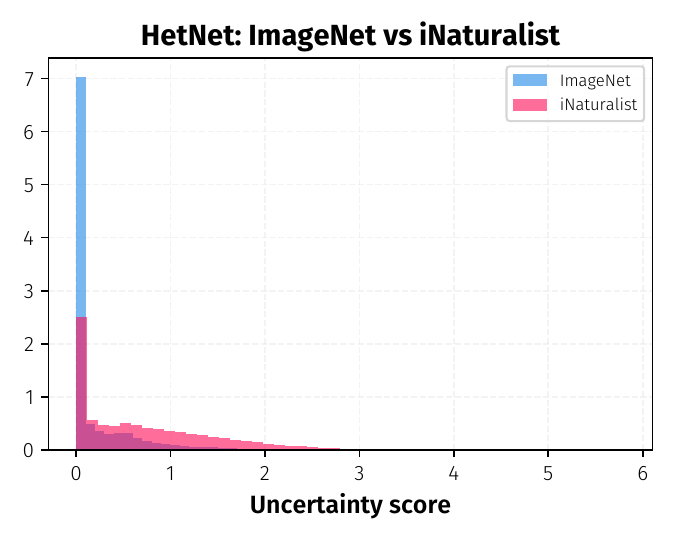}
\end{minipage}
\hfill
\begin{minipage}[c]{0.32\textwidth}
\centering
\includegraphics[width=\textwidth]{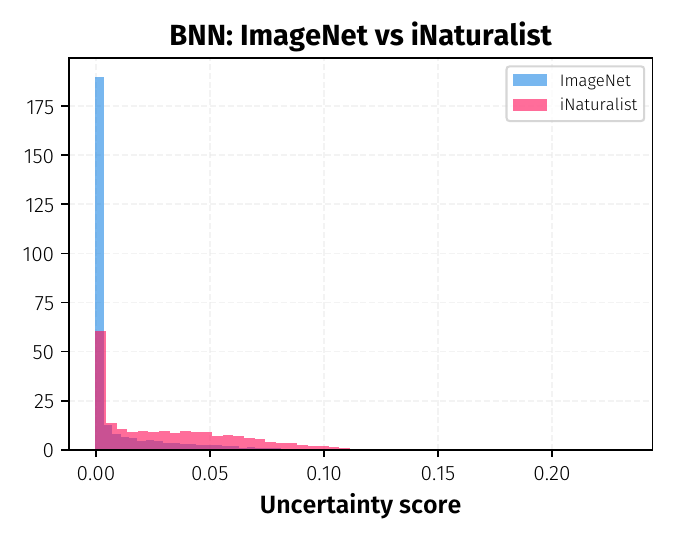}
\end{minipage}
\hfill
\begin{minipage}[c]{0.32\textwidth}
\includegraphics[width=\textwidth]{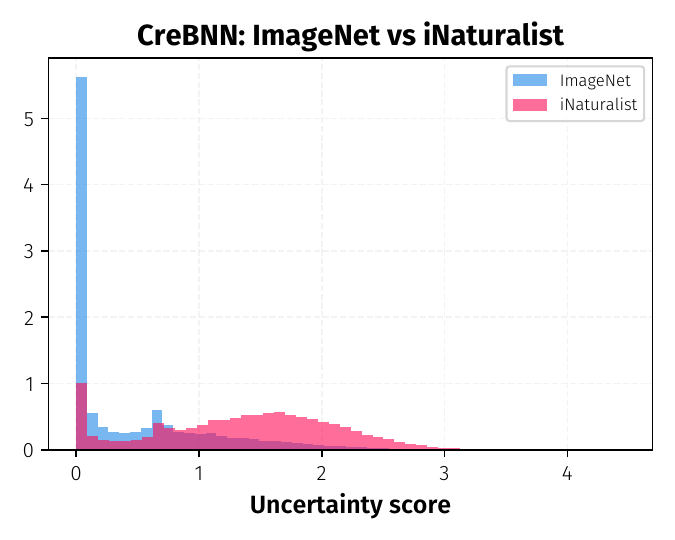}
\centering
\end{minipage}
\caption{A thorough evaluation across all methods in the \ac{ood} detection task for ImageNet. We compare on four \ac{ood} datasets of which a detailed analysis of three methods on the iNaturalist dataset is plotted in the histograms.}
\label{fig:imagenet-ood}
\end{figure}

\begin{figure}[ht]
\centering
\begin{minipage}[c]{0.49\textwidth}
\centering
\includegraphics[width=\linewidth]{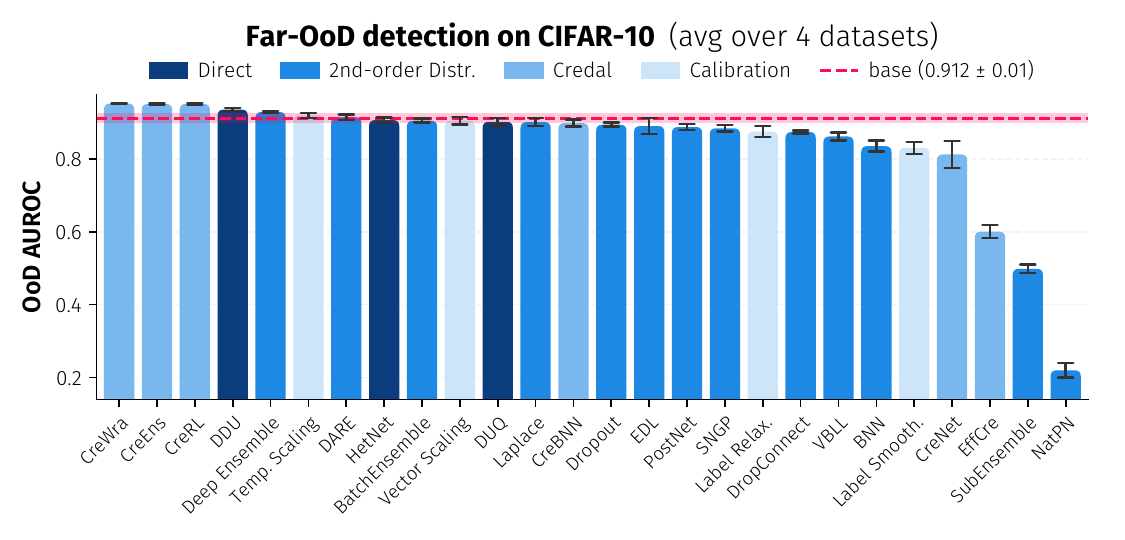}
\end{minipage}
\hfill
\begin{minipage}[c]{0.49\textwidth}
\centering
\includegraphics[width=\linewidth]{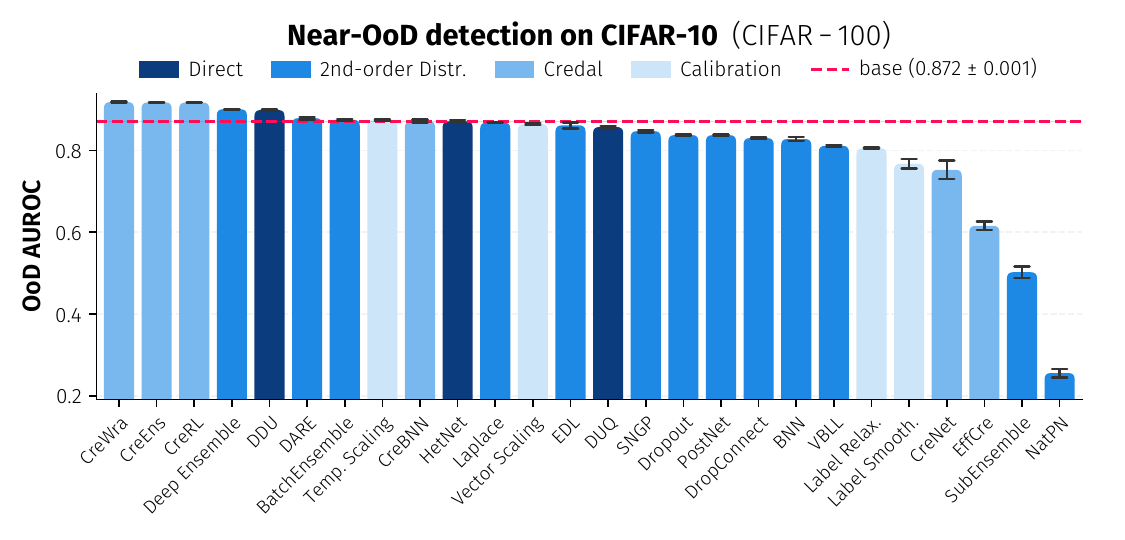}
\end{minipage}
\\
\begin{minipage}[c]{0.32\textwidth}
\centering
\includegraphics[width=\textwidth]{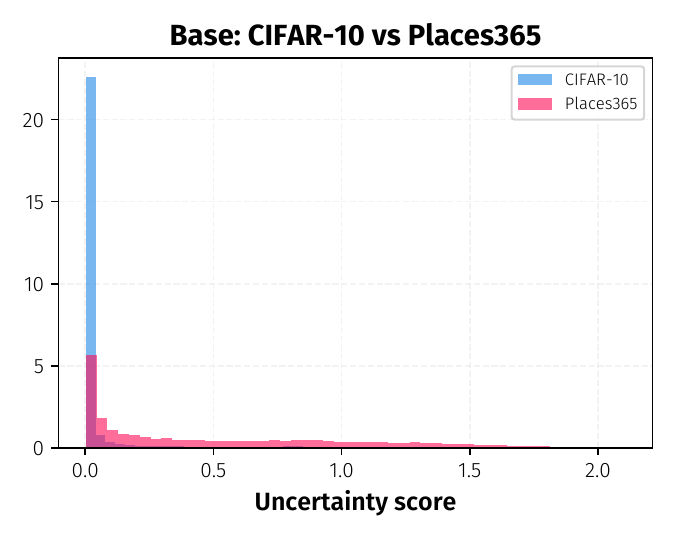}
\end{minipage}
\hfill
\begin{minipage}[c]{0.32\textwidth}
\centering
\includegraphics[width=\textwidth]{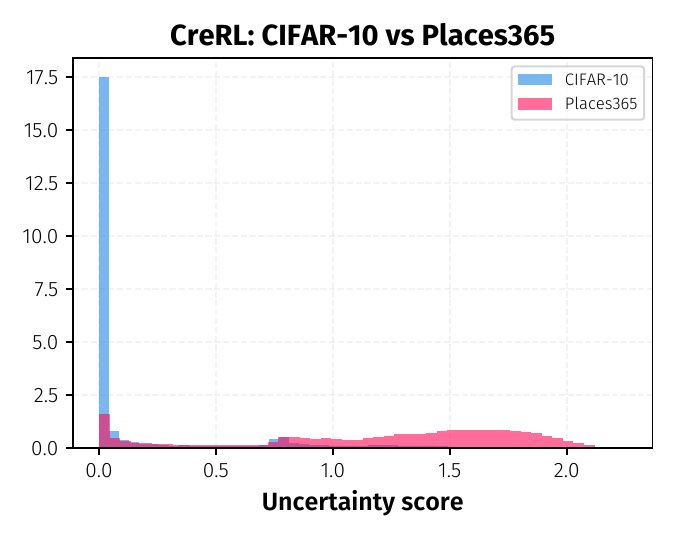}
\end{minipage}
\hfill
\begin{minipage}[c]{0.32\textwidth}
\includegraphics[width=\textwidth]{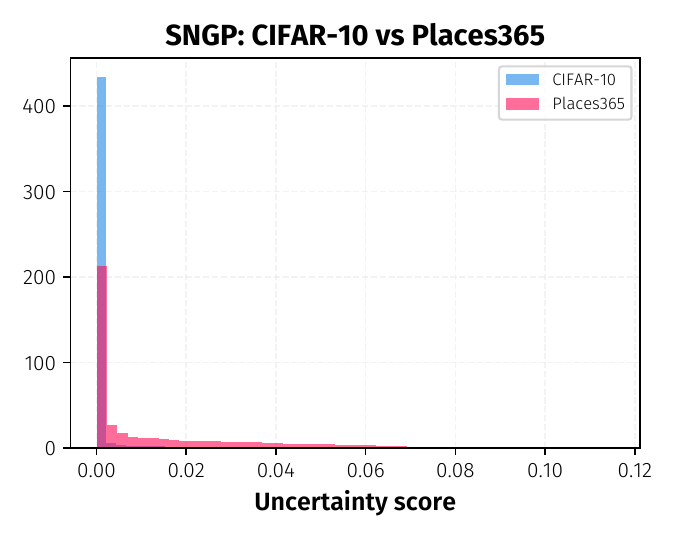}
\centering
\end{minipage}
\caption{A thorough evaluation across all methods in the \ac{ood} detection task for CIFAR-10. We compare on five \ac{ood} datasets of which a detailed analysis of three methods on the Places365 dataset is plotted in the histograms.}
\label{fig:cifar10-ood}
\end{figure}

\begin{figure}
\centering
\begin{minipage}[c]{0.49\textwidth}
\centering
\includegraphics[width=\linewidth]{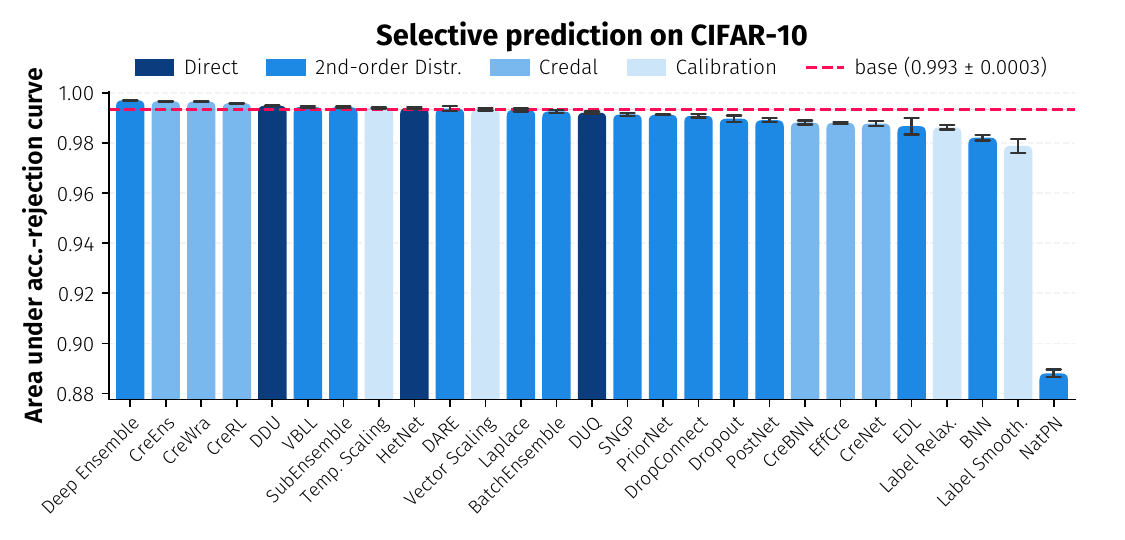}
\\
\includegraphics[width=0.75\linewidth]{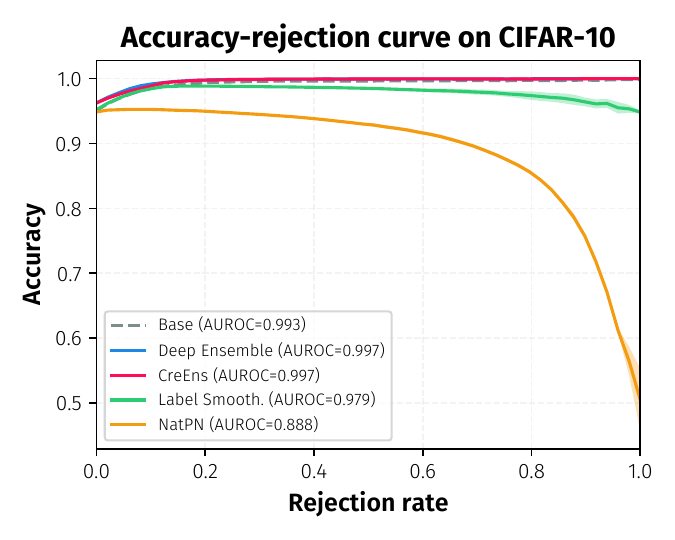}
\end{minipage}
\hfill
\begin{minipage}[c]{0.49\textwidth}
\centering
\includegraphics[width=\linewidth]{figures/appendix/sp/bar_sp_imagenet.pdf}
\\
\includegraphics[width=0.75\linewidth]{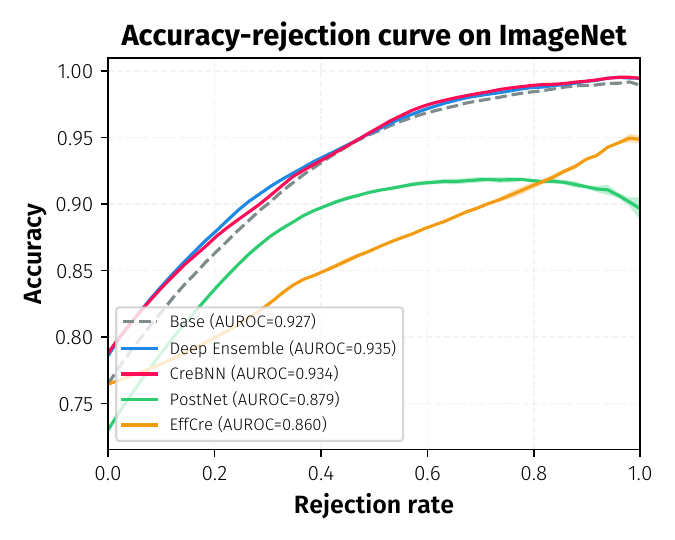}
\end{minipage}
\caption{Comparison of all methods on the selective prediction task for CIFAR-10 (left) and ImageNet (right). The arcs visualize the two best and two worst methods in this task.}
\label{fig:appendix/sp}
\end{figure}

This section presents supplemental results on additional datasets and settings
beyond the main body. For selective prediction (SP) and out-of-distribution
(OOD) detection, this includes evaluation on CIFAR-10
(see~\cref{tab:results-cifar10}; \cref{fig:cifar10-ood}) and a larger-scale
assessment on ImageNet (see~\cref{tab:results-imagenet};
\cref{fig:imagenet-ood}). Across these benchmarks, we evaluate 23 methods on
selective prediction and \ac{ood} detection. The extensive results for selective prediction are presented in \cref{fig:appendix/sp}.

\begin{table}[h]
\centering
\caption{Summary Results on ImageNet. For Selective prediction we report accuracy-rejection AUC (Acc-Rej) and for OoD AUROC.}
\label{tab:results-imagenet}
\footnotesize
\setlength{\tabcolsep}{7pt}
\renewcommand{\arraystretch}{1.15}
\begin{tabular}{@{}l@{\hspace{4pt}}l c c c@{}}
\toprule
\multicolumn{2}{@{}l}{\textbf{Method}} & \textbf{SP Acc-Rej} ($\uparrow$) & \textbf{Near-OoD AUROC} ($\uparrow$) & \textbf{Far-OoD AUROC} ($\uparrow$) \\
\hdashline\noalign{\vskip 0.45em}
\rowcolor[gray]{0.92}[0pt][0pt]
\multicolumn{5}{@{}l@{}}{\textbf{Deterministic baseline}} \\
\addlinespace[0.15em]
--- & Base & $.927_{\pm .000}$ & $.777_{\pm .002}$ & $.779_{\pm .003}$ \\
\addlinespace[0.4em]
\rowcolor[gray]{0.92}[0pt][0pt]
\multicolumn{5}{@{}l@{}}{\textbf{Direct}} \\
\addlinespace[0.15em]
\cite{mukhotiDDU2023} & DDU & $.928_{\pm .000}$ & $.681_{\pm .012}$ & $.798_{\pm .010}$ \\
\cite{collierHETNet2021} & HetNet & $.918_{\pm .001}$ & $.781_{\pm .010}$ & $.778_{\pm .013}$ \\
\addlinespace[0.4em]
\rowcolor[gray]{0.92}[0pt][0pt]
\multicolumn{5}{@{}l@{}}{\textbf{2nd-order Distribution}} \\
\addlinespace[0.15em]
\cite{galDropout2016} & Dropout & $.921_{\pm .000}$ & $.773_{\pm .003}$ & $.780_{\pm .003}$ \\
\cite{wanDropConnect2013} & DropConnect & $.919_{\pm .001}$ & $.765_{\pm .004}$ & $.773_{\pm .003}$ \\
\cite{blundellWeightUncertainty2015} & BNN & $.930_{\pm .000}$ & $.748_{\pm .003}$ & $.778_{\pm .001}$ \\
\cite{daxbergerLaplaceRedux2021} & Laplace & $.916_{\pm .001}$ & $.785_{\pm .006}$ & $.821_{\pm .005}$ \\
\cite{harrisonVBLL2024} & VBLL & $.917_{\pm .001}$ & $.683_{\pm .009}$ & $.708_{\pm .007}$ \\
\cite{laks_sa17} & Deep Ensemble & $\mathbf{.935_{\pm .000}}$ & $.783_{\pm .001}$ & $.820_{\pm .002}$ \\
\cite{wenBatchEnsemble2020} & BatchEnsemble & $.912_{\pm .000}$ & $.751_{\pm .005}$ & $.761_{\pm .001}$ \\
\cite{valdenegroToroSubEnsemble2023} & SubEnsemble & $.902_{\pm .000}$ & $.681_{\pm .001}$ & $.733_{\pm .002}$ \\
\cite{liuSNGP2020} & SNGP & $.892_{\pm .001}$ & $.752_{\pm .004}$ & $.767_{\pm .003}$ \\
\cite{sens_ed18} & EDL & $.907_{\pm .000}$ & $.735_{\pm .002}$ & $.743_{\pm .005}$ \\
\cite{charpentierPosteriorNetwork2020} & PostNet & $.879_{\pm .001}$ & $.547_{\pm .021}$ & $.564_{\pm .012}$ \\
\addlinespace[0.4em]
\rowcolor[gray]{0.92}[0pt][0pt]
\multicolumn{5}{@{}l@{}}{\textbf{Credal}} \\
\addlinespace[0.15em]
\cite{wangCredalWrapper2025} & CreWra & $.930_{\pm .000}$ & $.800_{\pm .001}$ & $\mathbf{.828_{\pm .001}}$ \\
\cite{nguyenCredalEnsembling2025} & CreEns & $.931_{\pm .000}$ & $.800_{\pm .001}$ & $.825_{\pm .002}$ \\
\cite{caprioCredalBNNs2024} & CreBNN & $.934_{\pm .000}$ & $.800_{\pm .001}$ & $.823_{\pm .001}$ \\
\cite{wangCredalIntervalNNs2025} & CreNet & $.921_{\pm .000}$ & $.754_{\pm .006}$ & $.755_{\pm .004}$ \\
\cite{loehrCredalPrediction2025} & CreRL & $.927_{\pm .000}$ & $\mathbf{.806_{\pm .001}}$ & $.827_{\pm .001}$ \\
\cite{hofmanEfficientCredalPrediction2026} & EffCre & $.860_{\pm .001}$ & $.644_{\pm .003}$ & $.678_{\pm .002}$ \\
\addlinespace[0.4em]
\rowcolor[gray]{0.92}[0pt][0pt]
\multicolumn{5}{@{}l@{}}{\textbf{Calibration}} \\
\addlinespace[0.15em]
\cite{szegedyRethinkingInceptionArchitecture2016} & Label Smooth. & $.905_{\pm .001}$ & $.778_{\pm .002}$ & $.777_{\pm .002}$ \\
\cite{DBLP:conf/aaai/LienenH21} & Label Relax. & $.927_{\pm .000}$ & $.780_{\pm .001}$ & $.777_{\pm .002}$ \\
\cite{guoOnCalibration2017} & Vector Scaling & $.887_{\pm .001}$ & $.726_{\pm .006}$ & $.738_{\pm .010}$ \\
\cite{guoOnCalibration2017} & Temp. Scaling & $.914_{\pm .000}$ & $.788_{\pm .001}$ & $.803_{\pm .003}$ \\
\bottomrule
\end{tabular}
\end{table}

\begin{table}[ht]
\centering
\caption{Summary Results on CIFAR-10. For Selective prediction we report accuracy-rejection AUC (Acc-Rej) and for OoD AUROC.}
\label{tab:results-cifar10}
\footnotesize
\setlength{\tabcolsep}{7pt}
\renewcommand{\arraystretch}{1.15}
\begin{tabular}{@{}l@{\hspace{4pt}}l c c c@{}}
\toprule
\multicolumn{2}{@{}l}{\textbf{Method}} & \textbf{SP Acc-Rej} ($\uparrow$) & \textbf{Near-OoD AUROC} ($\uparrow$) & \textbf{Far-OoD AUROC} ($\uparrow$) \\
\hdashline\noalign{\vskip 0.45em}
\rowcolor[gray]{0.92}[0pt][0pt]
\multicolumn{5}{@{}l@{}}{\textbf{Deterministic baseline}} \\
\addlinespace[0.15em]
--- & Base & $.993_{\pm .000}$ & $.872_{\pm .001}$ & $.912_{\pm .014}$ \\
\addlinespace[0.4em]
\rowcolor[gray]{0.92}[0pt][0pt]
\multicolumn{5}{@{}l@{}}{\textbf{Direct}} \\
\addlinespace[0.15em]
\cite{mukhotiDDU2023} & DDU & $.995_{\pm .000}$ & $.899_{\pm .001}$ & $.935_{\pm .004}$ \\
\cite{vanAmersfoortDUE2021} & DUQ & $.992_{\pm .000}$ & $.857_{\pm .003}$ & $.902_{\pm .010}$ \\
\cite{collierHETNet2021} & HetNet & $.994_{\pm .000}$ & $.871_{\pm .003}$ & $.906_{\pm .007}$ \\
\addlinespace[0.4em]
\rowcolor[gray]{0.92}[0pt][0pt]
\multicolumn{5}{@{}l@{}}{\textbf{2nd-order Distribution}} \\
\addlinespace[0.15em]
\cite{galDropout2016} & Dropout & $.990_{\pm .001}$ & $.838_{\pm .002}$ & $.894_{\pm .005}$ \\
\cite{wanDropConnect2013} & DropConnect & $.991_{\pm .001}$ & $.830_{\pm .002}$ & $.874_{\pm .005}$ \\
\cite{blundellWeightUncertainty2015} & BNN & $.982_{\pm .001}$ & $.828_{\pm .005}$ & $.835_{\pm .015}$ \\
\cite{daxbergerLaplaceRedux2021} & Laplace & $.993_{\pm .001}$ & $.868_{\pm .000}$ & $.901_{\pm .011}$ \\
\cite{harrisonVBLL2024} & VBLL & $.994_{\pm .000}$ & $.812_{\pm .002}$ & $.861_{\pm .011}$ \\
\cite{laks_sa17} & Deep Ensemble & $\mathbf{.997_{\pm .000}}$ & $.901_{\pm .001}$ & $.928_{\pm .003}$ \\
\cite{wenBatchEnsemble2020} & BatchEnsemble & $.993_{\pm .001}$ & $.875_{\pm .002}$ & $.905_{\pm .005}$ \\
\cite{valdenegroToroSubEnsemble2023} & SubEnsemble & $.994_{\pm .000}$ & $.502_{\pm .014}$ & $.498_{\pm .012}$ \\
\cite{deMathelinDARE2023} & DARE & $.994_{\pm .001}$ & $.879_{\pm .003}$ & $.914_{\pm .007}$ \\
\cite{liuSNGP2020} & SNGP & $.991_{\pm .001}$ & $.847_{\pm .003}$ & $.884_{\pm .009}$ \\
\cite{sens_ed18} & EDL & $.987_{\pm .003}$ & $.861_{\pm .007}$ & $.890_{\pm .022}$ \\
\cite{charpentierPosteriorNetwork2020} & PostNet & $.989_{\pm .001}$ & $.837_{\pm .002}$ & $.887_{\pm .008}$ \\
\cite{charpentierNaturalPosteriorNetwork2022} & NatPN & $.888_{\pm .001}$ & $.256_{\pm .010}$ & $.220_{\pm .020}$ \\
\addlinespace[0.4em]
\rowcolor[gray]{0.92}[0pt][0pt]
\multicolumn{5}{@{}l@{}}{\textbf{Credal}} \\
\addlinespace[0.15em]
\cite{wangCredalWrapper2025} & CreWra & $\mathbf{.997_{\pm .000}}$ & $\mathbf{.918_{\pm .001}}$ & $\mathbf{.952_{\pm .002}}$ \\
\cite{nguyenCredalEnsembling2025} & CreEns & $\mathbf{.997_{\pm .000}}$ & $\mathbf{.918_{\pm .001}}$ & $.951_{\pm .002}$ \\
\cite{caprioCredalBNNs2024} & CreBNN & $.988_{\pm .001}$ & $.872_{\pm .004}$ & $.898_{\pm .010}$ \\
\cite{wangCredalIntervalNNs2025} & CreNet & $.988_{\pm .001}$ & $.753_{\pm .023}$ & $.812_{\pm .037}$ \\
\cite{loehrCredalPrediction2025} & CreRL & $.996_{\pm .000}$ & $.917_{\pm .000}$ & $.951_{\pm .003}$ \\
\cite{hofmanEfficientCredalPrediction2026} & EffCre & $.988_{\pm .000}$ & $.616_{\pm .010}$ & $.600_{\pm .018}$ \\
\addlinespace[0.4em]
\rowcolor[gray]{0.92}[0pt][0pt]
\multicolumn{5}{@{}l@{}}{\textbf{Calibration}} \\
\addlinespace[0.15em]
\cite{szegedyRethinkingInceptionArchitecture2016} & Label Smooth. & $.979_{\pm .003}$ & $.767_{\pm .011}$ & $.830_{\pm .016}$ \\
\cite{DBLP:conf/aaai/LienenH21} & Label Relax. & $.986_{\pm .001}$ & $.806_{\pm .002}$ & $.875_{\pm .015}$ \\
\cite{guoOnCalibration2017} & Vector Scaling & $.993_{\pm .001}$ & $.865_{\pm .002}$ & $.904_{\pm .011}$ \\
\cite{guoOnCalibration2017} & Temp. Scaling & $.994_{\pm .000}$ & $.874_{\pm .002}$ & $.919_{\pm .007}$ \\
\bottomrule
\end{tabular}
\end{table}

\clearpage
\subsection{Active Learning}
\begin{figure}[ht]
    \centering
    \begin{minipage}[c]{0.75\linewidth}
        \includegraphics[width=\linewidth]{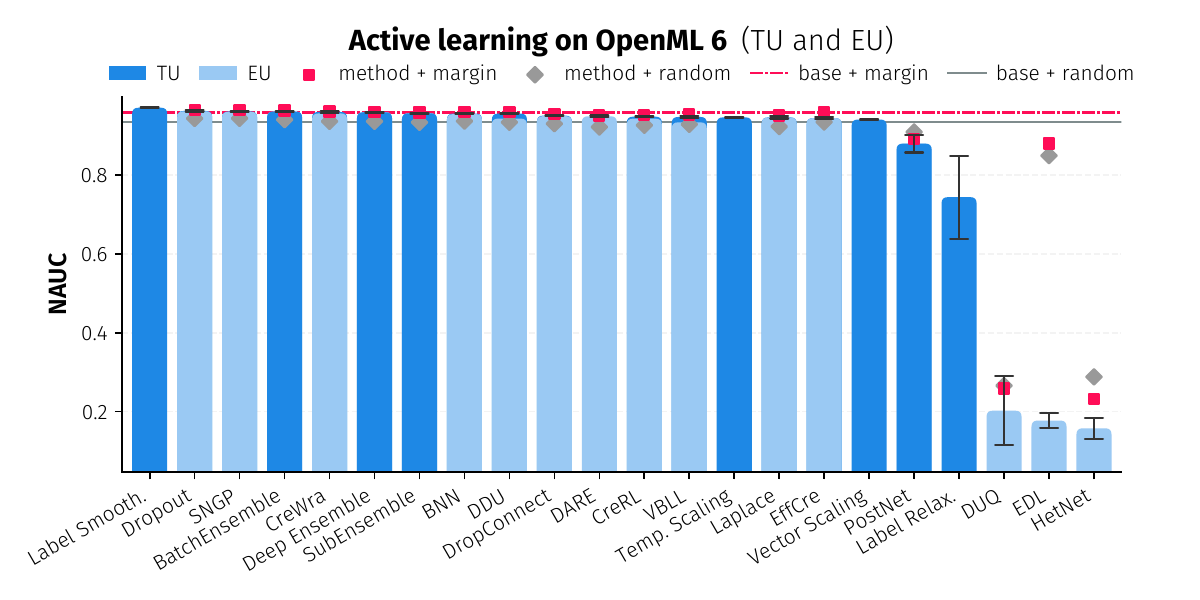}
    \end{minipage}
    \\
    \hfill
    \begin{minipage}[c]{0.32\linewidth}
        \includegraphics[width=\linewidth]{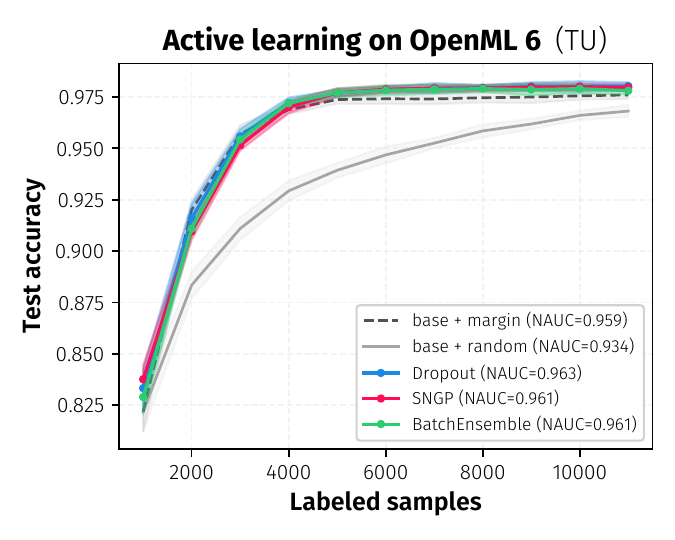}
    \end{minipage}
    \begin{minipage}[c]{0.32\linewidth}
        \includegraphics[width=\linewidth]{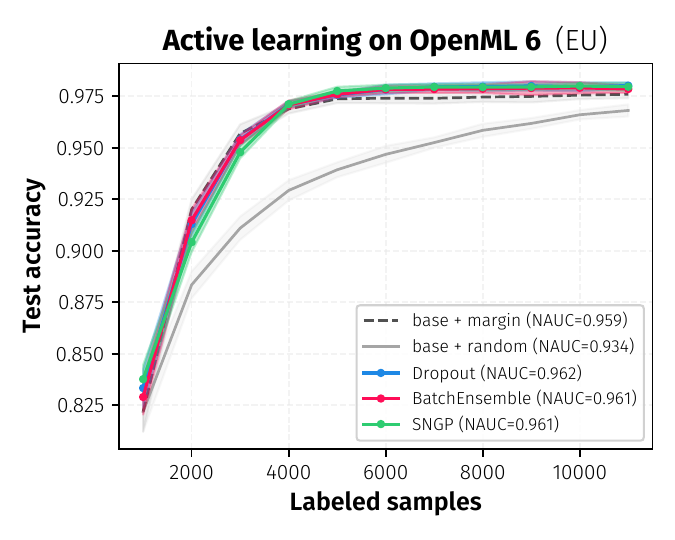}
    \end{minipage}
    \begin{minipage}[c]{0.32\linewidth}
        \includegraphics[width=\linewidth]{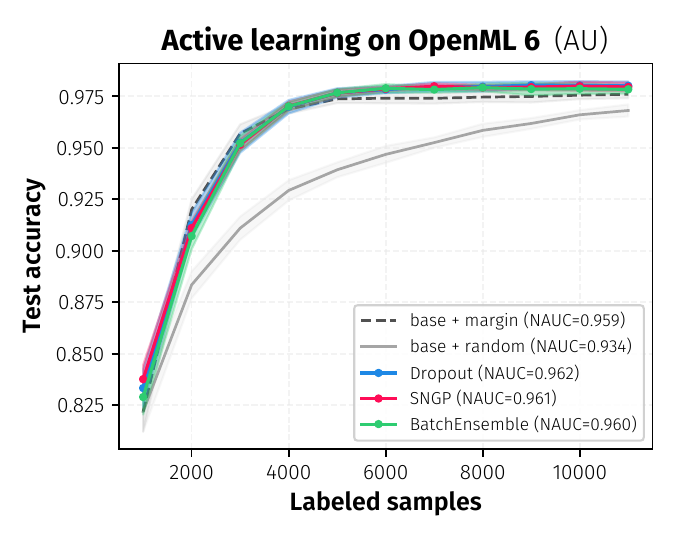}
    \end{minipage}
    \hfill
    \caption{Mean normalized area under the budget curve (NAUC, $\uparrow$) \textbf{(top)} and budget performance plot on OpenML-6 for the five best-performing methods for each notion of uncertainty \textbf{(bottom)}.}
\label{fig:al_results:openml_6}
\vspace{-0.5cm}
\end{figure}

\begin{figure}[h]
    \centering
    \begin{minipage}[c]{0.75\linewidth}
        \includegraphics[width=\linewidth]{figures/appendix/al/bar_al_eu_tu_openml_156.pdf}
    \end{minipage}
    \\
    \hfill
    \begin{minipage}[c]{0.32\linewidth}
        \includegraphics[width=\linewidth]{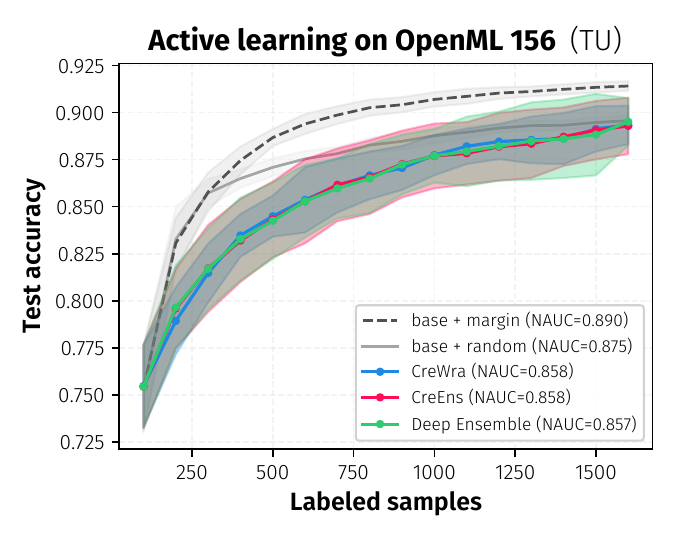}
    \end{minipage}
    \begin{minipage}[c]{0.32\linewidth}
        \includegraphics[width=\linewidth]{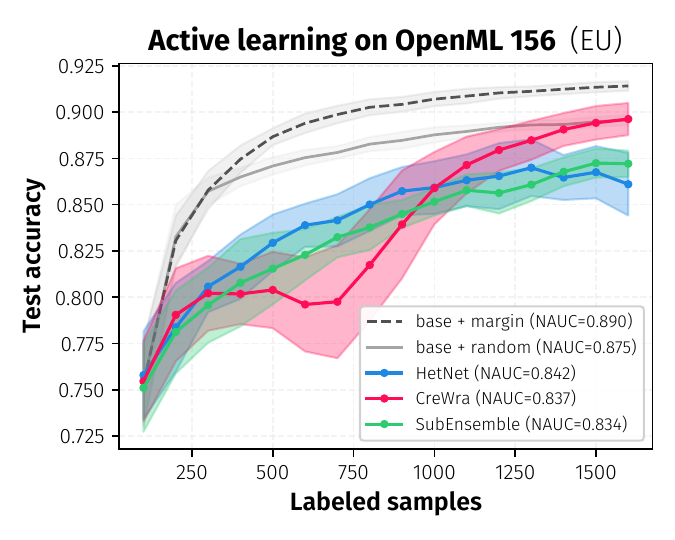}
    \end{minipage}
    \begin{minipage}[c]{0.32\linewidth}
        \includegraphics[width=\linewidth]{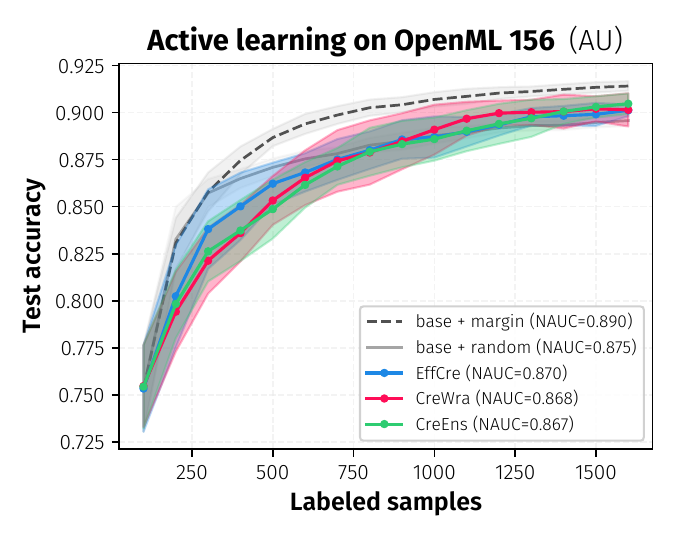}
    \end{minipage}
    \hfill
    \caption{Mean normalized area under the budget curve (NAUC, $\uparrow$) \textbf{(top)} and budget performance plot on OpenML-156 for the five best-performing methods for each notion of uncertainty \textbf{(bottom)}.}
\label{fig:al_results:openml_156}
\vspace{-0.5cm}
\end{figure}

\begin{figure}[h]
    \centering
    \begin{minipage}[c]{0.75\linewidth}
        \includegraphics[width=\linewidth]{figures/appendix/al/bar_al_eu_tu_cifar10.pdf}
    \end{minipage}
    \\
    \begin{minipage}[c]{0.32\linewidth}
        \includegraphics[width=\linewidth]{figures/appendix/al/al_curves_one_col_cifar10_total.pdf}
    \end{minipage}
    \hspace{0.02\linewidth}
    \begin{minipage}[c]{0.32\linewidth}
        \includegraphics[width=\linewidth]{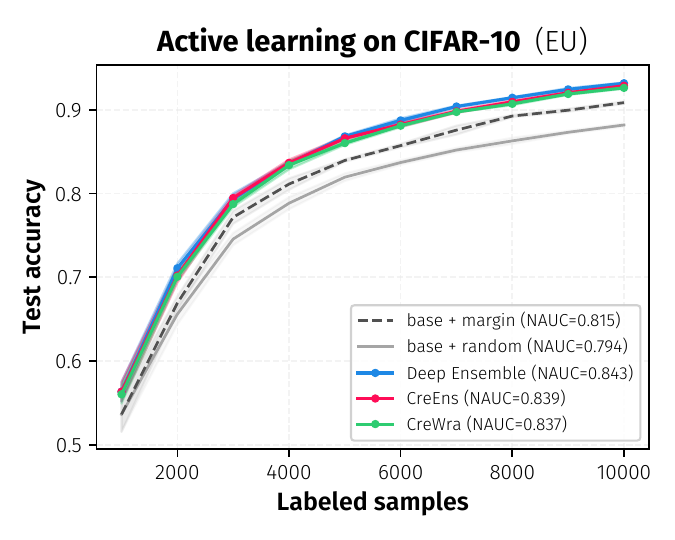}
    \end{minipage}
    \caption{Mean normalized area under the budget curve (NAUC, $\uparrow$) \textbf{(top)} and budget performance plot on CIFAR-10 for the three best-performing methods for total and epistemic uncertainty \textbf{(bottom)}.}
\label{fig:al_results:cifar10}
\vspace{-0.5cm}
\end{figure}

Besides \ac{ood} detection and selective prediction, active learning is another standard downstream evaluation task for uncertainty methods. We present an extensive evaluation and comparison to state-of-the-art baselines for two common benchmark datasets from the OpenML benchmark suite and for CIFAR-10 in \cref{tab:results-al}, respectively. A more visual comparison for the two tabular datasets is presented in \cref{fig:al_results:openml_6} and \cref{fig:al_results:openml_156}, and for CIFAR-10 in \cref{fig:al_results:cifar10}.

\begin{table}[t]
\centering
\caption{Active learning (NAUC). LS denotes Label Smoothing, LR denotes Label Relaxation, temp denotes temperature scaling, and vec vector scaling. A superscript names the notion a value comes from when a method has no run for the column's notion.}
\label{tab:results-al}
\footnotesize
\setlength{\tabcolsep}{6pt}
\renewcommand{\arraystretch}{1.15}
\begin{tabular}{@{}l cc cc cc@{}}
\toprule
\multicolumn{1}{@{}l}{} & \multicolumn{2}{c}{\textbf{OpenML 6}} & \multicolumn{2}{c}{\textbf{OpenML 156}} & \multicolumn{2}{c}{\textbf{CIFAR-10}} \\
\cdashline{2-3} \cdashline{4-5} \cdashline{6-7}
\noalign{\vskip 0.45em}
\multicolumn{1}{@{}l}{\textbf{Method}} & \textbf{EU} & \textbf{TU} & \textbf{EU} & \textbf{TU} & \textbf{EU} & \textbf{TU} \\
\hdashline\noalign{\vskip 0.45em}
\rowcolor[gray]{0.92}[0pt][0pt]
\multicolumn{7}{@{}l@{}}{\textbf{Baselines}} \\
\addlinespace[0.15em]
base + margin & $.959_{\pm .002}$ & $.959_{\pm .002}$ & $.890_{\pm .004}$ & $.890_{\pm .004}$ & $.815_{\pm .003}$ & $.815_{\pm .003}$ \\
base + random & $.934_{\pm .003}$ & $.934_{\pm .003}$ & $.875_{\pm .004}$ & $.875_{\pm .004}$ & $.794_{\pm .002}$ & $.794_{\pm .002}$ \\
\addlinespace[0.4em]
\rowcolor[gray]{0.92}[0pt][0pt]
\multicolumn{7}{@{}l@{}}{\textbf{Direct}} \\
\addlinespace[0.15em]
DDU & $.943_{\pm .004}$ & $.956_{\pm .001}$ & $.693_{\pm .017}$ & $.839_{\pm .015}$ & $.814_{\pm .002}$ & $.819_{\pm .003}^{\mathrm{AU}}$ \\
DUQ & $.202_{\pm .088}$ & $.202_{\pm .088}$ & $.803_{\pm .016}$ & $.803_{\pm .016}$ & $.824_{\pm .002}^{\mathrm{TU}}$ & $.824_{\pm .002}$ \\
HetNet & $.157_{\pm .026}$ & $.157_{\pm .026}$ & $.842_{\pm .010}$ & $.842_{\pm .010}$ & $.800_{\pm .003}^{\mathrm{AU}}$ & $.800_{\pm .003}^{\mathrm{AU}}$ \\
\addlinespace[0.4em]
\rowcolor[gray]{0.92}[0pt][0pt]
\multicolumn{7}{@{}l@{}}{\textbf{2nd-order Distribution}} \\
\addlinespace[0.15em]
Dropout & $.962_{\pm .002}$ & $.963_{\pm .001}$ & $.655_{\pm .015}$ & $.822_{\pm .015}$ & $.817_{\pm .002}$ & $.817_{\pm .003}$ \\
DropConnect & $.949_{\pm .002}$ & $.951_{\pm .002}$ & $.552_{\pm .014}$ & $.716_{\pm .030}$ & $.813_{\pm .002}$ & $.816_{\pm .004}$ \\
BNN & $.955_{\pm .001}$ & $.956_{\pm .002}$ & $.316_{\pm .012}$ & $.435_{\pm .024}$ & $.783_{\pm .004}$ & $.784_{\pm .001}$ \\
Laplace & $.946_{\pm .002}$ & $.946_{\pm .003}$ & $.574_{\pm .042}$ & $.800_{\pm .025}$ & $.753_{\pm .018}$ & $.775_{\pm .007}$ \\
VBLL & $.934_{\pm .004}$ & $.947_{\pm .002}$ & $.818_{\pm .019}$ & $.779_{\pm .023}$ & $.824_{\pm .001}$ & $.822_{\pm .003}$ \\
Deep Ensemble & $.960_{\pm .002}$ & $.959_{\pm .002}$ & $.783_{\pm .015}$ & $.857_{\pm .016}$ & $\mathbf{.843_{\pm .002}}$ & $\mathbf{.843_{\pm .002}}$ \\
BatchEnsemble & $.961_{\pm .001}$ & $.961_{\pm .001}$ & $.690_{\pm .016}$ & $.844_{\pm .007}$ & $.821_{\pm .003}$ & $.818_{\pm .004}$ \\
SubEnsemble & $.955_{\pm .002}$ & $.956_{\pm .001}$ & $.847_{\pm .008}$ & $.864_{\pm .007}$ & $.808_{\pm .004}$ & $.805_{\pm .002}$ \\
DARE & $.949_{\pm .002}$ & $.949_{\pm .002}^{\mathrm{AU}}$ & $.802_{\pm .016}$ & $.802_{\pm .016}^{\mathrm{AU}}$ & $.770_{\pm .013}$ & $.770_{\pm .013}^{\mathrm{EU}}$ \\
SNGP & $.286_{\pm .074}$ & $.241_{\pm .042}$ & $.649_{\pm .003}$ & $.667_{\pm .004}$ & $.739_{\pm .020}$ & $.739_{\pm .020}^{\mathrm{EU}}$ \\
EDL & $.177_{\pm .019}$ & $.177_{\pm .019}$ & $.606_{\pm .037}$ & $.606_{\pm .037}$ & $.622_{\pm .017}$ & $.622_{\pm .017}^{\mathrm{EU}}$ \\
PostNet & $.880_{\pm .027}$ & $.880_{\pm .022}$ & $.390_{\pm .030}$ & $.391_{\pm .033}$ & $.795_{\pm .004}$ & $.796_{\pm .006}^{\mathrm{AU}}$ \\
\addlinespace[0.4em]
\rowcolor[gray]{0.92}[0pt][0pt]
\multicolumn{7}{@{}l@{}}{\textbf{Credal}} \\
\addlinespace[0.15em]
CreWra & $.956_{\pm .001}$ & $.960_{\pm .002}$ & $.837_{\pm .015}$ & $.858_{\pm .009}$ & $.837_{\pm .002}$ & $.835_{\pm .003}$ \\
CreEns & -- & -- & $.808_{\pm .021}$ & $.858_{\pm .016}$ & $.839_{\pm .002}$ & $.840_{\pm .003}$ \\
CreBNN & -- & -- & -- & -- & $.813_{\pm .003}$ & $.814_{\pm .003}$ \\
CreRL & $.943_{\pm .002}$ & $.948_{\pm .002}$ & $.480_{\pm .014}$ & $.510_{\pm .026}$ & $.813_{\pm .003}$ & $.811_{\pm .003}$ \\
EffCre & $.944_{\pm .002}$ & $.944_{\pm .002}$ & $.830_{\pm .015}$ & $.857_{\pm .011}$ & $.800_{\pm .004}$ & $.813_{\pm .005}$ \\
\addlinespace[0.4em]
\rowcolor[gray]{0.92}[0pt][0pt]
\multicolumn{7}{@{}l@{}}{\textbf{Calibration}} \\
\addlinespace[0.15em]
base + margin + LS & $\mathbf{.971_{\pm .001}}$ & $\mathbf{.971_{\pm .001}}$ & $\mathbf{.893_{\pm .004}}$ & $\mathbf{.893_{\pm .004}}$ & $.819_{\pm .003}$ & $.819_{\pm .003}$ \\
base + margin + LR & $.744_{\pm .105}$ & $.744_{\pm .105}$ & $.885_{\pm .008}$ & $.885_{\pm .008}$ & $.814_{\pm .002}$ & $.814_{\pm .002}$ \\
base + margin + temp & $.946_{\pm .001}$ & $.946_{\pm .001}$ & $.870_{\pm .006}$ & $.870_{\pm .006}$ & $.800_{\pm .003}$ & $.800_{\pm .003}$ \\
base + margin + vec & $.941_{\pm .001}$ & $.941_{\pm .001}$ & $.864_{\pm .006}$ & $.864_{\pm .006}$ & $.796_{\pm .002}$ & $.796_{\pm .002}$ \\
\bottomrule
\end{tabular}
\end{table}

\clearpage
\subsection{First-Order Data}
\textbf{Credal sets evaluation.}
A fundamental property of a credal set is whether it contains the true conditional label distribution. Given first-order data, this can be assessed directly by checking whether $\prob( \cdot \given \vx)$ lies within the predicted credal set $Q$ for each input $\vx$, i.e., 
$ \mathds{1}\big( \prob( \cdot \given \vx)\in Q\big)$.
Such evaluation of credal sets has been considered in recent works \cite{javanmardi2024conformalized, caprio2025conformalized, loehrCredalPrediction2025,hofmanEfficientCredalPrediction2026}. 
\begin{itemize}
    \item \textbf{ConvexHull coverage:}
Given a credal set described by an ensemble of $M$ label distributions $\{\pi_m(\cdot\given\vx)\}_{m=1}^M$, the convex hull coverage checks whether the true distribution lies in the convex hull of these $M$ distributions, i.e.,
\begin{equation}\label{eq:cov:hull}
\mathds{1}\left[ \prob( \cdot \given \vx)\in \operatorname{convexHull}\left(\{\pi_m(\cdot\given\vx)\}_{m=1}^M\right) \right].
\end{equation}
In practice, this is evaluated via an LP feasibility test, i.e., whether there exist convex weights $\boldsymbol{\lambda} \in \Delta_M$ such that $\sum_m \lambda_m \pi_m(\cdot \mid \vx_i) = p(\cdot \mid \vx_i)$ (up to a tolerance $\varepsilon = 5\times 10^{-3}$).

\item \textbf{Interval coverage:}
Interval coverage uses the component-wise envelope $\underline{\pi}_k(\vx) = \min_m \pi_m(k \mid \vx)$ and $\overline{\pi}_k(\vx) = \max_m \pi_m(k \mid \vx)$, and checks whether the true distribution lies within this range for all classes simultaneously:
\begin{equation}\label{eq:cov:interval}
\mathds{1}\left[\forall k:\ \underline{\pi}_k(\vx_i) \leq p(k \mid \vx) \leq \overline{\pi}_k(\vx)\right].
\end{equation}
The same precision tolerance is used here to ensure stable results. Note that the interval coverage is lower bounded by the convex hull coverage.
\item \textbf{Efficiency:}
For a given credal set, efficiency measures the average tightness of the interval envelope, with higher values indicating a more informative credal set:
\begin{equation}\label{eq:eff}
1 - \frac{1}{K} \sum_{k=1}^K \bigl(\overline{\pi}_k(\vx) - \underline{\pi}_k(\vx)\bigr).
\end{equation}
\end{itemize}

All metrics are computed for each test input, averaged across the dataset, and reported in \cref{tab:credal_coverage} accordingly.
\begin{table}[t]
\centering
\caption{Credal set coverage and efficiency.}
\footnotesize
\label{tab:credal_coverage}
\setlength{\tabcolsep}{4pt}
\renewcommand{\arraystretch}{1.15}
\begin{tabular}{@{}l@{\hspace{2pt}}cccc@{}}
\toprule
\multicolumn{2}{@{}l}{\textbf{Method}}
&\makecell[c]{\textbf{ConvexHull}\\[-1pt]\textbf{Coverage ($\uparrow$)}}
&\makecell[c]{\textbf{Interval}\\[-1pt]\textbf{Coverage ($\uparrow$)}}
&\textbf{Efficiency} \textbf{($\uparrow$)}\\
\hdashline\noalign{\vskip0.45em}
\rowcolor[gray]{0.92}[0pt][0pt]
\multicolumn{5}{@{}l@{}}{\textbf{micebone}}\\
\addlinespace[0.15em]
CreEns&&$0.031$&$0.036$&$\mathbf{0.877}$\\
CreRL&&$\mathbf{0.035}$&$\mathbf{0.045}$&$0.843$\\
CreWra&&$0.023$&$0.029$&$0.862$\\
\addlinespace[0.4em]
\rowcolor[gray]{0.92}[0pt][0pt]
\multicolumn{5}{@{}l@{}}{\textbf{qualitymri}}\\
\addlinespace[0.15em]
CreEns&&$0.267$&$0.267$&$0.870$\\
CreRL&&$\mathbf{0.833}$&$\mathbf{0.833}$&$0.331$\\
CreWra&&$0.233$&$0.233$&$\mathbf{0.874}$\\
\addlinespace[0.4em]
\rowcolor[gray]{0.92}[0pt][0pt]
\multicolumn{5}{@{}l@{}}{\textbf{synthetic}}\\
\addlinespace[0.15em]
CreEns&&$\mathbf{0.230}$&$\mathbf{0.305}$&$0.880$\\
CreRL&&$0.049$&$0.084$&$\mathbf{0.912}$\\
CreWra&&$0.213$&$0.296$&$0.880$\\
\bottomrule
\end{tabular}
\end{table}
\begin{table}[h]
\centering
\vspace{-1.2em}
\caption{Calibration error: TV distance (ensemble mean vs.\ true distribution).}
\footnotesize
\label{tab:tv_calibration}
\setlength{\tabcolsep}{4pt}
\renewcommand{\arraystretch}{1.15}
\begin{tabular}{@{}lccc@{}}
\toprule
\textbf{Dataset} & \textbf{CreEns} ($\downarrow$) & \textbf{CreRL} ($\downarrow$) & \textbf{CreWra} ($\downarrow$) \\
\hdashline\noalign{\vskip 0.45em}
\midrule
micebone   & $\mathbf{.219_{\pm .213}}$ & $.239_{\pm .209}$ & $.220_{\pm .213}$ \\
qualitymri & $\mathbf{.172_{\pm .140}}$ & $.207_{\pm .165}$ & $\mathbf{.171_{\pm .137}}$ \\
synthetic  & $\mathbf{.197_{\pm .121}}$ & $.298_{\pm .153}$ & $.201_{\pm .121}$ \\
\bottomrule
\end{tabular}
\end{table}

\paragraph{Conditional calibration error evaluation.} In the context of probabilistic prediction, calibration refers to the extent to which predicted probabilities align with observed frequencies. In the standard (zero-order) setting, calibration is typically assessed in an aggregate sense, for example, through metrics such as the expected calibration error (ECE). Access to the conditional label distribution $\prob( \cdot \given \vx)$ enables a direct, pointwise evaluation of calibration. Specifically, for a probabilistic predictor $\prob( \cdot \given \vx, h)$, one can compare the predicted distribution to the true conditional distribution at each input $\vx$, for instance via the total variation distance
$
d_{\mathrm{TV}}\bigl(\prob( \cdot \given \vx), \prob( \cdot \given \vx, h)\bigr).
$ \cref{tab:tv_calibration} provides such a comparison for different classifiers (each being the mean of the provided ensemble).

\paragraph{Conformal Prediction}
We apply split-conformal prediction to each model on the test fold. Since the DCIC test set provides first-order annotations, we sample a single hard label $\tilde{y} \sim p(\cdot \mid \vx)$ per instance to obtain a zero-order label. We then randomly partition the test instances into two equally sized halves: one for conformal calibration and the other for evaluation.
Calibration is performed on the sampled labels using four score functions, LAC, APS, RAPS, and SAPS, with target miscoverage level $\alpha = 0.1$. For each test input, we report coverage, defined as whether the sampled label is contained in the prediction set $\cC(\vx)$, and set size as the cardinality of $\cC$.
We also report conditional satisfaction, defined as the indicator of whether the accumulated probability mass exceeds $1 - \alpha$:
$$
\mathds{1}\left[\sum_{k \in \cC(\vx)} p(k \mid \vx) \geq 1 - \alpha\right].
$$
The three measures are computed and averaged over the evaluation set. This process is repeated over 10 different random seeds, and the mean and standard deviation across seeds are reported in the \cref{tab:cp}.

Overall, the presented results highlight the fact that first-order data can bring many insights, for instance in evaluating credal set predictors in terms of covering the true label distributions, highlighting the gap between marginal and conditional coverage in conformal prediction, as well as providing a measure of how well models perform in predicting the true label distributions, opening room for a fair and careful comparison of different baselines while also providing insights into how to improve them.

\subsection{Ablation on Uncertainty Measures}

The benchmarks in \cref{sec:bench-sp} and \cref{sec:bench-ood} evaluate every method with the uncertainty measure proposed in the paper that introduced it. For the sample-based methods, this is the entropy decomposition \cref{eq:entropy-decomposition}. This raises the question of how much the choice of measure as opposed to the choice of method, drives the results on the downstream tasks. We therefore repeat the evaluation for all methods whose representation is a sample of categorical distributions and exchange the entropy decomposition for two decompositions induced by other proper scoring rules. Both share the structure of \cref{eq:entropy-decomposition}. The total uncertainty of the averaged prediction $\bar{p}(\cdot \mid \vx)$ splits into the average uncertainty of the individual predictors (aleatoric) and the average divergence of the individual predictors from their mean (epistemic). The zero-one decomposition measures uncertainty by the expected zero-one loss of the most likely class, i.e. one minus the maximum predicted probability,
\begin{equation*}
    1 - \max_y \hat{p}(y \mid \vx) = \frac{1}{M}\sum_{m=1}^M \Big(1 - \max_y p(y \mid \vx, h_m)\Big) + \frac{1}{M}\sum_{m=1}^M \Big(\max_y p(y \mid \vx, h_m) - p(\hat{y} \mid \vx, h_m)\Big),
\end{equation*}
where $\hat{y} = \argmax_y \hat{p}(y \mid \vx)$ is the prediction of the averaged model.

The Brier decomposition replaces the entropy by $1 - \sum_y p(y)^2$, and the divergence by the squared Euclidean distance,
\begin{equation*}
    1 - \sum_y \hat{p}(y \mid \vx)^2 = \frac{1}{M}\sum_{m=1}^M \Big(1 - \sum_y p(y \mid \vx, h_m)^2\Big) + \sum_y \frac{1}{M}\sum_{m=1}^M \big(p(y \mid \vx, h_m) - \hat{p}(y \mid \vx)\big)^2 .
\end{equation*}
As in the main benchmark, selective prediction ranks instances by total uncertainty and \ac{ood} detection by epistemic uncertainty. Models, seeds, and evaluation data are identical across measures, so the rows of a method in \cref{tab:ablation-measure-cifar10,tab:ablation-measure-imagenet} differ only in the measure.

The measure matters differently for the two tasks. For selective prediction the three measures are nearly interchangeable on CIFAR-10 and ImageNet. For \ac{ood} detection the entropy decomposition yields the highest AUROC for almost every method and dataset on both benchmarks. The zero-one decomposition falls far behind because its epistemic term is exactly zero whenever all members agree on the most likely class, which is the case for most inputs, in and out of distribution, so the resulting score barely separates the two. The drop is largest for methods whose members rarely disagree, e.g. SubEnsemble, BatchEnsemble and SNGP. The Brier decomposition results are close to the entropy decomposition results for most methods on CIFAR-10, but lower for ImageNet. Hence, conclusions about which method quantifies epistemic uncertainty best are not invariant to the measure, and a benchmark should state the measure as explicitly as the method.

\begin{table}[t]
\centering
\caption{Ablation on the uncertainty measure on CIFAR-10. Selective prediction (SP) is reported as the area under the risk-coverage curve (AURC, lower is better). \ac{ood} detection is reported as the AUROC (higher is better) for each \ac{ood} dataset. Mean $\pm$ standard deviation over three seeds with the best measure per method and column in bold.}
\label{tab:ablation-measure-cifar10}
\footnotesize
\setlength{\tabcolsep}{3.5pt}
\begin{tabular}{llcccccc}
\toprule
 & & SP & \multicolumn{5}{c}{\ac{ood} detection, AUROC $\uparrow$} \\
\cmidrule(lr){3-3} \cmidrule(lr){4-8}
Method & Measure & AURC $\downarrow$ & CIFAR-100 & MNIST & SVHN & Textures & Places365 \\
\midrule
Deep Ensemble & Entropy & 0.30{\scriptsize$\,\pm\,$0.02} & \textbf{90.1{\scriptsize$\,\pm\,$0.1}} & \textbf{93.9{\scriptsize$\,\pm\,$0.4}} & \textbf{94.8{\scriptsize$\,\pm\,$0.4}} & \textbf{93.0{\scriptsize$\,\pm\,$0.2}} & \textbf{89.7{\scriptsize$\,\pm\,$0.2}} \\
 & Zero-one & 0.30{\scriptsize$\,\pm\,$0.01} & 79.9{\scriptsize$\,\pm\,$0.2} & 85.9{\scriptsize$\,\pm\,$1.5} & 87.7{\scriptsize$\,\pm\,$1.0} & 84.6{\scriptsize$\,\pm\,$0.5} & 79.6{\scriptsize$\,\pm\,$0.3} \\
 & Brier & \textbf{0.30{\scriptsize$\,\pm\,$0.02}} & 89.3{\scriptsize$\,\pm\,$0.1} & 93.0{\scriptsize$\,\pm\,$0.3} & 93.8{\scriptsize$\,\pm\,$0.3} & 92.1{\scriptsize$\,\pm\,$0.1} & 88.9{\scriptsize$\,\pm\,$0.2} \\
\midrule
SubEnsemble & Entropy & 0.58{\scriptsize$\,\pm\,$0.04} & 50.2{\scriptsize$\,\pm\,$1.7} & 49.3{\scriptsize$\,\pm\,$1.0} & 50.2{\scriptsize$\,\pm\,$1.8} & 49.8{\scriptsize$\,\pm\,$1.5} & 50.0{\scriptsize$\,\pm\,$1.5} \\
 & Zero-one & 0.62{\scriptsize$\,\pm\,$0.05} & 50.4{\scriptsize$\,\pm\,$0.1} & 50.5{\scriptsize$\,\pm\,$0.1} & 50.6{\scriptsize$\,\pm\,$0.2} & 50.5{\scriptsize$\,\pm\,$0.1} & 50.4{\scriptsize$\,\pm\,$0.0} \\
 & Brier & \textbf{0.57{\scriptsize$\,\pm\,$0.04}} & \textbf{54.0{\scriptsize$\,\pm\,$0.3}} & \textbf{53.4{\scriptsize$\,\pm\,$1.5}} & \textbf{54.4{\scriptsize$\,\pm\,$1.2}} & \textbf{53.8{\scriptsize$\,\pm\,$0.8}} & \textbf{53.5{\scriptsize$\,\pm\,$1.3}} \\
\midrule
BatchEnsemble & Entropy & 0.74{\scriptsize$\,\pm\,$0.06} & \textbf{87.5{\scriptsize$\,\pm\,$0.2}} & \textbf{91.1{\scriptsize$\,\pm\,$0.4}} & \textbf{92.6{\scriptsize$\,\pm\,$1.3}} & \textbf{90.6{\scriptsize$\,\pm\,$0.2}} & \textbf{87.6{\scriptsize$\,\pm\,$0.2}} \\
 & Zero-one & 0.74{\scriptsize$\,\pm\,$0.05} & 73.0{\scriptsize$\,\pm\,$0.6} & 78.1{\scriptsize$\,\pm\,$1.8} & 80.9{\scriptsize$\,\pm\,$3.2} & 77.3{\scriptsize$\,\pm\,$0.6} & 73.8{\scriptsize$\,\pm\,$0.8} \\
 & Brier & \textbf{0.73{\scriptsize$\,\pm\,$0.05}} & 87.0{\scriptsize$\,\pm\,$0.1} & 90.4{\scriptsize$\,\pm\,$0.2} & 91.9{\scriptsize$\,\pm\,$1.2} & 89.8{\scriptsize$\,\pm\,$0.2} & 86.9{\scriptsize$\,\pm\,$0.1} \\
\midrule
Dropout & Entropy & 1.04{\scriptsize$\,\pm\,$0.13} & \textbf{83.8{\scriptsize$\,\pm\,$0.2}} & \textbf{94.5{\scriptsize$\,\pm\,$0.3}} & \textbf{89.8{\scriptsize$\,\pm\,$0.9}} & \textbf{86.5{\scriptsize$\,\pm\,$0.4}} & \textbf{86.7{\scriptsize$\,\pm\,$0.5}} \\
 & Zero-one & 1.07{\scriptsize$\,\pm\,$0.04} & 60.8{\scriptsize$\,\pm\,$0.2} & 93.2{\scriptsize$\,\pm\,$3.3} & 63.3{\scriptsize$\,\pm\,$0.1} & 62.6{\scriptsize$\,\pm\,$0.2} & 63.2{\scriptsize$\,\pm\,$0.3} \\
 & Brier & \textbf{1.00{\scriptsize$\,\pm\,$0.10}} & 81.6{\scriptsize$\,\pm\,$0.0} & 93.7{\scriptsize$\,\pm\,$0.3} & 86.9{\scriptsize$\,\pm\,$0.8} & 84.5{\scriptsize$\,\pm\,$0.4} & 84.6{\scriptsize$\,\pm\,$0.5} \\
\midrule
DropConnect & Entropy & 0.92{\scriptsize$\,\pm\,$0.07} & \textbf{83.0{\scriptsize$\,\pm\,$0.2}} & 93.0{\scriptsize$\,\pm\,$0.2} & \textbf{86.4{\scriptsize$\,\pm\,$0.4}} & \textbf{84.3{\scriptsize$\,\pm\,$0.7}} & \textbf{85.7{\scriptsize$\,\pm\,$0.5}} \\
 & Zero-one & 0.92{\scriptsize$\,\pm\,$0.05} & 60.3{\scriptsize$\,\pm\,$0.1} & 91.9{\scriptsize$\,\pm\,$3.9} & 60.6{\scriptsize$\,\pm\,$0.8} & 62.0{\scriptsize$\,\pm\,$1.2} & 62.7{\scriptsize$\,\pm\,$0.2} \\
 & Brier & \textbf{0.90{\scriptsize$\,\pm\,$0.05}} & 80.4{\scriptsize$\,\pm\,$0.3} & \textbf{93.2{\scriptsize$\,\pm\,$0.1}} & 83.1{\scriptsize$\,\pm\,$0.6} & 82.1{\scriptsize$\,\pm\,$0.9} & 83.3{\scriptsize$\,\pm\,$0.5} \\
\midrule
BNN & Entropy & 1.79{\scriptsize$\,\pm\,$0.11} & \textbf{82.8{\scriptsize$\,\pm\,$0.5}} & \textbf{83.7{\scriptsize$\,\pm\,$2.3}} & 83.6{\scriptsize$\,\pm\,$2.4} & \textbf{83.9{\scriptsize$\,\pm\,$1.1}} & \textbf{82.8{\scriptsize$\,\pm\,$0.2}} \\
 & Zero-one & 1.87{\scriptsize$\,\pm\,$0.09} & 68.2{\scriptsize$\,\pm\,$0.4} & 69.9{\scriptsize$\,\pm\,$3.6} & 71.4{\scriptsize$\,\pm\,$3.4} & 69.5{\scriptsize$\,\pm\,$0.9} & 68.5{\scriptsize$\,\pm\,$0.1} \\
 & Brier & \textbf{1.79{\scriptsize$\,\pm\,$0.12}} & 81.2{\scriptsize$\,\pm\,$0.6} & 82.5{\scriptsize$\,\pm\,$2.6} & \textbf{84.0{\scriptsize$\,\pm\,$1.6}} & 82.3{\scriptsize$\,\pm\,$1.2} & 80.7{\scriptsize$\,\pm\,$0.1} \\
\midrule
Laplace & Entropy & 0.68{\scriptsize$\,\pm\,$0.07} & \textbf{86.8{\scriptsize$\,\pm\,$0.0}} & \textbf{90.1{\scriptsize$\,\pm\,$2.5}} & \textbf{92.2{\scriptsize$\,\pm\,$1.0}} & \textbf{90.1{\scriptsize$\,\pm\,$0.3}} & \textbf{87.9{\scriptsize$\,\pm\,$0.7}} \\
 & Zero-one & 0.68{\scriptsize$\,\pm\,$0.06} & 68.9{\scriptsize$\,\pm\,$0.2} & 71.2{\scriptsize$\,\pm\,$3.9} & 73.3{\scriptsize$\,\pm\,$3.5} & 72.1{\scriptsize$\,\pm\,$1.2} & 70.2{\scriptsize$\,\pm\,$0.3} \\
 & Brier & \textbf{0.68{\scriptsize$\,\pm\,$0.06}} & 86.4{\scriptsize$\,\pm\,$0.1} & 89.7{\scriptsize$\,\pm\,$2.3} & 91.8{\scriptsize$\,\pm\,$1.0} & 89.4{\scriptsize$\,\pm\,$0.4} & 87.4{\scriptsize$\,\pm\,$0.6} \\
\midrule
SNGP & Entropy & 0.86{\scriptsize$\,\pm\,$0.05} & \textbf{84.7{\scriptsize$\,\pm\,$0.3}} & 89.3{\scriptsize$\,\pm\,$1.6} & 88.4{\scriptsize$\,\pm\,$1.5} & \textbf{88.4{\scriptsize$\,\pm\,$0.2}} & \textbf{87.4{\scriptsize$\,\pm\,$0.2}} \\
 & Zero-one & \textbf{0.84{\scriptsize$\,\pm\,$0.05}} & 56.3{\scriptsize$\,\pm\,$0.2} & 58.3{\scriptsize$\,\pm\,$0.7} & 56.8{\scriptsize$\,\pm\,$0.9} & 56.9{\scriptsize$\,\pm\,$0.5} & 58.0{\scriptsize$\,\pm\,$0.1} \\
 & Brier & 0.84{\scriptsize$\,\pm\,$0.05} & 84.6{\scriptsize$\,\pm\,$0.3} & \textbf{89.3{\scriptsize$\,\pm\,$1.6}} & \textbf{88.5{\scriptsize$\,\pm\,$1.4}} & 88.3{\scriptsize$\,\pm\,$0.1} & 87.1{\scriptsize$\,\pm\,$0.2} \\
\midrule
DARE & Entropy & \textbf{0.63{\scriptsize$\,\pm\,$0.10}} & \textbf{87.9{\scriptsize$\,\pm\,$0.3}} & \textbf{92.4{\scriptsize$\,\pm\,$0.5}} & \textbf{92.9{\scriptsize$\,\pm\,$1.9}} & \textbf{90.5{\scriptsize$\,\pm\,$0.4}} & \textbf{90.0{\scriptsize$\,\pm\,$0.2}} \\
 & Zero-one & 0.74{\scriptsize$\,\pm\,$0.09} & 60.0{\scriptsize$\,\pm\,$0.7} & 62.7{\scriptsize$\,\pm\,$0.5} & 61.8{\scriptsize$\,\pm\,$3.4} & 61.5{\scriptsize$\,\pm\,$1.4} & 62.1{\scriptsize$\,\pm\,$0.8} \\
 & Brier & 0.67{\scriptsize$\,\pm\,$0.09} & 79.5{\scriptsize$\,\pm\,$0.6} & 85.6{\scriptsize$\,\pm\,$0.8} & 85.6{\scriptsize$\,\pm\,$3.2} & 82.8{\scriptsize$\,\pm\,$1.1} & 82.1{\scriptsize$\,\pm\,$1.1} \\
\bottomrule
\end{tabular}
\end{table}

\begin{table}[t]
\centering
\caption{Ablation on the uncertainty measure on ImageNet. Selective prediction (SP) is reported as the area under the risk-coverage curve (AURC, lower is better). \ac{ood} detection is reported as the AUROC (higher is better) for each \ac{ood} dataset. Mean $\pm$ standard deviation over three seeds with the best measure per method and column in bold.}
\label{tab:ablation-measure-imagenet}
\footnotesize
\setlength{\tabcolsep}{3.5pt}
\begin{tabular}{llccccc}
\toprule
 & & SP & \multicolumn{4}{c}{\ac{ood} detection, AUROC $\uparrow$} \\
\cmidrule(lr){3-3} \cmidrule(lr){4-7}
Method & Measure & AURC $\downarrow$ & SSB-hard & NINCO & Textures & iNaturalist \\
\midrule
Deep Ensemble & Entropy & 6.55{\scriptsize$\,\pm\,$0.03} & \textbf{71.7{\scriptsize$\,\pm\,$0.1}} & \textbf{84.9{\scriptsize$\,\pm\,$0.1}} & \textbf{82.5{\scriptsize$\,\pm\,$0.3}} & \textbf{81.5{\scriptsize$\,\pm\,$0.2}} \\
 & Zero-one & 6.76{\scriptsize$\,\pm\,$0.03} & 66.3{\scriptsize$\,\pm\,$0.1} & 78.0{\scriptsize$\,\pm\,$0.1} & 76.2{\scriptsize$\,\pm\,$0.3} & 75.3{\scriptsize$\,\pm\,$0.1} \\
 & Brier & \textbf{6.49{\scriptsize$\,\pm\,$0.03}} & 70.5{\scriptsize$\,\pm\,$0.1} & 78.7{\scriptsize$\,\pm\,$0.1} & 79.7{\scriptsize$\,\pm\,$0.4} & 78.5{\scriptsize$\,\pm\,$0.1} \\
\midrule
SubEnsemble & Entropy & 9.80{\scriptsize$\,\pm\,$0.00} & \textbf{55.9{\scriptsize$\,\pm\,$0.1}} & \textbf{80.3{\scriptsize$\,\pm\,$0.2}} & \textbf{81.5{\scriptsize$\,\pm\,$0.2}} & \textbf{65.1{\scriptsize$\,\pm\,$0.2}} \\
 & Zero-one & \textbf{7.52{\scriptsize$\,\pm\,$0.00}} & 51.3{\scriptsize$\,\pm\,$0.0} & 56.9{\scriptsize$\,\pm\,$0.3} & 53.7{\scriptsize$\,\pm\,$0.1} & 52.6{\scriptsize$\,\pm\,$0.0} \\
 & Brier & 7.71{\scriptsize$\,\pm\,$0.00} & 55.1{\scriptsize$\,\pm\,$0.2} & 66.8{\scriptsize$\,\pm\,$0.5} & 70.8{\scriptsize$\,\pm\,$0.5} & 60.8{\scriptsize$\,\pm\,$0.2} \\
\midrule
BatchEnsemble & Entropy & \textbf{8.79{\scriptsize$\,\pm\,$0.06}} & \textbf{68.4{\scriptsize$\,\pm\,$0.2}} & \textbf{81.8{\scriptsize$\,\pm\,$1.1}} & \textbf{75.2{\scriptsize$\,\pm\,$0.2}} & \textbf{77.1{\scriptsize$\,\pm\,$0.1}} \\
 & Zero-one & 8.87{\scriptsize$\,\pm\,$0.07} & 52.0{\scriptsize$\,\pm\,$0.1} & 55.0{\scriptsize$\,\pm\,$1.0} & 53.3{\scriptsize$\,\pm\,$0.5} & 53.6{\scriptsize$\,\pm\,$0.1} \\
 & Brier & 8.83{\scriptsize$\,\pm\,$0.06} & 67.3{\scriptsize$\,\pm\,$0.1} & 78.7{\scriptsize$\,\pm\,$0.8} & 73.0{\scriptsize$\,\pm\,$0.4} & 74.9{\scriptsize$\,\pm\,$0.1} \\
\midrule
Dropout & Entropy & 7.95{\scriptsize$\,\pm\,$0.05} & \textbf{70.7{\scriptsize$\,\pm\,$0.1}} & \textbf{84.0{\scriptsize$\,\pm\,$0.5}} & \textbf{76.9{\scriptsize$\,\pm\,$0.7}} & \textbf{79.1{\scriptsize$\,\pm\,$0.1}} \\
 & Zero-one & 8.50{\scriptsize$\,\pm\,$0.03} & 64.7{\scriptsize$\,\pm\,$0.1} & 75.3{\scriptsize$\,\pm\,$0.6} & 70.2{\scriptsize$\,\pm\,$0.6} & 71.8{\scriptsize$\,\pm\,$0.2} \\
 & Brier & \textbf{7.93{\scriptsize$\,\pm\,$0.05}} & 68.7{\scriptsize$\,\pm\,$0.2} & 75.9{\scriptsize$\,\pm\,$1.0} & 73.7{\scriptsize$\,\pm\,$0.3} & 75.2{\scriptsize$\,\pm\,$0.2} \\
\midrule
DropConnect & Entropy & 8.08{\scriptsize$\,\pm\,$0.07} & \textbf{70.5{\scriptsize$\,\pm\,$0.2}} & \textbf{82.4{\scriptsize$\,\pm\,$0.7}} & \textbf{76.0{\scriptsize$\,\pm\,$0.5}} & \textbf{78.6{\scriptsize$\,\pm\,$0.2}} \\
 & Zero-one & 8.54{\scriptsize$\,\pm\,$0.04} & 64.4{\scriptsize$\,\pm\,$0.1} & 74.2{\scriptsize$\,\pm\,$0.5} & 68.6{\scriptsize$\,\pm\,$0.2} & 71.3{\scriptsize$\,\pm\,$0.2} \\
 & Brier & \textbf{8.06{\scriptsize$\,\pm\,$0.07}} & 68.3{\scriptsize$\,\pm\,$0.3} & 74.7{\scriptsize$\,\pm\,$0.6} & 72.7{\scriptsize$\,\pm\,$0.4} & 74.7{\scriptsize$\,\pm\,$0.1} \\
\midrule
BNN & Entropy & \textbf{7.02{\scriptsize$\,\pm\,$0.03}} & \textbf{69.2{\scriptsize$\,\pm\,$0.2}} & \textbf{80.3{\scriptsize$\,\pm\,$0.7}} & \textbf{77.4{\scriptsize$\,\pm\,$0.1}} & \textbf{78.1{\scriptsize$\,\pm\,$0.1}} \\
 & Zero-one & 8.00{\scriptsize$\,\pm\,$0.09} & 57.7{\scriptsize$\,\pm\,$0.1} & 64.9{\scriptsize$\,\pm\,$1.3} & 62.9{\scriptsize$\,\pm\,$0.2} & 63.2{\scriptsize$\,\pm\,$0.1} \\
 & Brier & 7.05{\scriptsize$\,\pm\,$0.04} & 66.9{\scriptsize$\,\pm\,$0.2} & 75.4{\scriptsize$\,\pm\,$0.1} & 73.6{\scriptsize$\,\pm\,$0.3} & 74.9{\scriptsize$\,\pm\,$0.1} \\
\midrule
Laplace & Entropy & 8.43{\scriptsize$\,\pm\,$0.15} & \textbf{70.0{\scriptsize$\,\pm\,$0.3}} & \textbf{87.0{\scriptsize$\,\pm\,$1.2}} & \textbf{84.2{\scriptsize$\,\pm\,$0.7}} & \textbf{79.9{\scriptsize$\,\pm\,$0.4}} \\
 & Zero-one & \textbf{7.77{\scriptsize$\,\pm\,$0.11}} & 69.4{\scriptsize$\,\pm\,$0.4} & 80.9{\scriptsize$\,\pm\,$2.1} & 80.3{\scriptsize$\,\pm\,$0.8} & 77.7{\scriptsize$\,\pm\,$0.8} \\
 & Brier & 7.89{\scriptsize$\,\pm\,$0.12} & 69.5{\scriptsize$\,\pm\,$0.5} & 75.8{\scriptsize$\,\pm\,$2.7} & 80.7{\scriptsize$\,\pm\,$0.8} & 76.9{\scriptsize$\,\pm\,$1.0} \\
\midrule
SNGP & Entropy & 10.80{\scriptsize$\,\pm\,$0.15} & \textbf{68.3{\scriptsize$\,\pm\,$0.1}} & \textbf{82.2{\scriptsize$\,\pm\,$0.9}} & \textbf{75.2{\scriptsize$\,\pm\,$0.5}} & \textbf{78.1{\scriptsize$\,\pm\,$0.3}} \\
 & Zero-one & \textbf{10.39{\scriptsize$\,\pm\,$0.15}} & 52.3{\scriptsize$\,\pm\,$0.2} & 52.0{\scriptsize$\,\pm\,$0.8} & 47.5{\scriptsize$\,\pm\,$0.5} & 50.6{\scriptsize$\,\pm\,$0.3} \\
 & Brier & 10.50{\scriptsize$\,\pm\,$0.16} & 58.7{\scriptsize$\,\pm\,$0.1} & 53.1{\scriptsize$\,\pm\,$1.0} & 57.1{\scriptsize$\,\pm\,$1.4} & 56.7{\scriptsize$\,\pm\,$1.4} \\
\bottomrule
\end{tabular}
\end{table}

\subsection{Practical Applications of \probly}

\subsubsection{\probly with Large Language Models}\label{sec:appendix:demo:llm}

Uncertainty quantification for \ac{llm} differs substantially from standard predictive settings, since the output space is not a finite-dimensional Euclidean space with a fixed set of labels or coordinates. A common strategy is therefore to generate multiple responses to a given prompt and infer uncertainty from their heterogeneity. In this view, semantically similar generations are grouped together, and uncertainty is quantified from the induced structure over groups or similarities.

\probly supports this setting through two \ac{llm}-specific pipelines that wrap arbitrary Hugging Face causal language models. \emph{Semantic entropy}~\citep{kuhnsemantic} clusters generated responses using an entailment relation produced by a natural language inference model and quantifies uncertainty over the resulting semantic clusters. \emph{Spectral entropy}~\citep{walhaSpectralUncertainty2026} instead embeds the generated responses and derives uncertainty from the spectrum of the corresponding similarity kernel. Both methods follow a shared two-stage structure: first, multiple responses are generated for each prompt; second, these responses are aggregated into an uncertainty score. This design preserves \probly's representer--quantification split, allowing the same downstream uncertainty measures, such as TU, AU, and EU, to be applied across \ac{llm}-based pipelines.

A key implementation challenge is that efficiency must be addressed not only in the uncertainty computation itself, but also in the querying procedure. Since both methods require multiple generations per prompt, naive evaluation can incur substantial computational overhead, making batching, caching, and careful model invocation central to practical use.

\begin{figure}[h]
\centering
\includegraphics[width=\linewidth]{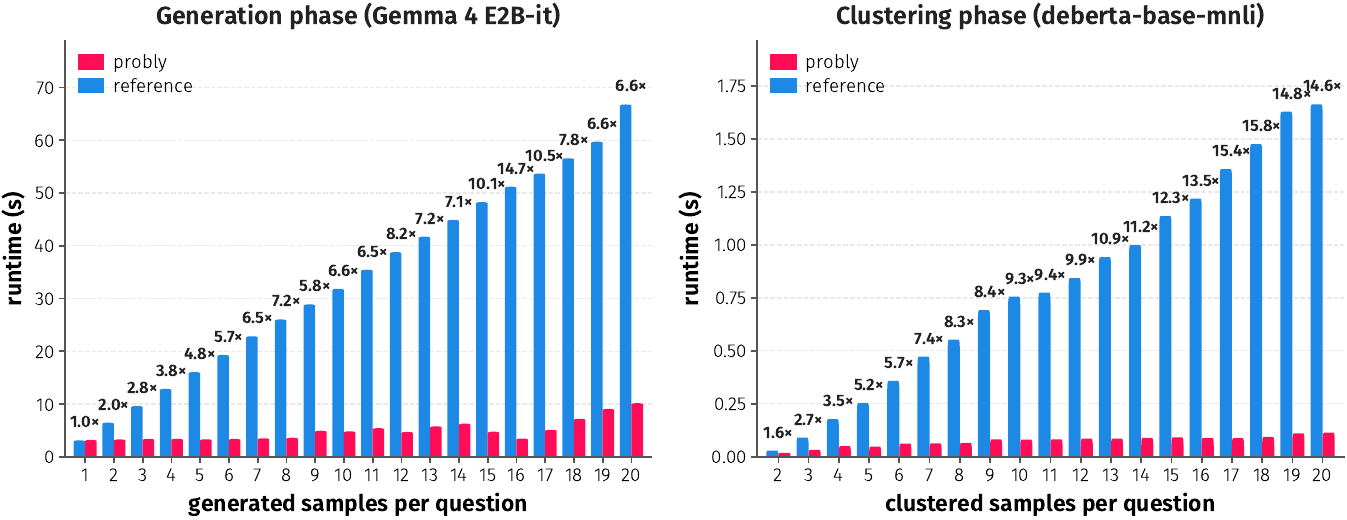}
\caption{Per-step runtime of \probly{}'s LLM uncertainty pipeline against an
unbatched reference, across the full sweep of generated / clustered samples
per question. \emph{Left:} response generation with Gemma~4~E2B-it (1--20
samples per question, two questions). \emph{Right:} NLI clustering with
deberta-base-mnli (2--20 samples per question). Numbers above each bar pair
report the speedup of the batched \probly{} path over the reference. Both
benchmarks were run on a single NVIDIA RTX 4090.}
\label{fig:demollm_full}
\end{figure}

\paragraph{Setup.}
We benchmark the two cost-dominant stages of the semantic-entropy pipeline against a naive, unbatched reference implementation that processes one generated response or one NLI pair at a time, following the pseudocode of the original method. Response generation uses Google's Gemma~4~E2B-it, while NLI-based clustering uses \texttt{microsoft/deberta-base-mnli}. All experiments are run on a single NVIDIA RTX~4090. For each configuration, we report the median runtime across three generation runs and five NLI repetitions, and define speedup as the ratio between the corresponding median runtimes. \Cref{fig:demollm_full} reports the full sweep over $1$--$20$ generated responses per question for generation and $2$--$20$ clustered responses per question for NLI clustering.

\paragraph{Findings.}
The unbatched reference implementation scales approximately linearly with the number of samples, whereas \probly's batched implementation keeps per-step latency nearly constant over the measured range. At the largest configuration, \probly achieves a $6.6\times$ end-to-end speedup for response generation and a $14.6\times$ speedup for NLI clustering. For semantic entropy with $20$ samples, this reduces the user-visible runtime from approximately $66$~s to $10$~s for generation and from $1.66$~s to $0.11$~s for clustering, without changing the user-facing API shown in \Cref{fig:demollm}. The speedup increases monotonically over the full range of evaluated sample counts, with no observed saturation up to $20$ responses per question.

\begin{figure}[h]
\centering
\includegraphics[width=\linewidth]{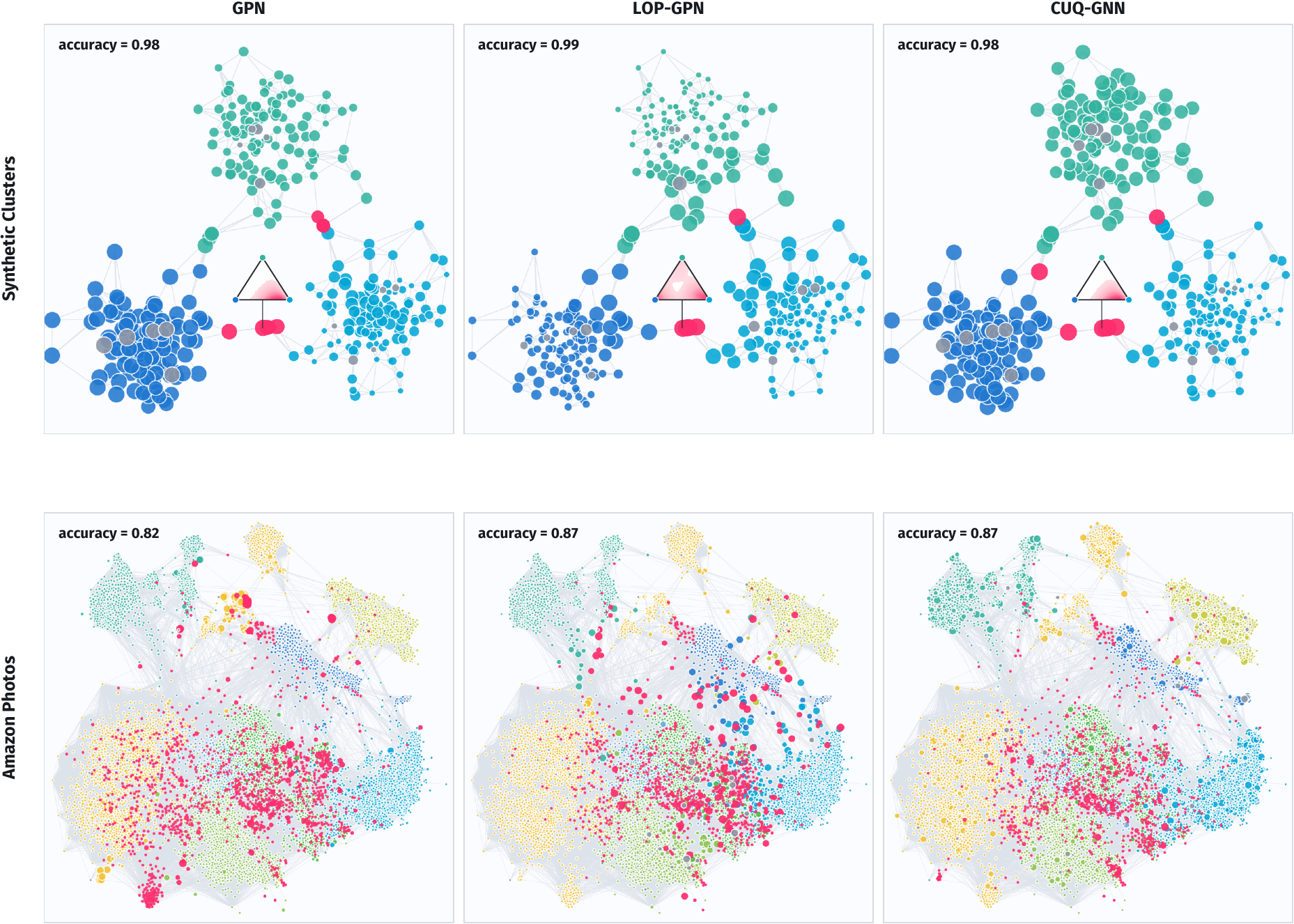}
\caption{%
    \Acl*{eu} of \acs*{gpn}~\citep{stadler2021}, \acs*{lopgpn}~\citep{damke2024}, and \acs*{cuqgnn}~\citep{damke2024a}
    on a small-scale synthetic cluster dataset and on the Amazon Photos dataset \citep{mcauley2015}.%
    Vertex size indicated \acs*{eu}, misclassified vertices are colored red.
    On the synthetic datasets, the predicted second-order distributions of the three models are shown for a single misclassified vertex.
    The misclassification of \acs*{gpn} and \acs*{cuqgnn} are symptoms of predicting a Dirichlet distribution with a mode close to the wrong class.
    \Acs*{lopgpn} also misclassifies this vertex but its \acs*{eu} is induced by a mixture over Dirichlet distributions with modes at all three classes, reflecting the fact that the misclassified node is in the low density region between clusters.
}
\label{fig:demogpn_full}
\end{figure}

\subsubsection{\probly with \river Data Stream Models}\label{sec:appendix:demo:data_stream}

Online learning from incremental data streams differs fundamentally from traditional supervised machine learning \cite{Gama.2014}. Whereas classical models are fit on a fixed dataset that is assumed to capture the underlying data distribution, an online learner observes only one sample at a time and updates its parameters \textit{incrementally} as the stream unfolds. The standard evaluation protocol is \emph{prequential}: at each step the learner first predicts on the new sample which yields a running performance metric. Then, the learner performs an update from the labeled instance. As the underlying distribution may change while the learner is running, the notion of \emph{concept drift} becomes central. Concept drift is any change in the joint distribution $p(X, Y)$ over time. The decomposition $p(X, Y) = p(X)\, p(Y \mid X)$ splits drift into two regimes: \emph{virtual drift}, where only the input distribution $p(X)$ shifts while the labeling rule $p(Y \mid X)$ stays fixed, and \emph{real concept drift}, where $p(Y \mid X)$ itself changes.

\begin{figure}[ht]
    \centering
    \begin{minipage}{0.6\linewidth}
        \centering
        \includegraphics[width=\linewidth]{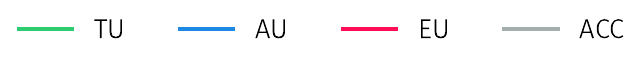}
    \end{minipage}\\[4pt]
    \begin{minipage}[t]{0.49\linewidth}
        \centering
        \includegraphics[width=\linewidth]{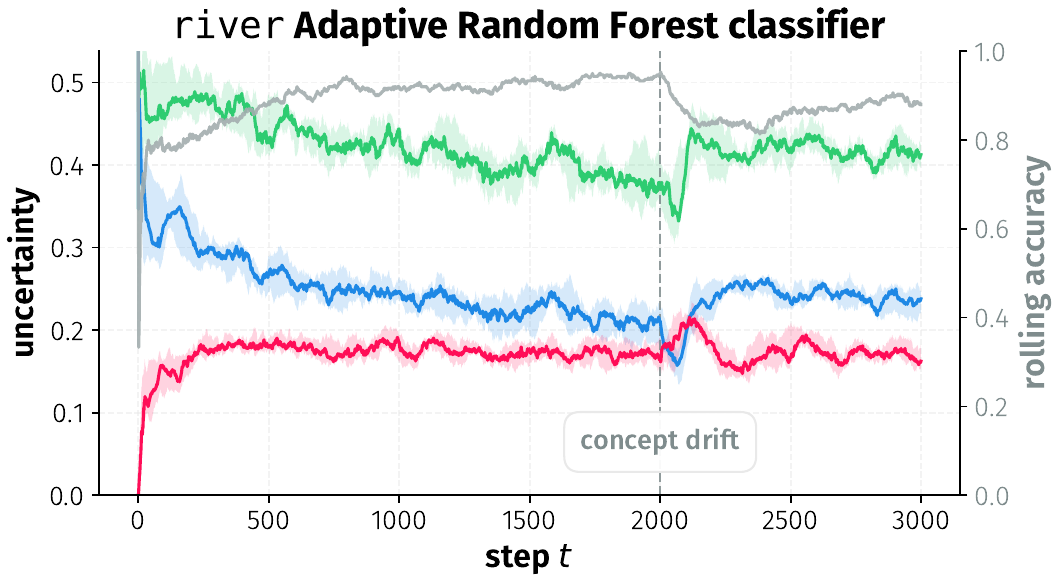}
        \subcaption{Abrupt virtual concept drift.}
        \label{fig:appendix:demo:data_stream:abrupt}
    \end{minipage}\hfill
    \begin{minipage}[t]{0.49\linewidth}
        \centering
        \includegraphics[width=\linewidth]{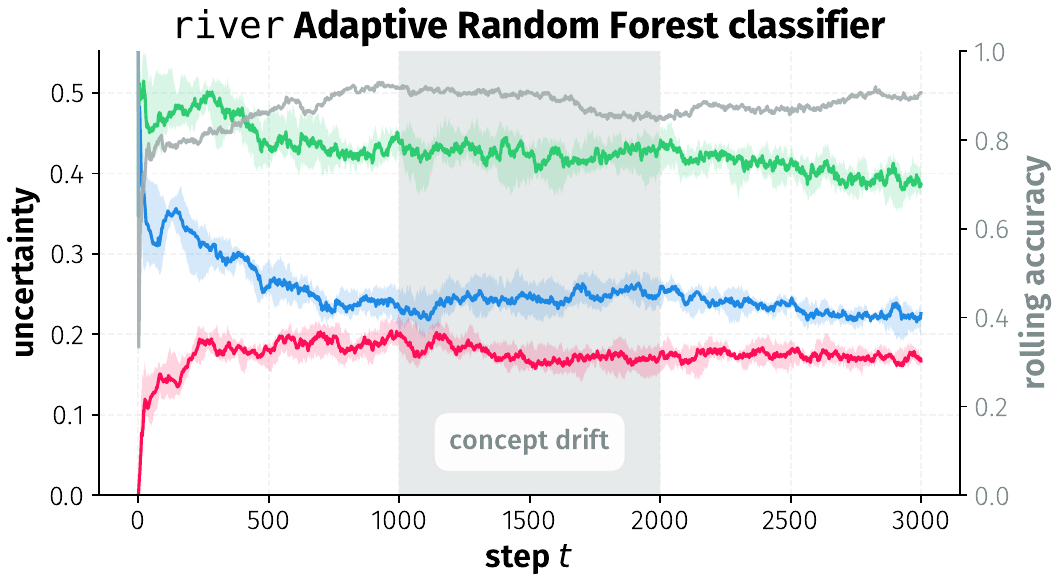}
        \subcaption{Gradual virtual concept drift.}
        \label{fig:appendix:demo:data_stream:gradual}
    \end{minipage}
    \vspace{0.6em}
    \begin{minipage}[t]{0.49\linewidth}
        \centering
        \includegraphics[width=\linewidth]{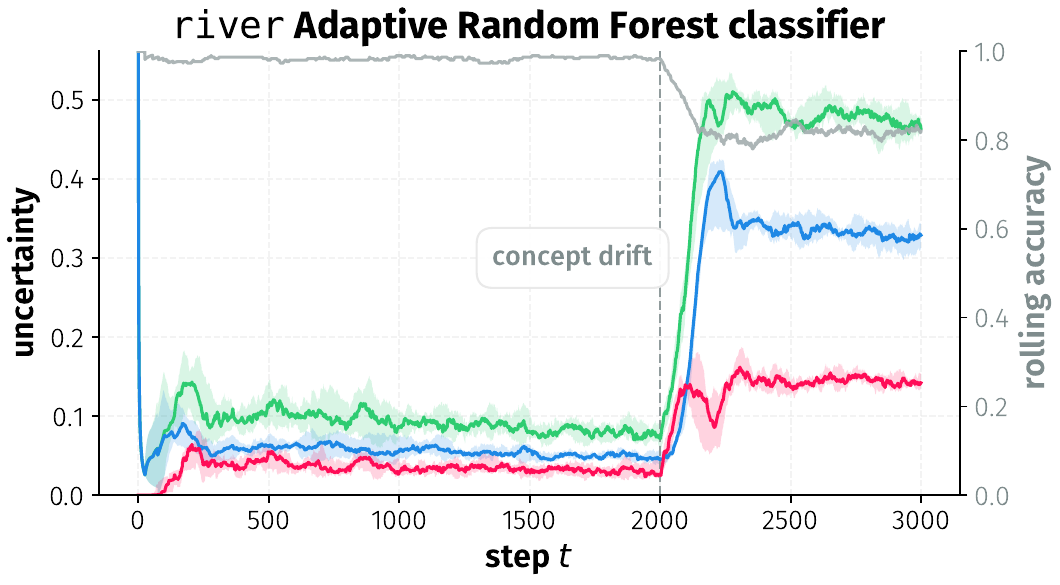}
        \subcaption{Real concept drift $f_7\!\to\!f_4$.}
        \label{fig:appendix:demo:data_stream:7to4}
    \end{minipage}\hfill
    \begin{minipage}[t]{0.49\linewidth}
        \centering
        \includegraphics[width=\linewidth]{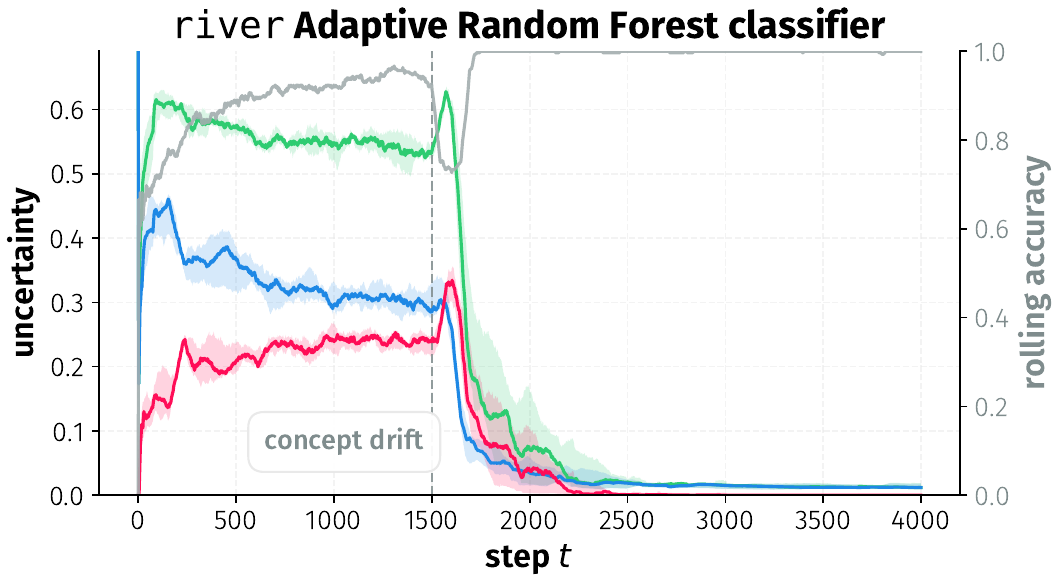}
        \subcaption{Real concept drift $f_0\!\to\!f_9$.}
        \label{fig:appendix:demo:data_stream:0to9}
    \end{minipage}
    \caption{TU/AU/EU decomposition with rolling accuracy on the right axis under the four drift scenarios described in \cref{sec:appendix:demo:data_stream}. \emph{Top row -- virtual drift:} the abrupt jump in \cref{fig:appendix:demo:data_stream:abrupt} produces a small EU spike that quickly subsides, while the gradual fade in \cref{fig:appendix:demo:data_stream:gradual} leaves EU largely unaffected. \emph{Bottom row -- real concept drift:} \cref{fig:appendix:demo:data_stream:7to4} shows that the \ac{arf} is unable to fully relearn $f_4$ after the rule flip, substantially increasing uncertainty in the model; in \cref{fig:appendix:demo:data_stream:0to9} the simple linear-threshold rule $f_9$ lets the ensemble converge and EU reduces to zero.}
    \label{fig:appendix:demo:data_stream}
\end{figure}

\paragraph{Setup.}
We use \river's synthetic \textit{Agrawal} \cite{AgrawalIS93} loan-approval stream and an \ac{arf} classifier with $K{=}10$ trees. Each panel of \cref{fig:appendix:demo:data_stream,fig:demo:data_stream} is run prequentially. At every step the model first predicts, \probly returns the TU/AU/EU decomposition, and then the model is updated. Every panel aggregates ten random seeds, plotting the per-step median and interquartile range of the performance and uncertainty scores. The Agrawal generator consists ten classification functions $f_0, \dots, f_9$. \cref{tab:appendix:demo:drift_configs} summarizes the pre- and post-drift conditions of the four stream scenarios in \cref{fig:appendix:demo:data_stream}.

\begin{table}[H]
    \centering
    \small
    \caption{Drift configurations for the four panels of \cref{fig:appendix:demo:data_stream}. $\mathcal{S}_+ = \{\text{age} \geq 40 \wedge \text{salary} \geq \$75\text{k}\}$ and $\mathcal{S}_- = \{\text{age} < 40 \wedge \text{salary} < \$75\text{k}\}$ are disjoint Agrawal \citep{AgrawalIS93} subpopulations.}
    \vspace{0.5em}
    \label{tab:appendix:demo:drift_configs}
    \begin{tabular}{@{}llllc@{}}
        \toprule
        \textbf{Panel} & \textbf{Drift type} & $\bm{p(Y \mid X)}$ & $\bm{p(X)}$ & \textbf{Drift Location} \\
        \midrule
        \cref{fig:appendix:demo:data_stream:abrupt}  & virtual & $f_4$ (fixed)  & $\mathcal{S}_+ \to \mathcal{S}_-$ & abrupt, $t{=}2000$ \\
        \cref{fig:appendix:demo:data_stream:gradual} & virtual & $f_4$ (fixed)  & $\mathcal{S}_+ \to \mathcal{S}_-$ & sigmoid fade, $t \in [1000, 2000]$ \\
        \cref{fig:appendix:demo:data_stream:7to4}    & real    & $f_7 \to f_4$ & full Agrawal (fixed) & abrupt, $t{=}2000$ \\
        \cref{fig:appendix:demo:data_stream:0to9}    & real    & $f_0 \to f_9$ & full Agrawal (fixed) & abrupt, $t{=}1500$ \\
        \bottomrule
    \end{tabular}
\end{table}

\section{Guidance for Practitioners}
The benchmark in \cref{sec:benchmark} compares many different uncertainty approaches, but does not show a single best method. We will here provide guidance for practitioners on the selection of uncertainty methods for the different tasks that were considered. In particular, we only consider the methods and settings evaluated in \cref{sec:benchmark}, as we do not wish to hypothesize about possible empirical behaviors outside of these. Importantly, any choice involves inherent trade-offs; we will elaborate on those as we discuss the methods.

We start with a discussion of the typical downstream tasks of selective prediction, out-of-distribution detection, and active learning from \cref{sec:bench-sp,sec:bench-ood,sec:bench-al}, respectively. Here, conflicting rankings are to be expected, since the three tasks evaluate different components of the uncertainty methods. Selective prediction tests the ability of the total uncertainty component to establish an ordinal ranking of in-distribution data. Out-of-distribution detection assesses the epistemic component of a representation and quantification combination, and rewards approaches that respond strongly to changes in data distribution. Active learning has a different goal altogether, a well-performing uncertainty approach should sample the most informative instances, that is, those instances that provide the most information upon receiving the correct label. Due to these different objectives, no universal ranking can be established \citep{hofmanOneSize2026}. If the goal is to perform optimally on a single task, one can take the best performing method for the respective task. However, if one requires a method that is performant across all tasks, deep ensembles are consistently amongst the best-performing methods. This finding is aligned with the literature on ensembles, which consistently shows good performance of these methods \citep{snoekPredictiveUncertaintyShift2019}. If efficiency is an additional requirement, the deep deterministic uncertainty (DDU) approach is a sound alternative as it only requires the training of a single neural network with minor modification to the original architecture.

The first-order experiments in \cref{sec:bench-fo} require a more subtle treatment as they inherently optimize two criteria: coverage and efficiency. For the credal predictors, the relative likelihood-based predictors (CreRL and EffCre) and credal ensembling allow one to navigate the Pareto front of coverage and efficiency according to the users’ requirements, while the other methods offer a fixed coverage and efficiency. For conformal prediction, a similar trade-off arises. Any conformal predictor attains the marginal coverage guarantee by construction, so the guarantee itself does not discriminate between methods; the relevant question is which one achieves it most efficiently, i.e. with the smallest prediction sets, where LAC \citep{sadinle2019least} typically performs best. A second criterion is adaptivity: whether set size reflects the instance-wise difficulty of the estimate, growing for hard inputs and shrinking for easy ones; here APS \citep{romanoAPS2020}, RAPS \citep{angelopoulosRAPS2021}, and SAPS \citep{huang2024conformal} are preferable to LAC. Finally, although conformal prediction does not provide conditional coverage guarantees, a method that empirically comes closer to conditional coverage, as measured by the conditional-coverage criterion we use in the paper, is preferable, and in that case APS \citep{romanoAPS2020} clearly outperforms the others.

\clearpage
\section*{NeurIPS Paper Checklist}

\begin{enumerate}

\item {\bf Claims}
    \item[] Question: Do the main claims made in the abstract and introduction accurately reflect the paper's contributions and scope?
    \item[] Answer: \answerYes{} 
    \item[] Justification: Our claims are supported by \cref{sec:practical,sec:benchmark,sec:showcase}. 

\item {\bf Limitations}
    \item[] Question: Does the paper discuss the limitations of the work performed by the authors?
    \item[] Answer: \answerYes{} 
    \item[] Justification: The limitations are discussed in \cref{sec:conclusion}.

\item {\bf Theory assumptions and proofs}
    \item[] Question: For each theoretical result, does the paper provide the full set of assumptions and a complete (and correct) proof?
    \item[] Answer: \answerNA{} 
    \item[] Justification: We do not provide theoretical results.

    \item {\bf Experimental result reproducibility}
    \item[] Question: Does the paper fully disclose all the information needed to reproduce the main experimental results of the paper to the extent that it affects the main claims and/or conclusions of the paper (regardless of whether the code and data are provided or not)?
    \item[] Answer: \answerYes{} 
    \item[] Justification: The full experimental details are discussed in \cref{app:implementation-details,sec:appendix_data_availability_reproducibility}.

\item {\bf Open access to data and code}
    \item[] Question: Does the paper provide open access to the data and code, with sufficient instructions to faithfully reproduce the main experimental results, as described in supplemental material?
    \item[] Answer: \answerYes{} 
    \item[] Justification: Code is provided.

\item {\bf Experimental setting/details}
    \item[] Question: Does the paper specify all the training and test details (e.g., data splits, hyperparameters, how they were chosen, type of optimizer) necessary to understand the results?
    \item[] Answer: \answerYes{} 
    \item[] Justification: The full details are described in \cref{app:implementation-details,sec:appendix_data_availability_reproducibility}.

\item {\bf Experiment statistical significance}
    \item[] Question: Does the paper report error bars suitably and correctly defined or other appropriate information about the statistical significance of the experiments?
    \item[] Answer: \answerYes{} 
    \item[] Justification: We provide information about the error bars for every figure and table.

\item {\bf Experiments compute resources}
    \item[] Question: For each experiment, does the paper provide sufficient information on the computer resources (type of compute workers, memory, time of execution) needed to reproduce the experiments?
    \item[] Answer: \answerYes{} 
    \item[] Justification: We provide information on the compute resources in \cref{sec:appendix_data_availability_reproducibility}.
    
\item {\bf Code of ethics}
    \item[] Question: Does the research conducted in the paper conform, in every respect, with the NeurIPS Code of Ethics \url{https://neurips.cc/public/EthicsGuidelines}?
    \item[] Answer: \answerYes{} 
    \item[] Justification: We have read the NeurIPS Code of Ethics and confirm that our work complies with it.

\item {\bf Broader impacts}
    \item[] Question: Does the paper discuss both potential positive societal impacts and negative societal impacts of the work performed?
    \item[] Answer: \answerYes{} 
    \item[] Justification: We provide broader impacts statements in \cref{sec:conclusion}
    
\item {\bf Safeguards}
    \item[] Question: Does the paper describe safeguards that have been put in place for responsible release of data or models that have a high risk for misuse (e.g., pre-trained language models, image generators, or scraped datasets)?
    \item[] Answer: \answerNA{} 
    \item[] Justification: We do not contribute any such models or datasets.

\item {\bf Licenses for existing assets}
    \item[] Question: Are the creators or original owners of assets (e.g., code, data, models), used in the paper, properly credited and are the license and terms of use explicitly mentioned and properly respected?
    \item[] Answer: \answerYes{} 
    \item[] Justification: We provide references to the models used and additionally include licenses for datasets in \cref{sec:appendix_data_availability_reproducibility}.

\item {\bf New assets}
    \item[] Question: Are new assets introduced in the paper well documented and is the documentation provided alongside the assets?
    \item[] Answer: \answerNA{} 
    \item[] Justification: We do not introduce any new assets.

\item {\bf Crowdsourcing and research with human subjects}
    \item[] Question: For crowdsourcing experiments and research with human subjects, does the paper include the full text of instructions given to participants and screenshots, if applicable, as well as details about compensation (if any)? 
    \item[] Answer: \answerNA{} 
    \item[] Justification: This paper does not involve crowdsourcing nor research with human subjects.

\item {\bf Institutional review board (IRB) approvals or equivalent for research with human subjects}
    \item[] Question: Does the paper describe potential risks incurred by study participants, whether such risks were disclosed to the subjects, and whether Institutional Review Board (IRB) approvals (or an equivalent approval/review based on the requirements of your country or institution) were obtained?
    \item[] Answer: \answerNA{} 
    \item[] Justification: This paper does not involve crowdsourcing nor research with human subjects.

\item {\bf Declaration of LLM usage}
    \item[] Question: Does the paper describe the usage of LLMs if it is an important, original, or non-standard component of the core methods in this research? Note that if the LLM is used only for writing, editing, or formatting purposes and does \emph{not} impact the core methodology, scientific rigor, or originality of the research, declaration is not required.
    \item[] Answer: \answerNA{} 
    \item[] Justification: LLMs are not a core component of this research.

\end{enumerate}

\end{document}